\documentclass{article}

\usepackage{microtype}
\usepackage{graphicx}
\usepackage{subfigure}
\usepackage{booktabs} 

\usepackage{hyperref}
\usepackage[algo2e, inoutnumbered, algoruled,vlined]{algorithm2e} %

\usepackage[accepted]{icml2021}
\usepackage{macros}
\usepackage{thm-restate}

\icmltitlerunning{Optimal Thompson Sampling strategies for support-aware CVaR bandits}

\begin{document}
	
	\twocolumn[
	\icmltitle{Optimal Thompson Sampling strategies for support-aware CVaR bandits}	
	\icmlsetsymbol{equal}{*}
	
\begin{icmlauthorlist}
		\icmlauthor{Dorian Baudry}{sequel}
		\icmlauthor{Romain Gautron}{cirad,cgiar}
		\icmlauthor{Emilie Kaufmann}{sequel}
		\icmlauthor{Odalric-Ambrym Maillard}{sequel}
\end{icmlauthorlist}
	
\icmlaffiliation{sequel}{Univ. Lille, CNRS, Inria, Centrale Lille, UMR 9198-CRIStAL, F-59000 Lille, France}
\icmlaffiliation{cirad}{CIRAD, UPR AIDA, F-34398 Montpellier, France}
\icmlaffiliation{cgiar}{CGIAR Platform for Big Data in Agriculture, Alliance of CIAT and Bioversity International, Km 17 Recta Cali-Palmira, Apartado Aéreo 6713, Cali, Colombia}

\icmlcorrespondingauthor{Dorian Baudry}{dorian.baudry@inria.fr}
	
	\icmlkeywords{Bandits, Thompson Sampling, Risk-Awareness, CVaR}
	
	\vskip 0.3in
	]
	
	
	\printAffiliationsAndNotice{}  


\begin{abstract}
	In this paper we study a multi-arm bandit problem in which the quality of each arm is measured by the \textit{Conditional Value at Risk} (CVaR) at some level $\alpha$ of the reward distribution.
	While existing works in this setting mainly focus on \textit{Upper Confidence Bound} algorithms, we introduce a new \textit{Thompson Sampling} approach for CVaR bandits on bounded rewards that is flexible enough to solve a variety of problems grounded on physical resources. Building on a recent work by \citet{Honda}, we introduce \CVTS{} for continuous bounded rewards and \MCVTS{} for multinomial distributions. On the theoretical side, we provide a non-trivial extension of their analysis that enables to theoretically bound their CVaR regret minimization performance. Strikingly, our results show that these strategies are the first to provably achieve \textit{asymptotic optimality} in CVaR bandits, matching the corresponding asymptotic lower bounds for this setting. Further, we illustrate empirically the benefit of Thompson Sampling approaches both in a realistic environment simulating a use-case in agriculture and on various synthetic examples.
\end{abstract}
\section{Introduction}\label{sec::intro}
Over the past few years, a number of works have focused  on adapting multi-armed bandit strategies (see e.g. \citet{Banditbook}) to optimize an other criterion than the \textit{expected} cumulative reward. 
\citet{sani_mv}, \citet{vakili2}, \citet{vakiliZ16}, \citet{zimin_ra} consider a mean-variance criterion,
 \cite{szorenyi2015qualitative} studies a  quantile  (Value-at-Risk)  criterion, \cite{oda_raucb} focuses on Entropic-value-at-risk.
The \textit{Conditional Value at Risk} (CVaR) as well as more generic \textit{coherent spectral risk measures} \cite{acerbi2002coherence} have received specific attention from the bandit community (\citet{galichet2013exploration,galichet2015contributions, cassel,chang2020riskconstrained,tamkindistributionally,cvar_conc_prashanth}  to cite a few).
Indeed, in a large number of application domains (healthcare, agriculture, marketing,...), one needs to take into account personalized \textit{preferences} of the practitioner that are not captured by the \textit{expected} reward.  We consider an illustrative use-case in agriculture in section~\ref{sec::experiments}, where an algorithm recommends planting dates to farmers.

The \textit{Conditional Value at Risk} (CVaR) at level $\alpha\in[0,1]$ (see \citet{mandelbrot1997variation}, \citet{artzner}) is easily interpretable as the expected reward in the worst $\alpha$-fraction of the outcomes, and hence captures different preferences, from being neutral to the shape of the distribution ($\alpha=1$, mean criterion) to trying to maximize the reward in the worst-case scenarios ($\alpha$ close to 0, typically in finance or insurance). It is further a coherent spectral measure in the sense of \citet{rockafellar2000optimization}, see \citet{acerbi2002coherence}).
%
Several definitions of the CVaR exist in the literature, depending on whether the samples are considered as \textit{losses} or as \textit{rewards}. \citet{brown_ineq}, \citet{thomas_concentration} and \citet{agrawal2020optimal} consider the \textit{loss} version of CVaR.
We here follow \citet{galichet2013exploration} and \citet{tamkindistributionally} who use the \textit{reward} version,
defined  for arm $k$ with distribution $\nu_k$
as \begin{equation}\label{eq::cvar_def}\CVAR_{\alpha}(\nu_k) = \sup_{x \in \R} \left\{x - \frac{1}{\alpha} \bE_{X \sim \nu_k}\left[\left(x-X\right)^+\right] \right\}\;. \end{equation}
This implies that the best arm is the one with the \textit{largest} CVaR. To simplify the notation we write $c_k^\alpha=\CVAR_\alpha(\nu_k)$ in the sequel.
Following e.g. \citet{tamkindistributionally}, for unknown arm distributions $\bm\nu = (\nu_1, \dots, \nu_K)$  we measure the CVaR regret at time $T$ for some risk-level $\alpha$ of a sequential sampling strategy $\cA=(A_t)_{t\in \mathbb{N}}$ as
\begin{align}
\!\!\cR^{\alpha}_{\bm\nu}(T)\!=&\bE_{\bm\nu}\!\!\left[\!\sum_{t=1}^T \!\left(\!\max_k c_k^\alpha \!-\! c_{A_t}^\alpha \!\right) \!\right]\!\!=\!\!
\sum_{k=1}^K \!\!\Delta_k^\alpha \bE_{\bm\nu}[N_k(T)],\!\!\!\!\label{def::regret}\end{align}

where $\Delta_k^\alpha = \max_{k'} c_{k'}^\alpha - c_k^\alpha$ is the gap in CVaR between arm $k$ and the best arm, and $N_k(t) =\sum_{s=1}^{t}\ind(A_s=k)$ is the number of selections of arm $k$ up to round $t$. 
Other notions of regret have been studied for risk-averse bandits, e.g. computing the risk metric of the full trajectory of observed rewards (\citet{sani_mv,cassel,oda_raucb}), but are less interpretable. 

\paragraph{Related work}
At a high level, the multi-armed bandit literature on the CVaR is largely inspired from adapting the popular Upper Confidence Bounds (UCB) algorithms (\citet{auer2002finite}) for bounded distributions to work under this criterion, hence rely on concentration tools for the CVaR.  
Two main approaches can be distinguished: using an empirical CVaR estimate plus a confidence bound as considered in
MaRaB (\citet{galichet2013exploration,galichet2015contributions}, U-UCB \cite{cassel}, or exploiting the link between the CVaR and the CDF to build an optimistic CDF as in \CVAR-UCB \cite{tamkindistributionally}, resorting to the celebrated Dvoretzky--Kiefer--Wolfowitz (DKW) concentration inequality (see \citet{massart1990}). Indeed DKW inequality has been used for example by \citet{brown_ineq} and \citet{thomas_concentration} to develop concentration inequalities for the empirical \CVAR{} of bounded distributions. 
These strategies provably achieve a logarithmic CVaR regret in bandit models with bounded distributions\footnote{\citet{cassel} gives an upper bound on the proxy regret of U-UCB, which is also valid for the smaller CVaR regret. For completeness, we provide in Appendix~\ref{app::brown_ucb} an analysis of U-UCB specifically tailored to the CVaR regret.}, with a scaling in $\frac{K\log T}{\alpha^2 \Delta}$ where $\Delta$ is the smallest (positive) CVaR gap $\Delta_k^{\alpha}$. However, the asymptotic optimality of these strategies is not established.
Strikingly, few works have tried to adapt to the CVaR setting the \textit{asymptotically optimal} bandit strategies for the mean criterion that provably match the lower bound on the regret given by \cite{LaiRobbins85}, such as 
KL-UCB \cite{KL_UCB}, Thompson Sampling (TS) \cite{TS_1933,agrawal13optTS,TS_Emilie} or IMED \cite{Honda15IMED}. 
We note that \citet{chang2020riskconstrained} adapts TS to the slightly different \textit{risk-constrained setting} introduced by \citet{kagrecha20riskconstrained} for which the goal is to maximize the mean rewards under the constraint that arms with a small \CVAR{} are not played too often. Unfortunately the analysis is limited to Gaussian distributions and does not target optimality.
(A TS algorithm was also proposed by \citet{chang2020riskconstrained} for the mean-variance criterion.)

We believe the reason is two-fold: 
First, despite asymptotic optimal strategies being appealing to improve practical performances, such strategies were, until recently, relying on assuming known parametric family (\citet{ HondaTakemura10,Honda15IMED, korda13TSexp, KL_UCB} to name a few), such as one-parameter exponential families, deriving one specific algorithm for each family. Unfortunately, assuming a simple parametric distributions may not be meaningful to model complex, realistic situations.
Rather, the most accessible information to the practitioner is often whether or not the distribution is discrete, and for the continuous case how it is bounded.  
That is typically the case in applications such as agriculture, healthcare, or resource management, when the reward distributions are grounded on physical realities. Indeed the practitioner can realistically assume that the support of the distributions is known and bounded, with bounds that can be  either natural or provided by experts.  For instance, in the use-case we consider in section~\ref{sec::experiments} the algorithm recommends planting dates to farmers to maximize the yield of a maize field, that is naturally bounded. Further, distributions in these settings can have shapes that are not well captured by standard parametric families of distributions, as for instance they can be multi-modal with an unknown number of modes that depend on  external factors unknown at the decision time (weather conditions, illness, pests, \dots). This suggests one may prefer algorithms that can cover a variety of possible shapes for the distributions, rather than targeting a specific known family. UCB-type strategies assuming only boundedness are thus handy even though not optimal.

Second, targeting asymptotic optimality for CVaR bandits is challenging: Massart's bound for DKW-inequality was already a non-trivial result, solving a long-lasting  open question back at the time, and yet only provides a ``Hoeffding version" of the CDF concentration. Adapting this to work e.g. with Kullback-Leibler, plus considering that the CVaR writes as an optimization problem, makes the quest for a tight analysis even more challenging, and providing regret guarantees for a CVaR equivalent of kl-ucb and empirical KL-UCB \cite{KL_UCB} is an interesting direction for future work. Looking at the CVaR community, recent works \citep{KagrechaNJ19,holland2020learning, cvar_conc_prashanth} have developed new tools for CVaR concentration. Unfortunately, they may not be adapted for this purpose since they aim at capturing properties of heavy-tail distributions in a highly risk-averse setup. The setting considered in this paper is different, and applying the optimistic principle for CVaR bandits to achieve asymptotic optimality may be a daunting task. This suggests the idea to turn towards alternative methods, such as e.g. TS strategies.


As it turns out, two powerful variants of TS were introduced recently by \citet{Honda} for the mean criterion, that enable to overcome the ``parametric" limitation, in the sense that these approaches reach the minimal achievable regret given by the lower bound of \citet{burnetas96LB}, respectively for discrete and bounded distributions. This timely contribution opens the room to overcome the two previous limitations and achieve the first provably optimal strategy for CVaR bandit for such practitioner-friendly assumptions.
\begin{remark}
	In finance CVaR is often associated to heavy-tail distributions.
	Other variants of bandits have been considered to deal with possibly heavy-tail distributions, or weak  moment conditions: In \cite{carpentierExtreme}, the authors study regret minimization for extreme statistics (the maximum), for Weibull of Frechet-like distributions. 	In \cite{lattimore2017scale}, a median-of-mean estimator is studied to minimize regret for distributions with bounded kurtosis.  
	A CVaR strategy has been proposed for the different pure exploration setting \citep{KagrechaNJ19,agrawal2020optimal}, under weak moment conditions. These works consider a different setup and objective.
\end{remark}

\paragraph{Contributions}
In this paper, we purposely focus on minimizing the CVaR regret considering either distributions with discrete, finite support, or with continuous and bounded support,
as we believe this has great practical relevance and is still a relatively unexplored topic in the literature.
More precisely, we target first-order \textit{asymptotic optimality} for these (sometimes called ``non-parametric") families and first derive in Theorem~\ref{thm:LB}  a lower-bound on the CVaR regret, adapting that of \cite{LaiRobbins85, burnetas96LB} to the CVaR criterion. This simple result highlights the right complexity term that should appear when deriving regret upper bounds.
We then introduce in Section~\ref{sec::algorithms} \CVTS{} for CVaR bandits with bounded support, and \MCVTS{} for CVaR bandits with \text{multinomial} arms, adapting the strategies proposed by \citet{Honda} for the CVaR.
We provide in Theorem~\ref{th::regret_MCVTS} and Theorem~\ref{th::regret_CVTS}
the regret bound of each algorithm, proving asymptotic optimality of these strategies.
Up to our knowledge, these  are the first results showing asymptotic optimality of a Thompson Sampling based CVaR regret minimization strategy.
As expected, adapting the regret analysis from \citet{Honda} is non-trivial; we highlight the main challenges of this adaption in section~\ref{sec::technical_tools_main}. 
For instance, one of the key challenge was to handle boundary crossing probability for the CVaR, and another difficulty comes in the analysis of the non-parametric  \CVTS{} due to regularity properties of the 
Kulback-Leibler projection.
In Section~\ref{sec::experiments}, we provide a case study in agriculture, making the well-established DSSAT  agriculture simulator \cite{hoogenboom2019dssat} available to the bandit community, and highlight the benefits of using strategies based on Thompson Sampling in this CVaR bandit setting against state-of-the-art baselines: We compare to  U-UCB and CVaR-UCB\footnote{MaRaB is similar to U-UCB but enjoys weaker guarantees.
} as they showcase two fundamentally different approaches to build a UCB strategy for a non-linear utility function. The first one is closely related to UCB, the second one exploits properties of the underlying CDF, which may generalize to different risk metrics. As claimed in \citet{tamkindistributionally}, our experiments confirm that CVaR-UCB generally performs better than U-UCB.
However, both TS strategies outperform UCB algorithms that tend to suffer from non-optimized confidence bounds. 
We complete this study with more classical experiments on synthetic data that also confirm the benefit of TS.

\section{Thompson Sampling Algorithms}\label{sec::algorithms}

We present two novel algorithms based on Thompson Sampling and targeting the lower bound of Theorem~\ref{thm:LB} on the CVaR-regret, for any specified value of $\alpha \in (0,1]$. These algorithms are inspired by the first algorithms based on Thompson Sampling matching the Burnetas and Katehakis lower bound for bounded distributions in the expectation setting, recently proposed by \citet{Honda}. 

\paragraph{Notations} We introduce the notation $C_\alpha(\cX, p)$ for the CVaR of the distribution of support $\cX$ and probability $p\in \cP^{|\cX|}$, where $\cP^n$ denotes the probability simplex of size $n$. 
For a multinomial arm $k$ we denote its known support $\cX_k=(x_k^1, \dots, x_{k}^{M_k})$ for some $M_k\in \N$, and its true probability vector $p_k$. We also define $N_k^i(t)$ as the number of times the algorithm has observed $x_k^i$ for arm $k$ before the time $t$. 
For general bounded distributions we denote $\nu_k$ the distribution of arm $k$ and introduce $\cX_{k, t}$ the set of its observed rewards before time $t$, augmented with a known upper bound $B_k$ for the support of $\nu_k$. We further introduce $\cD_n$ as the uniform distribution on the simplex $\cP^n$, corresponding to the Dirichlet distribution $\mathrm{Dir}((1,...,1)$).

\paragraph{\MCVTS} Thompson Sampling (or posterior sampling) is a general Bayesian principle that can be traced back to the work of \citet{TS_1933}, and that is now investigated for many sequential decision making problems (see \citet{russo18survey} for a survey). Given a prior distribution on the bandit model, Thompson Sampling is a randomized algorithm that selects each arm according to its posterior probability of being optimal. This can be implemented by drawing a possible model from the posterior distribution, and acting optimally in the sampled model. For multinomial distribution \MCVTS{} (Multinomial-CVaR-Thompson-Sampling), described in Algorithm~\ref{algo:MCVTS}, follows this principle. For each arm $k$, $p_k$ is assumed to be drawn from $\cD_{M_k}$, the uniform prior on $\cP^{M_k}$. The posterior distribution at a time $t$ is $\mathrm{Dir}(\beta_{k, t})$, with $\beta_{k, t}=(N_k^{i}(t)+1)_{i \in \{1, \dots, M_k\}}$. At time $t$, \MCVTS{} draws a sample $w_{k, t}\sim \mathrm{Dir}(\beta_{k, t})$ for each arm $k$ and computes $c_{k,t}^\alpha=C_\alpha(\cX_k, w_{k, t})$. Then, it selects $A_{t} = \aargmax_{k} c_{k, t}^\alpha$. For $\alpha=1$, this algorithm coincides with the Multinomial Thompson Sampling algorithm of \citet{Honda}. 
\begin{algorithm}[]
	\caption{\MCVTS \label{algo:MCVTS}}
		\SetKwInput{KwData}{Input} 
		 \KwData{Level $\alpha$, horizon $T$, $K$, supports $\cX_1, \dots, \cX_K$}
		 \SetKwInput{KwResult}{Init.}
		 \KwResult{$t \leftarrow 1$, $\forall k \in \{1, ..., K \}$, $\beta_k=\underbrace{(1, \dots, 1)}_{M_k}$}
		 \vspace{-5.5mm} 
		\For{$ t \in \{2,\dots, T\}$ }{
		\For{$k \in \K$}{
			Draw $w_k \sim Dir(\beta_k)$.\\
			Compute $c_{k, t} =C_\alpha(\cX_k, w_k)$.
		}
		Pull arm $A_t=\aargmax_{k \in \K} c_{k, t}$. \\ Receive reward $r_{t,A_t}$. \\
		Update $\beta_{A_t}(j) = \beta_{A_t}(j) + 1$, for $j$ as $r_{t,A_t}=x_k^j$
}	
\end{algorithm}

 \paragraph{\CVTS} We further introduce the \CVTS{} algorithm (for Bounded-CVaR-Thompson-Sampling) for general bounded distributions. \CVTS, stated as Algorithm~\ref{algo:CVTS}, bears some similarity with a Thompson Sampling algorithm, although it \textit{does not} explicitly use a prior distribution. The algorithm retains the idea of using a noisy version of $\nu_k$, obtained by a \textit{random re-weighting} of the previous observations. Hence, at a time $t$ the index used by the algorithm for an arm $k$ is simply $c_{k, t} = C_\alpha(\cX_{k, t}, w_{k, t})$, where $w_{k, t} \sim \cD_{N_k(t)}$ is drawn uniformly at random in the simplex $\cP^{|\cX_{k, t}|}$. \CVTS{} then selects the arm $A_{t} = \aargmax_{k} c_{k, t}$. For $\alpha = 1$, this algorithm coincides with the Non Parametric Thompson Sampling of \citet{Honda} (NPTS). NPTS can be seen as an algorithm that computes for each arm a random average of the past observations. Our extension to CVAR-bandits required to interpret this operation as the computation of the \emph{expectation} of a \textit{random perturbation} of the empirical distribution, which can be replaced by the computation of the CVaR of this new distribution. Note that this idea generalizes beyond using the CVaR, that  can be replaced with any criterion.

\begin{algorithm}[]
	\caption{\CVTS \label{algo:CVTS}}
	\SetKwInput{KwData}{Input} 
	\KwData{Level $\alpha$, horizon $T$, $K$, upper bounds $B_1, \dots, B_K$}
	\SetKwInput{KwResult}{Init.}
	\KwResult{$t=1$, $\forall k \in \{1, ..., K \}$, $\cX_k=\{B_k\}$, $N_k=1$} 
	\For{$ t \in \{2,\dots, T\}$ }{
		\For{$k \in \K$}{
			Draw $w_k \sim \cD_{N_k}$
			\\ Compute $c_{k, t} =C_\alpha(\cX_k, w_k)$
		}
		Pull arm $A_t=\aargmax_{k \in \K} c_{k, t}$. \\ Receive reward $r_{t,A_t}$. \\
		Update $\cX_{A_t} = \cX_{A_t} \cup \{r_{t, A_t}\}, N_{A_t}=N_{A_t}+1$.
	}	
\end{algorithm}

\begin{remark}
	Interestingly, \CVTS{} also applies to multinomial distributions (that are bounded). The resulting strategy differs from  \MCVTS{} due to the initialization step using the knowledge of the support in \MCVTS. 
\end{remark}
\section{Regret Analysis}\label{sec::analysis}

In this section we prove, after defining this notion, that \MCVTS{} and \CVTS{} are \textit{asymptotically optimal} in terms of the CVaR regret for the distributions they cover. 

\subsection{Asymptotic Optimality in CVaR bandits}\label{subsec::asymp_opt}

\citet{LaiRobbins85} first gave an asymptotic lower bound on the regret for parameteric distribution, that was later extended by \citet{burnetas96LB} to more general classes of distributions. We present below an intuitive generalization of this result for \CVAR{} bandits. 

\begin{definition}
	Let $\cC$ be a class of probability distributions, $\alpha \in (0,1]$, and $\KL(\nu,\nu')$ be the KL-divergence between $\nu\in \cC$ and $\nu' \in \cC$. 
	For any  $\nu\in \cC$ and $c\in \R$, we define \[\cK_{\inf}^{\alpha,\cC} (\nu,c) := \inf_{\nu' \in \cC, \nu' \neq \nu}\left\{ \mathrm{KL}(\nu,\nu') : \emph{CVaR}_{\alpha}(\nu') \geq c\right\}. \] 
\end{definition}

\begin{theorem}[Regret Lower Bound in CVaR bandits]\label{thm:LB}
	Let $\alpha \in (0,1]$. Let $\cF = \cF_1 \times \dots \times \cF_K$ be a set of bandit models $\bm\nu = (\nu_1,\dots,\nu_K)$ where each $\nu_k$ belongs to the class of distribution $\cF_k$. Let $\cA$ be a strategy satisfying $\cR_{\bm\nu}^{\alpha}(\cA,T) = o(T^\beta)$ for any $\beta>0$ and $\nu \in \cF$.  Then for any $\nu \in \cD$, for any sub-optimal arm $k$, under the strategy $\cA$ it holds that
	\[\lim_{T \rightarrow +\infty} \frac{\bE_{\bm\nu}[N_k(T)]}{\log T} \geq \frac{1}{\cK_{\inf}^{\alpha,\cF_k}(\nu_k, c^\star)},\]
	where $c^\star=\max_{i \in [K]} \emph{CVaR}_{\alpha}(\nu_i)$.
\end{theorem}

Using~\eqref{def::regret}, this result directly yields an asymptotic lower bound on the regret. The proof of Theorem~\ref{thm:LB} follows from a classical change-of-distribution argument, as that of any lower bound proof in the bandit literature. We detail it in Appendix~\ref{app:LB}, following the proof of Theorem 1 in \citet{menard_asymp} originally stated for $\alpha = 1$. We discuss in Appendix~\ref{app::discussion_scaling_lb} how this lower bound yields a weaker regret bound expressed in terms of the CVaR gaps (by Pinsker). 

In the next section we prove that \MCVTS{} matches the lower bound for the set of multinomial distribution when the support is known, and that \CVTS{} matches the lower bound for the set of continuous bounded distribution with a known upper bound. Hence, under these hypotheses, the two algorithms are \textit{asymptotically optimal}. Despite the recent development in CVaR bandits literature, to our knowledge no algorithm has been able to match this lower bound yet. These results are of particular interest because they show that this bound is attainable for CVaR bandit algorithms, at least for bounded distributions.

\subsection{Regret Guarantees for \MCVTS{} and \CVTS{}}

Our main result is the following regret bound for \MCVTS, showing that it is matching the lower bound of Theorem~\ref{thm:LB} for multinomial distributions. 

\begin{theorem}[Asymptotic Optimality of \MCVTS]\label{th::regret_MCVTS}
Let $\bm\nu$ be a bandit model with $K$ arms, where the distribution of each arm $k\in \K$ is multinomial with known support $\cX_k\subset \R^{M_k}$ for some $M_k\in \N$. The regret of \MCVTS{} satisfies
\[\cR_{\bm\nu}(T) \leq \sum_{k: \Delta_k^\alpha>0} \frac{\Delta_k^\alpha \log T}{\cK_{\inf}^{\alpha, \cX_k}(\nu_k, c_1^\alpha)} + o(\log T) \;.\]
\end{theorem}

We then provide a similar result for \CVTS{}, for bounded and continuous distributions with a known upper bound.

\begin{theorem}[Asymptotic Optimality of \CVTS]\label{th::regret_CVTS}
	Let $\bm\nu$ be a bandit model with $K$ arms, where for each arm $k\in\K$ its distribution $\nu_k$ belongs to $\cB_k$, the set of continuous bounded distributions, and its supports $\cX_k$ satisfies $\cX_k \subset [0, B_k]$ for some known $B_k>0$. Then the regret of \CVTS{} on $\bm\nu$ satisfies
	\[\cR_{\bm\nu}(T) \leq \sum_{k: \Delta_k^\alpha>0} \frac{\Delta_k^\alpha \log T}{\cK_{\inf}^{\alpha, \cB_k}(\nu_k, c_1^\alpha)} + o(\log T)\;.\]
\end{theorem}


We postpone the detailed proofs of Theorem~\ref{th::regret_MCVTS} and Theorem~\ref{th::regret_CVTS} respectively to Appendix~\ref{sec::proof_multi} and Appendix~\ref{sec::proof_cvts}, and we highlight their main ingredients in this section. First, using Equation~\eqref{def::regret} it is sufficient to upper bound $\bE[N_k(T)]$ for each sub-optimal arm $k$. To ease the notation we assume that arm 1 is optimal. Our analysis follows the general outline of that of \citet{Honda}, but requires several novel elements that are specific to CVaR bandits. First, the proof leverages some properties of the function $\cK_{\inf}^{\alpha}$ for the sets of distributions we consider. Secondly, it requires novel boundary crossing bounds for Dirichlet distributions that we detail in Section~\ref{sec::technical_tools_main}.

The first step of the analysis is almost identical for the two algorithms and consists in upper bounding the number of selections of a sub-optimal arm by a \textit{post-convergence} term (Post-CV) and a \textit{pre-convergence} term (Pre-CV). The first term controls the probability that a sub-optimal arm \textit{over-performs} when its empirical distribution is ``close" to the true distribution of the arm, while the second term considers the alternative case. To measure how close two distributions are we use the $L^{\infty}$ distance for multinomial distributions, while for general continuous arms we use the Levy distance (See Appendix~\ref{app::notations} for definitions and details). We state the decomposition in Equation~\ref{eq::control_nb_pulls} below for a generic distance $d(F_{k, t}, F_k)$ between the empirical cdf of the arm at a time $t$ and its true cdf. As in Section~\ref{sec::algorithms} we write  $c^{\alpha}_{k,t}$ for the index assigned to arm $k$ by the algorithm at time $t$. Then, for any $\varepsilon_1>0$ and $\epsilon_2>0$ we define the events
\begin{align*}
\cC_{t, k}^+ = &\left\{A_t=k, c_{k, t} \geq c_1^\alpha -\epsilon_1, d(F_{k, t}, F_k) \leq \epsilon_2 \right\} \;, \\
\cC_{t, k}^- = & \left\{A_t=k, c_{k, t} < c_1^\alpha -\epsilon_1\right\} \\ &\cup \left\{A_t=k, d(F_{k, t}, F_k) \geq \epsilon_2 \right\} \;.
\end{align*}

As $\{c_{k, t} \geq c_1^\alpha -\epsilon_1, d(F_{k, t}, F_k) \leq \epsilon_2\} $ is the complementary set of $\{c_{k, t} < c_1^\alpha -\epsilon_1\} \cup \{  d(F_{k, t}, F_k) > \epsilon_2 \}$ we obtain 
\begin{eqnarray}\label{eq::control_nb_pulls}
&\bE[N_k(T)] \leq \underbrace{\bE\!\left[\sum_{t=1}^T \ind(\cC_{t, k}^+)\right]}_{\text{(Post-CV)}} + \underbrace{\bE\!\left[\sum_{t=1}^T \ind(\cC_{t, k}^-)\right]}_{\text{(Pre-CV)}} \;.
\end{eqnarray}

For an arm $k$ satisfying the hypothesis of Theorem~\ref{th::regret_MCVTS}, for all $\epsilon>0$ we show that the corresponding Post-Convergence term of \MCVTS{} satisfies 
\begin{eqnarray}\label{eq::post_cv_mcvts}
\text{(Post-CV)} \leq \frac{(1+\epsilon) \log T}{\cK_{\inf}^{\alpha,\cX_k}(\nu_k, c_1^\alpha)} + \cO(1)\;,
\end{eqnarray}

while for an arm $k$ satisfying the hypothesis of Theorem~\ref{th::regret_CVTS}, for all $\epsilon>0$ the corresponding Post-Convergence term of \CVTS{} satisfies 
\begin{eqnarray}\label{eq::post_cv_cvts}
\text{(Post-CV)} \leq \frac{\log T}{\cK_{\inf}^{\alpha,\cB_k}(\nu_k, c_1^\alpha)-\epsilon} + \cO(1)\;. 
\end{eqnarray}
%
%

Finally, for both algorithms the Pre-Convergence term is asymptotically negligible for the families of distribution they cover, namely 
\begin{eqnarray}\label{eq::pre_cv}\text{(Pre-CV)} = \cO(1) \;.\end{eqnarray}

We detail these results in Appendix~\ref{sec::proof_multi} and Appendix~\ref{sec::proof_cvts}. In the next section we present some novel technical tools that we introduced in order to prove these results.
 
\subsection{Technical challenges and tools}\label{sec::technical_tools_main}

The proofs of \eqref{eq::post_cv_mcvts}, \eqref{eq::post_cv_cvts} and \eqref{eq::pre_cv} follow the outline of \citet{Honda}, respectively for Multinomial Thompson Sampling and Non Parametric Thompson Sampling. However, replacing the linear expectation by the CVaR that is non-linear, causes several technical challenges that make the adaptation non-trivial. This is particularly true for the \textit{boundary crossing probabilities} for Dirichlet random variables, that we define and analyze in this section. Our results aim at replacing the Lemma 13, 14, 15 and 17 of \citet{Honda} in the proofs of Theorem~\ref{th::regret_MCVTS} and Theorem~\ref{th::regret_CVTS}.

\paragraph{Boundary crossing probabilities} In this paragraph we highlight the construction of \textit{boundary crossing} probabilities for Dirichlet random variables, which consists in providing upper and lower bounds of some terms of the form
\[\bP_{w \sim \mathrm{Dir}(\beta)}\left(C_{\alpha}(\cX, w) \geq c \right),\]
for some known support $\cX=(x_1, \dots, x_n)$, parameter $\beta \in \R_+^{n}$ of the Dirichlet distribution, and some real value $c$ that will be defined in context. We introduce the set
\begin{eqnarray*}
\Sa &=& \left\{p \in \cP^{n}: C_\alpha(\cX,p) \geq c\right\}\,,
\end{eqnarray*}
following the notations of Section~\ref{sec::algorithms} for $C_\alpha(\cX, p)$.
Thanks to the expression of the CVaR in Equation~\eqref{eq::cvar_def} we have \begin{eqnarray}\label{eq::S}\Sa= \cup_{m=1}^n \Sm \;,\end{eqnarray} where we defined for all $m \in \{1,\dots,n\}$ the sets
\[\Sm \!=\left\{p \in \cP^{n},  x_m \!-\frac{1}{\alpha}\sum_{i=1}^{n} p_i \left(x_m-x_i\right)^+ \geq c \right\}.\]
This set is closed and convex, hence $\Sa$ is closed, and is the finite union of convex sets (but is not convex). These properties are crucial to prove the results of this section.

\paragraph{Bounded support size} We first study the case when the size of the support is $|\cX|=M$, for some known $M \in \N$ and  when the considered distributions are the \textit{frequency} of each observation in $\cX$ out of $n \in \N$ many observations, which we represent by the set
\[\mathcal{Q}_n^M= \left\{(\beta, p) \in \N^{*n} \times \cP^M: p=\frac{\beta}{n} \right\}\;. \]
We then express bounds for boundary crossing probabilities on this set, in terms of $n$ and $M$, where $n$ should be considered much larger than $M$. Lemma~\ref{lem::lemma13_main} and \ref{lem::lemma14_main} respectively provide an upper and lower bound on such probabilities.

\begin{restatable}[Upper Bound]{lemma}{lemupM}
	\label{lem::lemma13_main}
	For any $(\beta, p) \in \mathcal{Q}_n^M$, for any $c>C_\alpha(\cX, p)$, it holds that  
	\begin{eqnarray*}
	\bP_{w \sim Dir(\beta)}(w \!\in\! \Sa)\!\!\!\! &\leq&\!\!\!  C_1 M n^{M/2}\exp(-n \kinfcvx(p, c)) \;,
	\end{eqnarray*}
for some constant $C_1$.
\end{restatable}
\begin{restatable}[Lower Bound]{lemma}{lemlowM}
	\label{lem::lemma14_main}
	For any $(M, n)\in N^2$ and $(\beta, p) \in \mathcal{Q}_{n}^M$, let $p^\star \in \cP^M$ be any vector satisfying $C_\alpha(\cX, p^\star)\geq c$. Denoting $P_\beta=\bP_{w \sim \mathrm{Dir}(\beta)}\left(w \in \Sa \right)$, it holds that
	\begin{align*}
	 P_\beta & \geq  \frac{n!}{\prod_{i=1}^M\beta_i!}\frac{\beta_M}{n p_M^\star}\prod_{j=1}^M(p_j^\star)^{\beta_j}\\
	&\geq \frac{1}{n}\bP_{\mathrm{Mult}\left(n, p\right)}(\beta)\times e^{-n\kinfcvx\left(p, c\right)}\\
	& \geq C_M \frac{\exp\left(-n \kinfcvx(p, c)\right)}{n^{\frac{M+1}{2}}} \;,
	\end{align*}	
	where $C_M=\sqrt{2\pi}\left(\frac{M}{2.13}\right)^\frac{M}{2}$, and $\bP_{\mathrm{Mult}\left(n, q\right)}(\beta)$ denotes the probability that a vector drawn from a multinomial distribution with $n$ trials and probability $q$ is equal to $\beta$.
\end{restatable}

The detailed proofs of these two results are to be found in Appendix~\ref{app::auxiliary_results}. Lemma~\ref{lem::lemma13_main} hinges on the Lemma 13 of \citet{Honda} (see Appendix~\ref{app::auxiliary_results}), while the proof of Lemma~\ref{lem::lemma14_main} shares the core idea of the proof sketch of their Lemma 14. For both results we exploit the convexity of the sets $\Sm$ (equation~\eqref{eq::S}). Both results are core component of the proof of both \MCVTS{} and \CVTS{} due to the quantization arguments used in the latter. Furthermore, we alternatively use the three results provided in~\ref{lem::lemma14_main}.

\paragraph{General support size} We now detail some results that are specifically designed for the regret analysis of \CVTS. For this reason, we consider a support $\cX=(x_1, \dots, x_n)$ and the Dirichlet distribution $\cD_n$ defined in Section~\ref{sec::algorithms}. 
Here we focus on the Dirichlet sample, hence the support $\cX$ is known. We further denote $u_\cX$ the uniform distribution on $\cX$, and $C_\alpha(\cX)$ its CVaR. We first establish an upper bound.

\begin{restatable}{lemma}{lemupC}
	\label{lem::lem_15_main} Let $\cX=(x_0, \dots, x_{n}) \subset [0, B]^{n+1}$ for some known $B>0$ and $n\in \N$, assuming that $x_0=B$. For any $c > C_\alpha(\cX)$, and any $\eta>0$ small enough it holds that 
\[ \bP_{w \sim \cD_n}(C_\alpha(\cX, w) \geq c) \leq \frac{B}{\eta} \exp^{-n \left(\kinfcvgen(u_\cX,c) - \eta C(B, \alpha, c)  \right)}\;,\]
for some constant $C(B, \alpha, c)$.
\end{restatable} 

We prove this result in Appendix~\ref{app::auxiliary_results}. It relies on deriving the dual form of the functional $\kinfcvgen$ for discrete distributions, that is a result of independent interest.

\begin{lemma}\label{lem::kinf_dual_main}
	If a discrete distribution $F$ supported on $\cX$ satisfies $\bE_F\left[\frac{(y-c)\alpha}{(y-X)^+}\right]<1$, then for any $c > \CVAR_{\alpha}(F)$ it holds that
	\[\kinfcvgen(F, c) = \inf_{y \in \cX} \max_{\lambda \in \left[0, \frac{1}{\alpha(y-c)} \right)} g(y, \lambda, X) \;, \]
	with $g(y, \lambda, X) = \bE_F\left[\log(1-\lambda((y-c)\alpha)-(y-X)^+)\right]$.
	
	If $\bE_F\left[\frac{(y-c)\alpha}{(y-X)^+}\right] \geq 1$, then for any $c > \CVAR_{\alpha}(F)$ 
	\[\kinfcvgen(F, c) = \inf_{y \in \cX} \bE_F\left(\frac{(y-X)^+}{(y-c)\alpha}\right)\;.\]
\end{lemma}

The detailed proof of this result is provided in Appendix~\ref{app::asymp_opt}, where we also show that this expression matches the result of \citet{HondaTakemura10} for $\alpha=1$, and is similar to the one obtained by \cite{agrawal2020optimal}[Theorem 6] for a more complex set of distributions (which is hence less explicit). Furthermore, \citet{agrawal2020optimal}[Lemma 4] prove the continuity of $\kinfcvx$ under this condition, which is required in several part of our proofs. We propose a simplified proof of this result for the restriction to bounded distribution in Appendix~\ref{app::asymp_opt}.

The last result we report in this section is a lower bound on the probability that a \textit{noisy} CVaR in \CVTS{} exceeds the CVaR of the empirical distribution. 

\begin{restatable}{lemma}{lemlowC}
	\label{lem::lower_lem17_main}
Assume that $\cX=(x_1, \dots, x_n)$ and $x_1<\dots<x_n$, then $x_{\lceil n \alpha \rceil}$ is the empirical $\alpha$ quantile of the set and $x_1$ its minimum, and it holds that
\[\bP_{w \sim \cD_n}\left(C_\alpha(\cX, w) \geq C_\alpha(\cX)\right) \geq \frac{1}{25n^3}(x_{\left\lceil n \alpha \right\rceil}-x_1) \;.\]
\end{restatable}
This result is proved in Appendix~\ref{app::auxiliary_results}. Let us remark that in all the results presented in this section we consider a fixed support $\cX$, while in \CVTS{} the support is random and evolves with the time. This causes several challenges in the proof. In particular, the use of Lemma~\ref{lem::lower_lem17_main} in Appendix~\ref{app::A2} is not sufficient in itself to conclude and additional work is required to handle the random support.

\begin{remark}
The results presented in this section contains most of the difficulty induced by the replacement of the expectation by the CVaR in the proofs. Extending these results to other criterion is an interesting future work and may help generalize the Non Parametric Thompson Sampling algorithms to broader settings.
\end{remark}
\section{Experiments}\label{sec::experiments}

In this section we report the results of experiments on the algorithms presented in the previous sections, first on synthetic examples, and then on a use-case study in agriculture based on the DSSAT agriculture simulator.

\subsection{Preliminary Experiments}\label{sec::other_exp}

We first performed various experiments on synthetic data in order to check the good practical performance of \MCVTS{} and \CVTS{} on settings that are simple to implement and are good illustrative examples of the performance of the algorithms. Due to space limitation, we report a complete description of the experiments and and an analysis of the results in Appendix~\ref{app::experiments}. We tested the TS algorithms on specified difficult instances and on randomly generated problems, against U-UCB and CVaR-UCB. 

As an example of experiment with multinomial arms, we report in Table~\ref{tab::xp_mult_bayes_main} the results of an experiment with $10^3$ randomly generated problems with $5$ arms drawn uniformly at random in $\cP^{|\cX|}$, where $\cX=[0, 0.1, 0.2, \dots, 1]$, for $\alpha\in \{10\%, 50\%, 90\%\}$ and an horizon $10^4$. These experiments confirm the benefits of TS over UCB approaches, as M-CVTS significantly outperforms its competitors for all levels of the parameter $\alpha$. We also tested the algorithms with fixed instances (see Tables~\ref{tab::xp1_multapp}-\ref{tab::xp4_mult}), with the same results, and further illustrated the asymptotic optimality of \MCVTS{} in Figures~\ref{fig::xp1_mult10_lb} and ~\ref{fig::xp1_mult90_lb} by representing the lower bound presented in Section~\ref{sec::analysis} along with the regret of the algorithm in logarithmic scale. 

We also tested \CVTS{} on different problems, using truncated gaussian mixtures (TGM). The results are presented in Tables~\ref{fig::xp1_BBG}-\ref{fig::xp4_TBG}, and again show the merits of the TS approach. We also performed an experiment with a small level $\alpha=1\%$ (Table~\ref{tab::xp_small_alpha}) and show that \CVTS{} keeps the same level of performance in this case, while the other algorithm stay in the linear regime for the horizon we consider. Finally, we also experimented more arms ($K=30$) and randomly generated TGM problems and report the results in Table~\ref{tab::xp_MTG_main}. The means and variance of each arm satisfy $(\mu_k, \sigma_k) \sim \cU([0.25, 1]^10\times [0, 0.1]^10)$, and the probabilities of each mode are drawn uniformly, $p_k\sim \cD_{19}$.

\begin{table}[h]
	\caption{CVaR regret at time $T=10^4$, averaged over $10^3$ random instances with $5$ multinomial arms supported on $\cX=[0.1, 0.2, \dots, 1]$}
	\label{tab::xp_mult_bayes_main}
	\vskip 0.15in
	\begin{center}
		\begin{small}
			\begin{sc}
			\begin{tabular}{lccc}
		\textbf{$\alpha$}  &U-UCB & CVaR-UCB & \MCVTS \\
		\hline \\
		$10\%$  & 633.1& 219.7 &\textbf{38.8} \\
		$50\%$  & 368.8 & 187.9 & \textbf{48.9} \\
		$90\%$  & 188.5 & 186.2 & \textbf{42.7}\\
			\end{tabular}
			\end{sc}
		\end{small}
	\end{center}
	\vskip -0.1in
\end{table}

\begin{table}[h]
	\centering
	\caption{Results for TGM arms with 10 modes, at $T=10000$ averaged over 400 random instances with $K=30$, $\alpha=5\%$ (results: mean (std)).} \label{tab::xp_MTG_main}
	\begin{center}
		\begin{tabular}{llll}
			T  &U-UCB & CVaR-UCB & \CVTS \\
			\hline \\
			10000  & 2149.9 (263)& 2016.0 (265)& \textbf{210.9 (6.4)}  \\
			20000  & 4276.4 (538)& 3781.3 (521) & \textbf{237.1 (15.4)} \\
			40000  & 8493.4 (1085)& 6894.1 (985) & \textbf{263.5 (17.9)}\\
		\end{tabular}
	\end{center}
\end{table}

These very good results with synthetic data and its theoretical guarantees motivate using the \CVTS{} algorithm in the real-world application we introduce in the next section.  
\subsection{Bandit application in Agriculture}\label{sec::agriculture}

\paragraph{Motivation} Let us consider a farmer who must decide on a \textit{planting date} (action) for a rainfed crop. Farmers have been reported to primarily seek advice that reduces uncertainty in highly uncertain decision making \cite{mccown2002changing, hochman2011emerging, evans2017data}. 
Planting date is an example of such a decision as it will influence the probabilities of favorable meteorologic events during crop cultivation. These events are highly uncertain due to the length of crop growing cycles (e.g. 3 to 6 months for grain maize). For instance, because of the stochastic nature of the rainfalls and temperatures, a farmer will observe a range of different crop yields from year to year for the same planting date, all other technical choices being equal. Thus, assuming that the environment is stationary, each planting date corresponds to an underlying, unknown yield distribution, which can be modeled as an arm in a \textit{bandit problem}. 
Depending on her profile, a farmer may be more or less risk averse, and the \textit{Conditional Value at Risk} can be used to personalize her level of risk-aversion. For instance, a small-holder farmer looking for food security may seek to avoid very poor yields compromising auto-consumption (e.g $\alpha\leq 20\%$), while a market-oriented farmer may be more prone to risky choices in order to increase her profit but still not risk neutral (e.g $\alpha=80\%$). Yield distributions are supposed to be \textit{bounded}. Indeed, a finite yield potential can be defined under non-stressing conditions for a given crop and environment \cite{evans1999yield, tollenaar2002yield}. Observed yields can be modeled as following Von Liebig's law of minimum \cite{paris1992return}: limiting factors will determine how much of the yield potential can be expressed.

\begin{figure}[H]
	\centering
	\includegraphics[width=0.9\columnwidth]{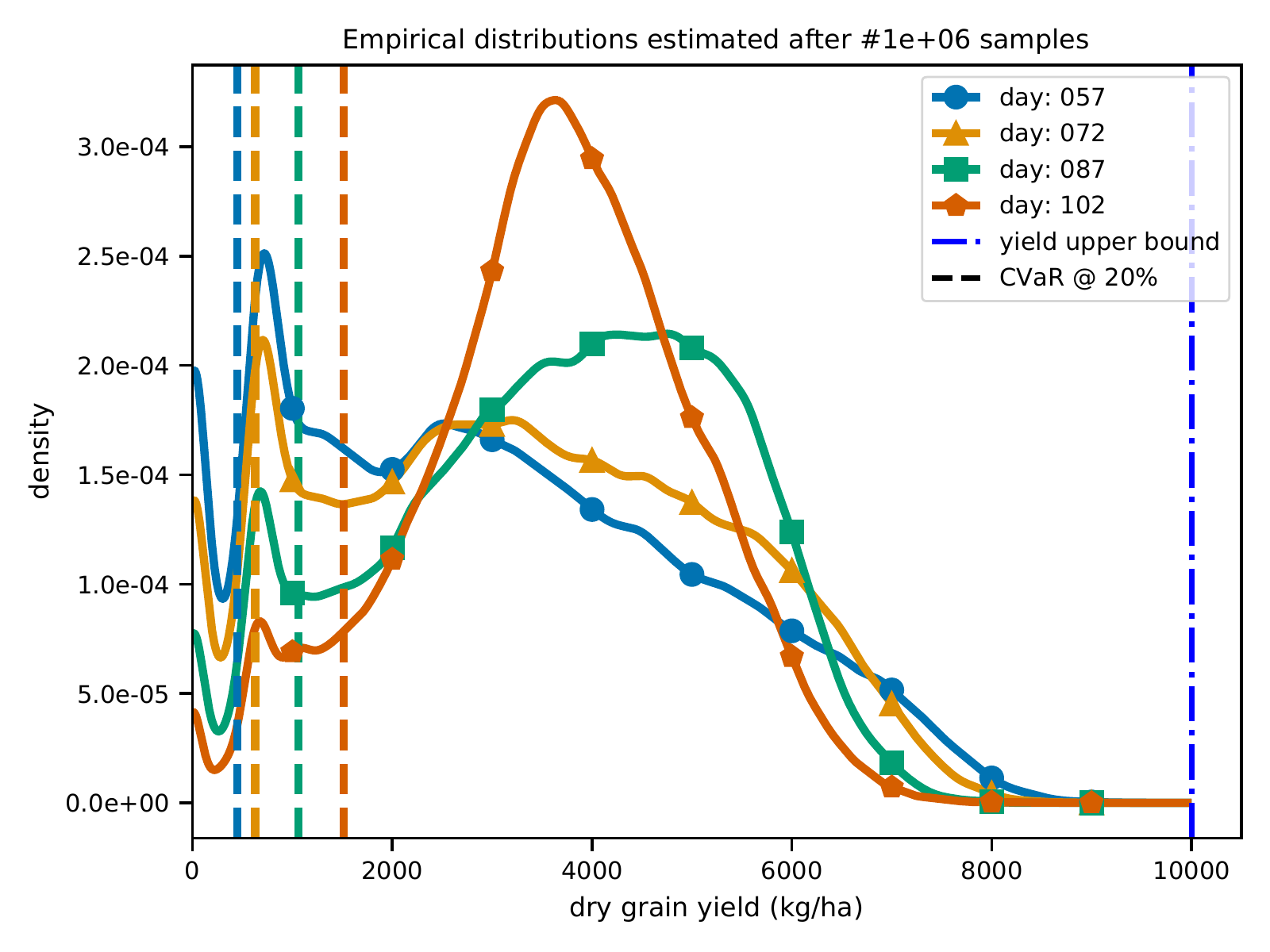}
	\vspace{-4mm}
	\caption{Empirical simulated yields and respective CVaRs at 20\% estimated after $10^6$ samples in \texttt{DSSAT} environment.}
	\label{fig::dssatDists}
\end{figure}

\paragraph{Setting} Planting date decision-making support requires extensive testing prior to any real-life application, due the potential impact of wrong action-making, particularly in subsistence farming. For this reason, we consider the problem of facing many times the decision of a planting date in the \href{https://dssat.net/}{\texttt{DSSAT}}\footnote{\texttt{DSSAT} is an Open-Source project maintained by the DSSAT Foundation, see \href{https://dssat.net/}{https://dssat.net/}.} simulator, to make an \textit{in silico} decision. \texttt{DSSAT}, standing for \emph{Decision Support System for Agrotechnology Transfer}, is a world-wide crop simulator, supporting 42 different crops, with more than 30 years of development \cite{hoogenboom2019dssat}. We specifically address maize planting date decision, as maize is a crucial crop for global food security \cite{shiferaw2011crops}. Each simulation is assumed to be realistic, and starts from the same field initial conditions as ground measured. The simulator takes as input historical weather data, field soil measures, crop specific genetic parameters and a given crop management plan. Modeling is based on simulations of atmospheric, soil and plants compartments and their interactions.  In the considered experiments, after a decision is made on planting date in the simulator, daily stochastic meteorologic features are generated according to historical data \cite{richardson1984wgen} and injected in the complex crop model. At the end of crop cycle, a maize grain yield is measured to evaluate decision-making. We parameterized the crop-model under challenging rainfed conditions on shallow sandy soils, i.e. with poor water retention and fertility. Such experiment intends to be representative of realistic conditions faced by small-holder farmers under heavy environmental constraints, such as in Sub-Saharan Africa. Thus, this setting can help picturing how CVaR bandits may perform in real-world conditions. 
For the sake of the experiments, we built a bandit-oriented Python wrapper to \texttt{DSSAT} that we made \href{https://github.com/rgautron/DssatBanditEnv}{available}\footnote{ \href{https://github.com/rgautron/DssatBanditEnv}{https://github.com/rgautron/DssatBanditEnv}} to the bandit community for reproducibility.

\paragraph{Experiments} We test bandit performances on the 4 armed \texttt{DSSAT} environment described in Table~\ref{tab::dssatDists}. To illustrate the non-parametric nature of these distributions, we report in Figure~\ref{fig::dssatDists} estimations of their density obtained with Monte-Carlo simulations, as well as of their CVaRs. The resulting distributions are typically \emph{multi-modal}, with one of their mode very close to zero (years of bad harvest), and with upper tails that cannot be properly characterized. However the practitioner can realistically assume that the distributions are upper-bounded, due to the physical constraints of crop-farming. The yield upper-bound is set to 10 t/ha thanks to expert knowledge for the considered conditions.

\begin{table}[!hbtp]
	\caption{Empirical yield distribution metrics in kg/ha estimated after $10^6$ samples in \texttt{DSSAT} environment} \label{tab::dssatDists}
	\vskip -0.1in
	\begin{center}
		\begin{tabular}{lllll}
			\makecell[t]{day (action)\hspace{-10mm}.}  & & & $\text{CVaR}_{\alpha}$ & \\ \hline
			& $5\%$ & $20\%$ & $80\%$ & \makecell[t]{$100\%$ (mean)} \\ \cmidrule{2-5}
			057  & 0 & 448 & 2238 & 3016\\
			072  & 46 & 627 & 2570 & 3273\\
			087  & 287 & 1059 & 3074 & \textbf{3629}\\
			102  & \textbf{538} & \textbf{1515} & \textbf{3120} & 3586\\
		\end{tabular}
	\end{center}
\end{table}

The presented \texttt{DSSAT} environment advocates for the use of algorithms specifically designed for CVaR bandits, as the optimal arm can change depending on the value of the parameter $\alpha$. Our experiment consists in running 64 trajectories for three algorithms U-UCB, CVaR-UCB and \CVTS{} defined in Section~\ref{sec::algorithms}.  Experiments are carried out with an horizon of $10^4$ time steps, and we compare the results for each algorithm for $\alpha \in \{5\%, 20\%, 80\%\}$ to see how the parameter impacts their performance. Indeed we want a strategy to perform well on all $\alpha$ choices, allowing to freely model any farmer's risk aversion level. As shown in Figure~\ref{fig::dssatE1RegretPlot} and Table~\ref{tab::dssatE1RegretTab}, \CVTS{} appears to be consistently better than its UCB counterparts in \texttt{DSSAT} environment for all tested $\alpha$ values, which is encouraging for real-life applications. 

\begin{table}[!hbtp]
	\caption{Empirical yield regrets at horizon $10^4$ in t/ha in \texttt{DSSAT} environment, for 1040 replications. Standard deviations in parenthesis.} \label{tab::dssatE1RegretTab}
	\vskip -0.1in
	\begin{center}
		\begin{tabular}{llll}
			\textbf{$\alpha$}  &U-UCB & CVaR-UCB & \CVTS \\
			\hline 
			$5\%$  & 3128 (3)& 760 (14)& \textbf{192 (11)} \\
			$20\%$  & 4867 (11)& 1024 (17) & \textbf{202 (10)} \\
			$80\%$  & 1411 (13)& 888 (13) & \textbf{287 (12)}\\
		\end{tabular}
	\end{center}
	\vskip -0.1in
\end{table}

Further experiments are reported in Appendix~\ref{app::experiments}. In particular we increase the number of arms, and empirically study the effect of over-estimating the support upper-bound: our results show that a "prudent" bound has little effect of the performance of the algorithms in the settings we consider. This property is of particular interest for the practitioner, as a proper tuning of the support upper bound is the main limitation of the use of \CVTS{} (and all bandit algorithms available for this problem). In most applications grounded on physical reality, the availability of such prudent upper-bound estimate is likely, and sufficient to ensure the practical performance of the \CVTS{} algorithm.

{\setlength\intextsep{0pt}
	\begin{figure}[H] 
		\begin{figure}[H]
			\centering
			\label{fig::algoComp5p}
			\includegraphics[width=\columnwidth]{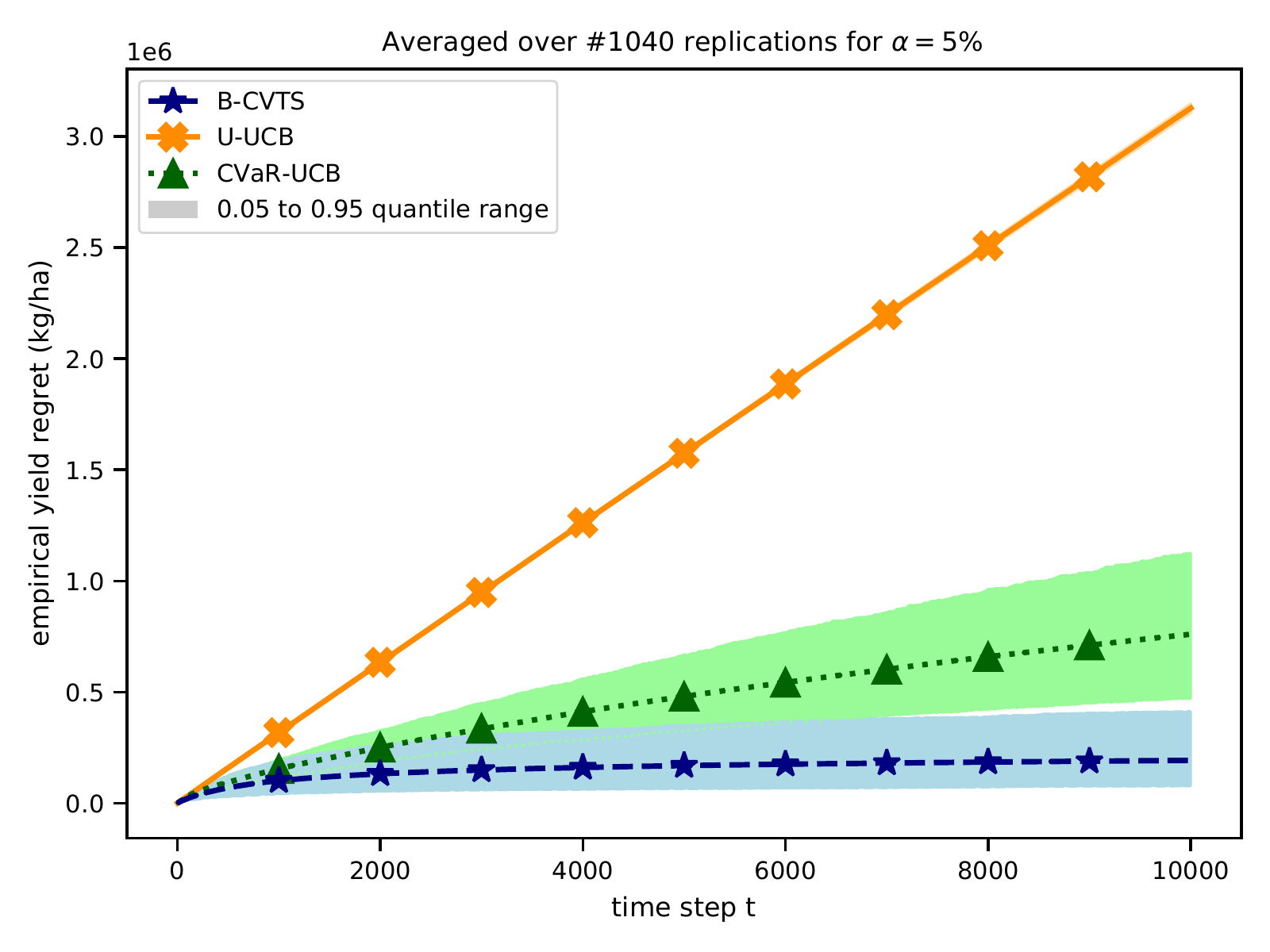}
			\vspace*{-2.5em}
			\caption*{$\alpha=5\%$}	
		\end{figure}
		\begin{figure}[H]
			\centering
			\label{fig::algoComp80p}
			\vspace*{-.2em}
			\includegraphics[width=\columnwidth]{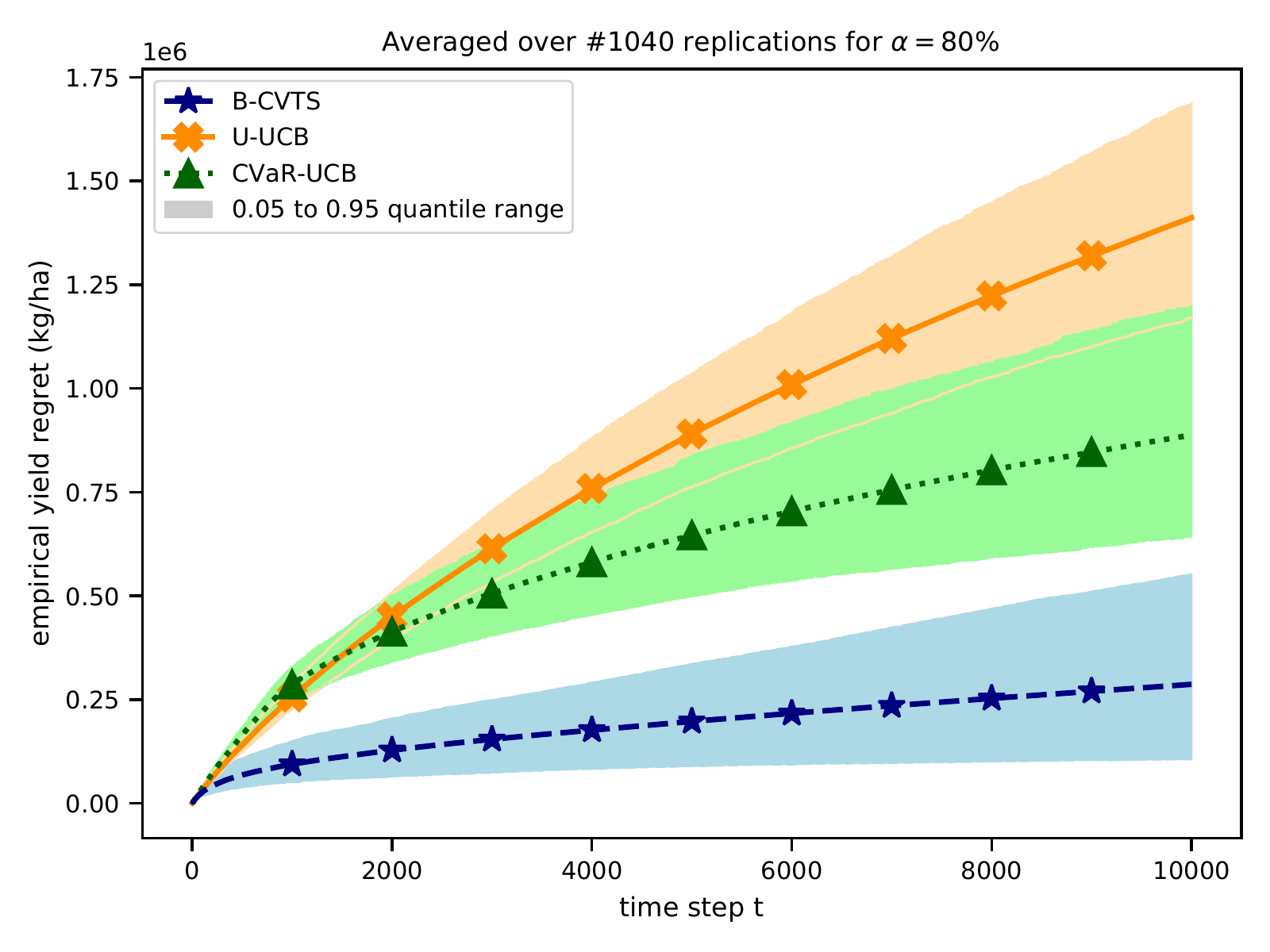}
			\vspace*{-2.5em}
			\caption*{$\alpha=80\%$}	
		\end{figure}
		\vspace{-3mm}
		\caption{Regret comparison in \texttt{DSSAT} environment, averaged over 1040 experiment replications.}
		\label{fig::dssatE1RegretPlot}
	\end{figure}
}

\paragraph{Perspectives} This first set of experiments using a challenging realistic crop simulator is promising, and motivates to further investigate the use of \CVTS{} algorithm for crop-management support and other problems that can be modeled as CVaR bandits. \CVTS{} enjoys appealing theoretical guarantees, and thanks to its simplicity and competitive empirical performances may be a good candidate for practitioners. In order to  address real-world crop-management challenges, many questions remain to be considered, e.g.  how to optimally generate mini-batches of recommendations to an ensemble of farmers in a semi-sequential procedure (in order to account for the long feedback time), how to incorporate distribution priors on crop-management options that could be pre-learnt \textit{in silico} and refining them adaptively in the real world (thus, minimizing random exploration in the real world), how to include contextual information such as soil characteristics and local weather forecasts, or how handle non-stationarity, incorporating climate change progressive impact on an optimal planting date. Furthermore, the simplicity of the Non-Parametric Thompson Sampling algorithms make them appealing for generalization to other risk-aware settings, e.g risk-constrained (maximizing the mean under a condition on the CVaR) or with other risk metrics (mean-variance, entropic risk, etc). All of these open questions make interesting challenges for future works.
\section*{Acknowledgements}
This work has been supported by the French Ministry of Higher Education and Research, Hauts-de-France region, Inria within the team-project Scool
and the MEL. The authors acknowledge the funding of the French National Research Agency under projects BADASS (ANR-16-CE40-0002) and BOLD (ANR-19-CE23-0026-04) and the I-Site ULNE regarding project R-PILOTE-19-004-APPRENF.
 
The PhD of Dorian Baudry is funded by a CNRS80 grant.
The PhD of Romain Gautron is co-funded by the French Agricultural Research Centre for International Development (CIRAD) and the Consortium of International Agricultural Research Centers (CGIAR) Big Data Platfrom.
	
Experiments presented in this paper were carried out using the Grid'5000 testbed, supported by a scientific interest group hosted by Inria and including CNRS, RENATER and several Universities as well as other organizations (see \hyperlink{https://www.grid5000.fr}{https://www.grid5000.fr}).

\bibliographystyle{abbrvnat}
\bibliography{biblio}

\appendix
\onecolumn

\section{Notations for the proofs}\label{app::notations}
In this section, we introduce for convenience several notations that are used in the regret analysis.

\subsection{General Notations}

\paragraph{Model for multinomial arms} In section~\ref{sec::proof_multi} we denote by $(\cX_1, \dots, \cX_K)$ the supports of $(\nu_1, \dots, \nu_K)$, where $\cX_k=(x_1^k, \dots, x_{M_k}^k)$ for some integer $M_k$. We also denote $\cP^{M}=\{p \in \R^{M}: \forall i, p_i\geq 0, \sum_{j=1}^{M}p_j=1\}$ the probability simplex in $\R^{M+1}$. Any multinomial distribution $\nu$ is characterized by its support and its probability vector $p_\nu$: $\nu=(\cX, p_\nu)$ with $p_\nu \in \cP^{|\cX|}$. Each arm $\nu_k$ is associated with a probability vector $p_k \in \cP^{M}$. 

\paragraph{Model for continuous bounded distributions} In section~\ref{sec::proof_cvts} we assume that each arm $k$ is supported in $[0, B_k]$ for some known value $B_k>0$. We could assume supports of the form $[a_k, b_k]$ where only an upper bound on each $b_k$ is known without loss of generality in the result, but we use this formulation for the sake of simplicity. We consider the set of continuous distributions in $[0, B_k]$, that we write $\cC^{B_k}$. We still denote $\cP^{M}=\{p \in \R^{M}: \forall i, p_i\geq 0, \sum_{j=1}^{M}p_j=1\}$ the probability simplex in $\R^{M}$. The distribution of an arm $k$ is $\nu_k$, and its CVaR at a level $\alpha$ is denoted $c_{k}^\alpha$. 

\paragraph{Notations for the CVaR} In the proofs in section of section~\ref{sec::proof_multi} and section~\ref{sec::proof_cvts} we will encounter three different CVaR formulations for which we propose convenient notations,

\begin{itemize}
	\item $\CVAR_\alpha(F)$ for the CVaR of a distribution with a specified cumulative distribution function $F$.
	\item $C_\alpha(\cX, w)$ for the CVaR of a discrete random variable of support $\cX \subset [0, B]$ associated with a probability vector $w \in \cP^{|\cX|}$. According to our previous notation, $C_\alpha(\cX, w)=\CVAR_\alpha\left(F_w\right)$, where for all $y \in \R$, \[F(y) = \frac{1}{|\cX|} \sum_{x \in \cX} w_x \ind(y \leq x)\;.\]
	\item $C_\alpha(\cX)$ the CVaR of the uniform distribution on the discret support $\cX$, which shortens the notations for the CVaR of the empirical distributions: $C_\alpha(\cX)= C_\alpha(\cX, 1_N)$ where $1_N=\underbrace{\left(\frac{1}{N}, \dots, \frac{1}{N}\right)}_{N terms}$.
\end{itemize}

We also introduce $c_k^\alpha = \CVAR_\alpha(F_k)$ the CVaR of each arm's distribution, further assuming without loss of generality that 
\[c_1^\alpha = \argmax{k \in \K} c_k^\alpha.\]

\paragraph{Algorithm} Both \MCVTS{} and \CVTS{} can be formulated as an index policy, that is 
\[A_t = \argmax{k \in [K]} \ c_{k,t}^{\alpha},\]
where $c_{k,t}^{\alpha}$ is the index used in round $t$. We recall that $N_k(t)=\sum_{s=1}^{t}\ind(A_s=k)$ denotes the number of selections of arms $k$. We define the $\sigma$-field $\cF_t = \{A_{\tau}, r_{A_\tau} \text{ for } \tau=\{1, \dots, t\}\}$. In particular, knowing $\cF_t$ allows to know $N_k(t)$ and the history of all arms available up to time $t$ (i.e. all observations drawn before and including $t$).

\paragraph{Distances} We define for multinomial distributions with same support $\cX$ the distance \[d: \cP^{|\cX|} \times \cP^{|\cX|} \rightarrow \R^+ : (p, q) \rightarrow ||p-q||_\infty=\sup_{m \in \{1,\dots, |\cX|\}} |p_m-q_m| \;.\]

We also introduce the notation $D_L(F, G)$ for the Levy distance between two distributions with cdf $F$, and $G$. Namely, if two distributions are supported in $[0, B]$ for some $B>0$

\[D_L(F, G) = \inf\left\{\epsilon>0: \forall x \in [0, B], F(x-\epsilon)-\epsilon \leq G(x) \leq F(x+\epsilon) + \epsilon   \right\} \;. \]

Furthermore, we recall the result from \cite{Honda} stating that for two distributions of cdf $F$ and $G$,
 \[D_L(F, G) \leq ||F-G ||_\infty \]

\subsection{Specific notations for multinomial arms}

We introduce for multinomial arms $N_k^m(t)$ as the number of times an element $x_m$ has been observed during these pulls, so that $N_k(t)=\sum_{m=1}^{M_k} N_k^m(t)$. The Dirichlet posterior distribution given the observation after $t$ rounds is then $\mathrm{Dir}(\beta_{k, t})$ where $\beta_{k, t} = (1+N_k^1(t-1), 1+N_k^1(t-1), \dots, 1+N_k^{M_k}(t-1))$. With this notation, the index is 
\[c_{k,t}^{\alpha} = C_\alpha\left(\cX_k, w_{k, t} \right)
\ \ \text{where} \ \ \ w_{k, t} \sim \mathrm{Dir}(\beta_{k, t}).\]

We denote by $\betaDir$ the parameter of the Dirichlet distribution sampled at round $t$ for arm $k$, and by $\MDir \in \cP^{M_k}$ the mean of this Dirichlet distribution: $\MDir = \frac{1}{N_k(t-1) + M_k} \beta_{k, t}$. Observe that this probability vector can also be viewed as a biased version of the empirical probability distribution of arm $k$. 

To ease the notation, we denote by $\cX=(x_1, \dots, x_M)$ the support of arm $1$, while the support of any sub-optimal arm $k$ is denoted by $\cX_k=(x_1^k, \dots, x_{M_k}^k)$.

\subsection{Notations for continuous arms}

For continuous arms we simply write $\cX_n^k$ the history of observations available after $n$ pulls of arm$k$,  $X_0^k, X_1^k, X_2^k, \dots, X_{n}^k$ in the order they have been collected, including as first term  $X_0^k=B_k$, the upper bound of the support of $k$ considered by the strategy.
When considering (optimal) arm $1$ we omit the exponent $k$ to simplify notations. The Dirichlet distributions that we consider in this case  are always of the form $\text{Dir}((1, \dots , 1))$,  where the size of the mean vector depends on the number of observation. We recall that this distribution is the uniform distribution on a simplex of fixed size $N$, and its average is the vector $\left(\frac{1}{N}, \dots, \frac{1}{N}\right)$. We denote this distribution $\cD_N$ to simplify the notations. 

\section{Proof of Theorem~\ref{th::regret_MCVTS} : analysis of \MCVTS{} with multinomial arms}\label{sec::proof_multi}

Thanks to Equation~\eqref{eq::control_nb_pulls} presented in Section~\ref{sec::analysis}, the proof of Theorem~\ref{th::regret_MCVTS} can be obtained by proving Equation~\eqref{eq::post_cv_mcvts} and Equation~\eqref{eq::pre_cv} for multinomial arms distributions. This consists in upper bounding the \textit{pre-convergence} (Pre-CV) and \textit{post-convergence} (Post-CV) terms presented in section~\ref{sec::analysis}. In this section we use the $L^{\infty}$ distance presented in Section~\ref{app::notations}.

\subsection{Proof of Equation~\eqref{eq::post_cv_mcvts} : Upper Bound on the Post-Convergence term}\label{sec::postcv_multi}


We upper bound the term $\text{(Post-CV)} = \bE\left[\sum_{t=1}^T \ind(A_t=k, c_{k, t}^\alpha \geq c_1^\alpha-\epsilon_1, d(\MDir, p_k) \leq \epsilon_2)\right]$, where $p_{k, t}=\frac{\beta_{k, t}}{N_k(t-1)}$ is the probability vector associated with the empirical distribution of an arm $k$, i.e the frequency of each item of the support $\cX_k$ in the history of arm $k$. We recall that this former quantity is biased because of the initialization step which set $\beta_{k, 0}=(1, \dots, 1)$ and $N_k^i(0)=1$ (introducing the fictitious time $t=0$ just before the algorithms starts). For any constant $n_0(T)$ we have
\begin{align*}
\text{(Post-CV)} \leq& \bE\left[\sum_{t=1}^T \ind(A_t=k, c_{k, t}^\alpha \geq c_1^\alpha-\epsilon_1, d(\MDir, p_k)\leq\epsilon_2)\right] \\
\leq & \sum_{t=1}^T \bE\left[\ind(A_t=k, N_k(t-1)\leq n_0(T), c_{k, t}^\alpha\geq c_1^\alpha-\epsilon_1, d(\MDir, p_k)\leq \epsilon_2) \right]\\
&+ \sum_{t=1}^T \bE\left[\ind(A_t=k, N_k(t-1)\geq n_0(T), c_{k, t}^\alpha\geq c_1^\alpha-\epsilon_1, d(\MDir, p_k)\leq \epsilon_2) \right]\;.
\end{align*} 
The first term is upper bounded by $n_0(T)$ as the event $(A_t=k, N_k(t-1)\leq n_0(T))$ can occur at most $n_0(T)$ times. So we have
\begin{align*}
\text{(Post-CV)}\leq & n_0(T) + \sum_{t=1}^T \bE\left[\ind(A_t=k, N_k(t-1)\geq n_0(T), c_{k, t}^\alpha\geq c_1^\alpha-\epsilon_1, d(\MDir, p_k)\leq \epsilon_2) \right]\\
\leq& n_0(T) + \sum_{t=1}^T \bE\left[\ind(N_k(t-1)\geq n_0(T), c_{k, t}^\alpha\geq c_1^\alpha-\epsilon_1, d(\MDir, p_k)\leq \epsilon_2) \right]\\
\leq & n_0(T) + \sum_{t=1}^T \bE\left[\ind\left(N_k(t-1)\geq n_0(T), d(\MDir, p_k)\leq \epsilon_2\right) \times \bE\left[\ind\left(c_{k,t}^{\alpha} \geq c_1^\alpha - \epsilon_1\right) | \cF_{t-1} \right]\right]\\
= & n_0(T) + \sum_{t=1}^T \bE\left[\ind\left(N_k(t-1)\geq n_0(T), d(\MDir, p_k)\leq \epsilon_2\right) \times  \bP_{w \sim \mathrm{Dir}(\betaDir)}\left(C_\alpha(\cX_k, w) \geq c_1^\alpha - \epsilon_1 \right)\right]\;,
\end{align*} where we upper bounded $\ind(A_t=k)$ by $1$, so that $\ind\left(c_{k,t}^{\alpha} \geq c_1^\alpha - \epsilon_1\right)$ is the only term that is not $\cF_{t-1}$-measurable, and then used the law of total expectation. Now we can use Lemma~\ref{lem::lemma13_main} to control the probability term inside the expectation by
{\small \[\bP_{w \sim \mathrm{Dir}(\beta_{k}(t))}\left(C_\alpha(\cX_k, w) \geq c_1^\alpha - \epsilon_1\right) \leq C_1 M_k(N_k(t-1)+M_k)^\frac{M_k}{2}\exp\left(-(N_k(t-1)+M_k)\cK_{\inf}^{\alpha, \cX_k}(\MDir, c_1^\alpha-\epsilon_1) \right).\]}

At this stage, our objective is to remove the randomness in this upper bound in order to bound uniformly the terms inside the expectation. To do this, we will use the fact that $d(\MDir, p_k) \leq \epsilon_2$ and $N_k(t-1)\geq n_0(T)$ together with the continuity of the function $\cK_{\inf}^{\alpha, \cX_k}$ in its second argument, established in Lemma~\ref{lem:berge} defined in Appendix~\ref{app::berge}. 


Let $\varepsilon_3 \in (0,\cK_{\inf}^{\alpha, \cX_k}(\MDir, c_1^\alpha))$. There exists by continuity small enough values of $\varepsilon_1$ and $\varepsilon_2$ such that, if $d(\MDir,p_k) \leq \varepsilon_2$,  
\begin{align*}
\cK_{\inf}^{\alpha, \cX_k}(\MDir, c_1^\alpha-\epsilon_1) &\geq \cK_{\inf}^{\alpha, \cX_k}(p_k, c_1^\alpha-\epsilon_1) - \frac{\epsilon_3}{4}  \geq \cK_{\inf}^{\alpha, \cX_k}(p_k, c_1^\alpha) - \frac{\epsilon_3}{2},
\end{align*} hence
\begin{align*}
D := (N_k(t-1)+M_k)^\frac{M_k}{2}\exp\left(-(N_k(t-1)+M_k) \cK_{\inf}^{\alpha, \cX_k}(p_{k, t}, c_1^\alpha-\epsilon_1)\right) \\
\leq (N_k(t-1)+M_k)^\frac{M_k}{2}\exp\left(-(N_k(t-1)+M_k) (\cK_{\inf}^{\alpha, \cX_k}(p_k, c_1^\alpha) - \epsilon_3/2)\right).
\end{align*}
Using the fact that for any $b>0$ and $b > \epsilon>0$ there exists a constant $C'$ such that $\forall t>0$: $t\exp(-bt) \leq C'\exp(-t(b-\epsilon))$, we further get 
\begin{align*}
D &\leq C'\exp\left(-(N_k(t-1)+M_k) (\cK_{\inf}^{\alpha, \cX_k}(p_k, c_1^\alpha) - \epsilon_3)\right) \\
&\leq C' \exp\left(-(n_0(T)+M_k) (\cK_{\inf}^{\alpha, \cX_k}(p_k, c_1^\alpha) - \epsilon_3)\right),
\end{align*}
provided that $N_k(t-1) \geq n_0(T)$. 

Putting things together, we proved that for every $\varepsilon_3 \in (0,\cK_{\inf}^{\alpha, \cX_k}(\MDir, c_1^\alpha))$, if $\varepsilon_1$ and $\varepsilon_2$ are small enough, then there exists a constant $C_1' > 0$ such that 
\begin{align*}
\text{(Post-CV)} \leq & n_0(T) + \sum_{t=1}^T  C_1' \exp\left(-(n_0(T)+M_k)(\cK_{\inf}^{\alpha, \cX_k} \left(p_k, c_1^\alpha\right)-\epsilon_3)\right) \\
\leq & n_0(T) + T C_1'\exp\left(-(n_0(T)+M_k)(\cK_{\inf}^{\alpha, \cX_k} \left(p_k, c_1^\alpha\right)-\epsilon_3)\right).\end{align*}

Choosing $n_0(T)=\frac{\log T}{\cK_{\inf}^{\alpha, \cX_k}\left(\nu_k, c_1^\alpha-\epsilon_1\right)-\epsilon_3}-(M_k+1)$  yields the upper bound
\[(\text{Post-CV}) \leq \frac{\log T}{\cK_{\inf}^{\alpha, \cX_k}\left(p_k, c_1^\alpha\right)-\epsilon_3} + O(1).\]

Finally, we have shown that for any $\epsilon_0>0$, if $\varepsilon_1$ and $\varepsilon_2$ are small enough, then
\[(\text{Post-CV}) \leq \frac{(1+\epsilon_0)\log T}{\cK_{\inf}^{\alpha, \cX_k}\left(p_k, c_1^\alpha\right)} + O(1),\]
which proves Equation~\eqref{eq::post_cv_mcvts}.

\subsection{Proof of Equation~\eqref{eq::pre_cv}: Upper Bound on the Pre-Convergence term for multinomial distributions}\label{sec::precv_multi}

In this section, we upper bound the term $\text{(Pre-CV)}=\bE\left(\sum_{t=1}^T \ind(A_t=k, \{c_{k, t}^\alpha< c_1^\alpha-\epsilon_1 \cup d(\MDir,p_k)>\epsilon_2\})\right)$.

We first decompose this term into 
\[\text{(Pre-CV)} \leq \bE\left(\sum_{t=1}^T \ind(A_t=k, c_{k, t}^\alpha< c_1^\alpha-\epsilon_1)\right) + \bE\left(\sum_{t=1}^T \ind(A_t=k, d(\MDir, p_k)>\epsilon_2)\right)\,.\]

Let us remark that the second term does not feature any CVaR, hence we can directly use the upper bound derived by \cite{Honda} to get that, for any $\varepsilon_2>0$, 
\[\bE\left(\sum_{t=1}^T \ind(A_t=k, d(\MDir,p_k)>\epsilon_2)\right) \leq K M \left(\frac{2M}{\epsilon_2}+\frac{2}{\epsilon_2^2}\right).\]

Hence, it remains to upper bound the term 
\[A := \bE\left(\sum_{t=1}^T \ind(A_t=k, c_{k, t}^\alpha< c_1^\alpha-\epsilon_1)\right) \]
We write
\begin{align*}
A \leq &\bE\left[\sum_{t=1}^T \ind(c_{A_t, t}^\alpha< c_1^\alpha-\epsilon_1)\right]\\
\leq &\sum_{t=1}^T \sum_{n=1}^T \bE\left[\ind(c_{A_t, t}^\alpha< c_1^\alpha-\epsilon_1, N_1(t)=n) \right]\\
\leq & \sum_{m=1}^T \sum_{n=1}^T \bE\left[\ind\left(\sum_{t=1}^T\ind(c_{A_t, t}^\alpha< c_1^\alpha-\epsilon_1, N_1(t)=n)\geq m \right)  \right] \;,
\end{align*}
where we used as in \cite{Honda} that for any series of events $(E_t)$ it holds that 
\[\sum_{t=1}^T \ind(E_t) \leq \sum_{m=1}^T \ind\left(\sum_{t=1}^T \ind(E_t) \geq m\right).\]

We then introduce a random sequence $(\tau_i^n)_{i \in \N}$ where $\tau_i^n\in \R \cup \{+\infty\}$ is the $i$-th time at which the event $\{ \max_{j>1} c_{j,t}^\alpha \!\leq\! c_1^\alpha \!-\! \epsilon_1, N_1(t)\!=\!n\}$ holds. In order to ensure that this event occurs at least $m$ times, then we need 1) that $\tau_i^n \leq +\infty$ for all $i \leq m$, and 2) $c_{1, \tau_i^n}^\alpha \leq c_1^\alpha - \epsilon_1$ for all $i \leq m$, otherwise arm $1$ would be drawn. Hence, we have the following inclusion of events \[\left\{\sum_{t=1}^T\ind(c_{A_t, t}^\alpha< c_1^\alpha-\epsilon_1, N_1(t)=n)\geq m \right\} \subset \left\{\tau_i^n < +\infty, c_{1, \tau_i^n}^\alpha \leq c_1^\alpha - \epsilon_1\forall i \in \{1,\dots, m\} \right\}.\]

We then use the following arguments, for a fixed $n$: 1) since arm 1 has been drawn $n$ times at all (finite) time steps $\tau_i^n$, the random variables $\beta_{1,\tau_i^n}$ for $i$ such that $\tau_i^n < \infty$ are all equal to some common value $\beta_n$, which is such that $\beta_n - 1$ follows a multinomial distribution $\mathrm{Mult}(n, p_1)$. 2) the $c_{1, \tau_i^n}^\alpha$ are independent conditionally to ${\beta}_n$ and follow a $\mathrm{Dir}(\beta_n)$ distribution.

Therefore, we write 

\begin{align*}
A \leq &\sum_{n=1}^T \sum_{m=1}^T \bE \left[\prod_{i=1}^m \ind\left(\tau_i^n < +\infty, c_{1, \tau_i^n}^\alpha \leq c_1^\alpha - \epsilon_1\right)\right]\\
\leq & \sum_{n=1}^T \sum_{m=1}^T \bE_{\beta-1 \sim \mathrm{Mult}(n,p_1)}\left[\prod_{i=1}^m \bP_{w_i \sim \mathrm{Dir}(\beta)}\left(C_\alpha(\cX, w_i)\leq c_1^\alpha-\epsilon_1\right)\right]\\
\leq & \sum_{n=1}^T \sum_{m=1}^T \bE_{\beta-1 \sim \mathrm{Mult}(n,p_1)}\left[\bP_{w \sim \mathrm{Dir}(\beta)}\left(C_\alpha(\cX, w)\leq c_1^\alpha-\epsilon_1\right)^m\right] \\
\leq & \sum_{n=1}^T \bE_{\beta-1 \sim \mathrm{Mult}(n,p_1)}\left[\sum_{m=1}^T \bP_{w \sim \mathrm{Dir}(\beta)}\left(C_\alpha(\cX, w)\leq c_1^\alpha-\epsilon_1\right)^m\right] \\
\leq & \sum_{n=1}^T \bE_{\beta-1 \sim \mathrm{Mult}(n,p_1)}\left[\frac{\bP_{w \sim \mathrm{Dir}(\beta)}\left(C_\alpha(\cX, w)\leq c_1^\alpha - \epsilon_1\right)}{1- \bP_{w \sim \mathrm{Dir}(\beta)}\left(C_\alpha(\cX, w)\leq c_1^\alpha - \epsilon_1\right)}\right]\,.
\end{align*}

Thank to this bound, we have transformed our problem into the study of properties of the Dirichlet distribution. Similarly, we may now upper bound the last term in the above inequality by considering different regions to which the mean of the Dirichlet distribution -- that  is, $\frac{\beta}{n+M}$ -- belongs. However,  in order to account for a general risk level $\alpha$, the analysis is more intricate as we need to split the simplex into sub-spaces defined by different values of the \CVAR{}, not the mean. This requires to establish new boundary crossing probabilities involving those sub-spaces, which we provide now.

We decompose the upper-bound on $A$ into $\quad A \leq A_1 + A_2 + A_3,\quad$ where:
\begin{itemize}
	\item $A_1=\sum_{n=1}^T \bE_{\beta-1 \sim \mathrm{Mult}(n,p_1)}\left[\frac{\bP_{w \sim \mathrm{Dir}(\beta)}\left(C_\alpha(\cX, w)\leq c_1^\alpha - \epsilon_1\right)}{1- \bP_{w \sim \mathrm{Dir}(\beta)}\left(C_\alpha(\cX, w)\leq c_1^\alpha - \epsilon_1\right)} \ind\left(C_\alpha\left(\cX, \frac{\beta}{n+M}\right) \geq c_1^\alpha -\epsilon_1/2\right)\right]$
	\item $A_2=\sum_{n=1}^T \bE_{\beta-1 \sim \mathrm{Mult}(n,p_1)}\left[\frac{\bP_{w \sim \mathrm{Dir}(\beta)}\left(C_\alpha(\cX, w)\leq c_1^\alpha - \epsilon_1\right)}{1- \bP_{w \sim \mathrm{Dir}(\beta)}\left(C_\alpha(\cX, w)\leq c_1^\alpha - \epsilon_1\right)} \ind\left(c_1^\alpha - \epsilon_1 \leq C_\alpha\left(\cX, \frac{\beta}{n+M}\right) \leq c_1^\alpha -\epsilon_1/2\right)\right]$
	\item $A_3=\sum_{n=1}^T \bE_{\beta-1 \sim \mathrm{Mult}(n,p_1)}\left[\frac{\bP_{w \sim \mathrm{Dir}(\beta)}\left(C_\alpha(\cX, w)\leq c_1^\alpha - \epsilon_1\right)}{1- \bP_{w \sim \mathrm{Dir}(\beta)}\left(C_\alpha(\cX, w)\leq c_1^\alpha - \epsilon_1\right)} \ind\left(C_\alpha\left(\cX, \frac{\beta}{n+M}\right) \leq c_1^\alpha -\epsilon_1\right)\right]$.
\end{itemize}
We now upper bound each of these three terms by a constant, for any value of $\varepsilon_1$. 

\subsubsection{Upper bound on $A_1$}\label{app::bound_A1_mcvts}

This term is the easiest to control. Indeed the set $\{p \in \cP^{M}: C_\alpha(\cX, p) \leq c_1^{\alpha}-\epsilon_1 \}$ is closed and convex, hence we can apply the boundary crossing probability of Lemma~13 in \cite{Honda} on this subset and write
\[\bP_{w \sim \mathrm{Dir}(\beta)}\left(C_\alpha(\cX, w)\leq c_1^\alpha - \epsilon_1\right) \leq C_1 (n+M)^\frac{M}{2} \exp\left(-(n+M)\KL\left(\frac{\beta}{n+M},p_\beta^*\right)\right)\,,\]
with $p^*_{\beta}=\argmin{p : C_\alpha(\cX,p)\leq c_1^\alpha - \epsilon_1} \KL(\frac{\beta}{n+M},p)$. Now the quantity
\[\delta=\inf_{\substack{q:C_\alpha(\cX,q)\geq c_1^\alpha-\epsilon_1/2 \\ p: C_\alpha(\cX,p)\leq c_1^\alpha-\epsilon_1}} \KL(q,p)\]
satisfies $\delta > 0$ due to the following argument: the infimum of the continuous function $(q,p) \mapsto \KL(q,p)$ on a compact set is necessarily achieved in a point $(q^*,p^*)$. Assuming $\delta = 0$ yields $q^* = p^*$ while $C_\alpha(\cX, q^*) \geq c_1^\alpha-\epsilon_1/2$ and $C_{\alpha}(\cX, p^*) \leq c_1^\alpha-\epsilon_1$ which is not possible as $\varepsilon_1 >0$. Hence, if the event $\left\{C_{\alpha}\left(\cX, \frac{\beta}{n+M}\right) \geq c_1^\alpha -\epsilon_1/2\right\}$ holds, one has 
\[\bP_{w \sim \mathrm{Dir}(\beta)}\left(C_\alpha(\cX, w)\leq c_1^\alpha - \epsilon_1\right) \leq C_1 (n+M)^\frac{M}{2} \exp\left(-(n+M) \delta\right).\]
For $n\geq n_1$ large enough so that $C_1(n+M)^{M/2}\exp(-(n+M)\delta)<1$ it follows that
\[A_1 \leq n_1 + \sum_{n=n_1+1}^T \bE_{\beta-1 \sim \mathrm{Mult}(n,p)}\left(\ind(C_\alpha(\cX, q)\geq c_1-\epsilon_1/2) \frac{C_1 (n+M)^\frac{M}{2} \exp(-(n+M)\delta)}{1-C_1 (n+M)^\frac{M}{2} \exp(-(n+M)\delta)}\right).\]

Furthermore, for any $\gamma>1$, there exist some $n_\gamma$ satisfying $\forall n\geq n_\gamma$, $\frac{1}{1-C_1 (n+M)^\frac{M}{2} \exp(-(n+M)\delta)}\leq \gamma$, therefore: 
\begin{align*}
A_1 \leq& \max(n_1, n_\gamma) + \sum_{n=\max(n_1, n_\gamma)+1}^T \bE_{\beta-1 \sim \mathrm{Mult}(n,p_1)}\left(\gamma \ind(C_\alpha(\cX,q)\geq c_1-\epsilon_1/2) C_1 (n+M)^\frac{M}{2} \exp(-(n+M)\delta)\right)\\
\leq & \max(n_1, n_\gamma) + \sum_{n=\max(n_1, n_\gamma)+1}^T \gamma C_1 (n+M)^\frac{M}{2} \exp(-(n+M)\delta),
\end{align*}
and the right-hand side can be upper bounded by a constant.

\subsubsection{Upper bound on $A_2$}\label{app::A2_MCVTS}

For the term $A_2$ we ignore the numerator and hence study \[A_2 \leq \sum_{n=1}^T\bE_{\beta-1 \sim \mathrm{Mult}(n,p)}\left[\ind\left(c_1^\alpha- \epsilon_1 \leq C_\alpha\left(\cX,\frac{\beta}{n+M}\right) \leq c_1^\alpha-\epsilon_1/2\right) \frac{1}{\bP_{w \sim \mathrm{Dir}(\beta)}(C_\alpha(\cX, w)\geq c_1^\alpha - \epsilon_1)}\right]\,.\]
Using Lemma~\ref{lem::lemma14_main} we obtain
\[\bP_{w \sim \mathrm{Dir}(\beta)}(C_\alpha(\cX, w)\geq c_1^\alpha - \epsilon_1) \geq \frac{C_M}{(n+M)^{\frac{M+1}{2}}}\exp\left(-(n+M)\kinfcvx\left(\frac{\beta}{n+M}, c_1^\alpha-\epsilon_1 \right)\right)\;.\]

Now we note that, by definition, $\kinfcvx\left(\frac{\beta}{n+M}, c_1^\alpha-\epsilon_1 \right)=0$ if $c_1^\alpha- \epsilon_1 \leq C_\alpha\left(\cX,\frac{\beta}{n+M}\right)$, so the exponential term is equal to 1 in that case, leading to 
\begin{align*}
A_2 \leq & \sum_{n=1}^T\bE_{\beta-1 \sim \mathrm{Mult}(n,p)}\left[\ind\left(c_1^\alpha- \epsilon_1 \leq C_\alpha\left(\cX,\frac{\beta}{n+M}\right) \leq c_1^\alpha-\epsilon_1/2\right) C_2^{-1}(n+M)^{3M/2+1}\right] \\
\leq & \sum_{n=1}^T \frac{n^\frac{M+1}{2}}{C_M} \bP_{\beta-1 \sim \mathrm{Mult}(n,p)}\left(C_\alpha\left(\cX,\frac{\beta}{n+M}\right) \leq c_1^\alpha - \epsilon_1/2\right)\;.
\end{align*}
To conclude, we make use of a concentration inequality for the empirical CVaR derived from Brown's inequality \citep{brown_ineq}. However, to express the probability in the right-hand side in terms of the CVaR of an empirical distribution, we need to handle the bias of $\beta$ induced by the initialization step. For this we use the fact that for any integers $n_k, M$: \[\left|\frac{n_k}{n}-\frac{n_k+1}{n+M}\right| = \frac{1}{n+M}\left|\frac{n_k}{n}M-1\right| \leq \frac{M}{n+M}\;. \]
As a direct consequence, when $n$ is large enough the biased empirical distribution $\frac{\beta}{n+M}$ can be made as close to the empirical distribution  $\frac{\beta-1}{n}$ as we want. Thanks to Lemma~\ref{lem::comp_cvar} we indeed get
\[\left|C_\alpha\left(\cX,\frac{\beta}{n+M}\right)-C_\alpha\left(\cX,\frac{\beta-1}{n}\right)\right| \leq \frac{M^2 x_M}{\alpha(n+M)}.\]

So for $n\geq n'$ large enough it holds that \[\bP_{\beta-1 \sim \mathrm{Mult}(n,p)}\left(C_\alpha\left(\cX,\frac{\beta}{n+M}\right) \leq c_1^\alpha - \epsilon_1/2\right) \leq \bP_{\beta-1 \sim \mathrm{Mult}(n,p)}\left(C_\alpha\left(\cX,\frac{\beta-1}{n}\right) \leq c_1^\alpha - \epsilon_1/3\right).\]

We are now ready to apply Lemma~\ref{lem:brown} (Brown inequality) in Appendix~\ref{app::brown_ucb}, which yields
\[
\bP_{\beta-1 \sim \mathrm{Mult}(n,p)}\left(C_\alpha\left(\cX,\frac{\beta-1}{n}\right) \leq c_1^\alpha - \epsilon_1/3\right)\leq \exp\left(-n \frac{2\alpha^2 \epsilon_1^2}{9} \right)\,.\]
This entails that we can upper bound $A_2$ by a constant: 
\[A_2 \leq \sum_{n=1}^T \frac{n^\frac{M+1}{2}}{C_M} \exp\left(-n \frac{2\alpha^2 \epsilon_1^2}{9} \right)=\mathcal{O}(1).\]

We remark that the assumption considered by Brown's inequality requires the random variables to be both \textit{positive} and \textit{bounded}. For instance, \citet{prashanth2019concentration}[Theorem 3.3] proved that a similar result (i.e an exponential inequality with a scaling in $n\alpha^2\epsilon^2$) hold for random variables that are non necessarily positive; their concentration bound is a little more complicated. Hence, we work with Brown's inequality in this proof for the sake of simplicity but our algorithm does not actually require the variables to be positive. Furthermore, note that Brown's inequality is not used here in order to  control the first order terms of our regret bound, hence we do not focus on obtaining the tightest concentration bounds for the concentration of the empirical CVaR, as it only affects second order terms.   

\subsubsection{Upper bound on $A_3$}\label{sec::bound_A3_M}

Similarly to what we presented in the previous section we write
\[A_3 \leq \sum_{n=1}^T\bE_{\beta-1 \sim \mathrm{Mult}(n,p)}\left(\ind\left(C_\alpha\left(\cX,\frac{\beta}{n+M}\right) < c_1^\alpha-\epsilon_1\right) \frac{1}{\bP_{w \sim \mathrm{Dir}(\beta)}(C_\alpha(\cX, w)\geq c_1^\alpha - \epsilon_1)}\right)\;.\]

We now use the second inequality in Lemma~\ref{lem::lemma14_main} in order to balance the probability that a specific count vector $\beta-1$ is observed after collecting $n$ observations from arm $1$ (hence $\beta$ includes the additional element in the initialization of the algorithm) and the probability that the re-sampled CVaR exceeds the threshold. We denote $\cM$ the set of vectors $\beta$ for which the CVaR of the distribution associated with $\frac{\beta}{n+M}$ is smaller than $c_1^\alpha-\epsilon_1$. Then, for any $n$
\begin{align*}
&\bE_{\beta-1 \sim \mathrm{Mult}(n,p)}\left[\ind\left(C_\alpha\left(\cX,\frac{\beta}{n+M}\right) < c_1^\alpha-\epsilon_1\right) \frac{1}{\bP_{w \sim \mathrm{Dir}(\beta)}(C_\alpha(\cX, w)\geq c_1^\alpha - \epsilon_1)}\right] \\ =& \sum_{\beta \in \cM} \frac{\bP_{\beta-1 \sim \mathrm{Mult}(n,p)}(\beta-1)}{\bP_{w \sim \mathrm{Dir}(\beta)}(C_\alpha(\cX, w)\geq c_1^\alpha - \epsilon_1)}   \\
\leq& \sum_{\beta \in \cM} n! \prod_{i=1}^M\frac{p_i^{\beta_i-1}}{(\beta_i-1)!} \times \frac{e^{(n+M) \kinfcvx\left(\frac{\beta}{n+M}, c_1^\alpha-\epsilon_1\right)}}{\Gamma(n+M)} \prod_{j=1}^M \frac{(\beta_j-1)!}{(\beta_j/(n+M))^{\beta_j}} \\
=& \sum_{\beta \in \cM} \frac{n!}{(n+M-1)!} \prod_{i=1}^M \left(\frac{p_i}{\beta_i/(n+M)}\right)^{\beta_i}e^{(n+M) \kinfcvx\left(\frac{\beta}{n+M}, c_1^\alpha-\epsilon_1\right)} \times \frac{1}{\prod_{j=1}^M p_j} \\
=& \sum_{\beta \in \cM} \frac{n!}{(n+M-1)!} \exp\left(-(n+M) \left(\KL\left(\frac{\beta}{n+M}, p_1\right)-\kinfcvx\left(\frac{\beta}{n+M}, c_1^\alpha-\epsilon_1\right)\right)\right) \times \frac{1}{\prod_{j=1}^M p_j} \\
\leq & \frac{n^M}{n^{M-1}}\exp\left(-(n+M) \left(\KL\left(\frac{\beta}{n+M}, p_1\right)-\kinfcvx\left(\frac{\beta}{n+M}, c_1^\alpha-\epsilon_1\right)\right)\right) \times \frac{1}{\prod_{j=1}^M p_j} \\
\end{align*}
Hence, denoting $J(p)= \prod_{i=1}^M \frac{1}{p_i}$ (which is a problem-dependent constant) we obtain 
\begin{align*}
A_3 & \leq J(p) \sum_{n=1}^T n \exp\left(-n \delta_1\right) \leq J(p) \frac{e^{-\delta_1}}{(1-e^{-\delta_1})^2} \\
\end{align*}
where we introduced the quantity \[\delta_1= \inf_{p : \CVAR_{\alpha}^{\cX}(p) \leq c_1^{\alpha} - \varepsilon_1} \left[\kinfcvx(p, c_1^\alpha)-\kinfcvx(p, c_1^\alpha-\epsilon_1)\right]\;.\]
Since $\kinfcvx$ is continuous in its first argument and we take the infimum over a compact set, this infimum is reached for some distribution $p_{\inf}^\alpha$. That is, we can write 
\[\delta_1 = \kinfcvx(p_{\inf}^{\alpha}, c_1^\alpha)-\kinfcvx(p_{\inf}^{\alpha}, c_1^\alpha-\epsilon_1),\]
and $\CVAR_{\alpha}^{\cX}(p_{\inf}^{\alpha}) \leq c_1^{\alpha} - \varepsilon_1$. 
Thanks to Lemma~\ref{lem:berge}, we know that the mapping $c \mapsto \kinfcvx(p_{\inf}^{\alpha}, c)$ is strictly increasing for $c \geq c_1^{\alpha} - \varepsilon_1$, thus $\delta_1 > 0$. This shows that A3 is upper-bounded by a constant depending only on $p$ and the CVaR level $\alpha$.
\section{Proof of Theorem~\ref{th::regret_CVTS} : analysis of \CVTS{} for continuous bounded distributions}\label{sec::proof_cvts}

Similarly to the proof techniques used to analyze \MCVTS{}, we use Equation~\eqref{eq::control_nb_pulls} presented in Section~\ref{sec::analysis} in the proof of  Theorem~\ref{th::regret_CVTS}.

In particular, we first prove Equation~\eqref{eq::post_cv_cvts} (Post-CV term) before proving Equation~\eqref{eq::pre_cv} (Pre-CV term), assuming that the arms are continuous, bounded, and that an upper bound on their support is known. In this section we use the Levy distance  presented in Appendix~\ref{app::notations} in order to compare the empirical cdf $F_{k, t}$ with $F_k$ for each arm $k$.

\subsection{Proof of Equation \eqref{eq::post_cv_cvts}: Upper Bound on the Post-Convergence term}\label{sec::postcv_gen}

We upper bound the term $\text{(Post-CV)} = \bE\left[\sum_{t=1}^T \ind(A_t=k, c_{k, t}^\alpha \geq c_1^\alpha-\epsilon_1, D_L(F_{k, t}, F_k) \leq \epsilon_2)\right]$. 
The change from using the $L^{\infty}$ distance to using the Levy metric does not affect any argument in the beginning of the proof used to upper bound (Post-CV). Hence, following the same steps as in section~\ref{sec::postcv_multi}, for any $n_0(T)$ it holds that
\begin{align*}
\text{(Post-CV)}\leq & n_0(T) + \sum_{t=1}^T \bE\left[\ind\left(N_k(t-1)\geq n_0(T), D_L(F_{k, t}, F_k)\leq \epsilon_2\right) \times  \bP_{w \sim \mathrm{\cD_{N_k(t)}}}\left(C_\alpha(\cX_{k, t}, w) \geq c_1^\alpha - \epsilon_1 \right)\right]\,.
\end{align*} 

We then use Lemma~\ref{lem::lem_15_main} in order to control the probability term inside the expectation as follows
{\small \[\bP_{w \sim \mathrm{\cD_{N_k(t)}}}\left(C_\alpha(\cX_{k, t}, w) \geq c_1^\alpha - \epsilon_1 \right) \leq \frac{1}{\eta} \exp\left(- N_k(t-1) \left(\kinfcvbk(F_k, c_1^\alpha-\epsilon_1)- \eta C(\alpha, B_k, c_1^\alpha-\epsilon_1) \right) \right)\;,\]}

where $\kinfcvbk(F_k, c_1^\alpha-\epsilon_1)$ is the functional defined in Section~\ref{sec::analysis}, applied on the set of continuous bounded distributions defined on $[0, B_k]$, $\cB_k$.

We then proceed by removing the randomness in this upper bound by bounding uniformly the terms inside the expectation, using that $D_L(F_{k, t}, F_k) \leq \epsilon_2$ and $N_k(t-1)\geq n_0(T)$. To do so, we now use the continuity of the mapping $\kinfcvbk$, which is proved using Lemma~4 from \cite{agrawal2020optimal}. 
Note that this is not a trivial result, as for instance the Levy topology does not coincide with the one induced by the Kullback-Leibler divergence. Combining these elements it holds that for any $\epsilon_0$, there exist $\eta>0$ such that 
\begin{align*}
\text{(Post-CV)} \leq & n_0(T) + \sum_{t=1}^T  \frac{1}{\eta} \exp\left(-n_0(T)\left(\kinfcvbk(F_k, c_1) - \epsilon_0\right)\right) \\
\leq & n_0(T) + \frac{1}{\eta}T \exp\left(-n_0(T)\left(\kinfcvbk(F_k, c_1) - \epsilon_0\right)\right) \;.\end{align*}

Choosing $n_0(T)=\frac{\log T}{\kinfcvbk(F_k, c_1^\alpha) - \epsilon_0}$ we upper bound the post-convergence term as
\[(\text{Post-CV}) \leq \frac{\log T}{\kinfcvbk\left(F_k, c_1^\alpha\right)-\epsilon_0} + O(1).\]

\subsection{Pre-Convergence term}\label{sec::precv_gen}

In this section, we now focus on providing an upper bound on the remainder term $\text{(Pre-CV)}=\bE\left(\sum_{t=1}^T \ind(A_t=k, \{c_{k, t}^\alpha< c_1^\alpha-\epsilon_1 \cup D_L(F_{k, t},F_k)>\epsilon_2\})\right)$.

We first decompose this term into
\[\text{(Pre-CV)} \leq \bE\left[\sum_{t=1}^T \ind(A_t=k, c_{k, t}^\alpha< c_1^\alpha-\epsilon_1)\right] + \bE\left[\sum_{t=1}^T \ind(A_t=k, D_L(F_{k, t}, F_k)>\epsilon_2)\right]\,.\]

Again, as the second term does not feature any CVaR, we can again use a result from \citet{Honda} (section D.1) to get that, for any $\varepsilon_2>0$, 
\[\bE\left[\sum_{t=1}^T \ind(A_t=k, D_L(F_{k, t},F_k)>\epsilon_2)\right] \leq K(M+1)\left(1+\sum_{n=2}^{+\infty} 2(n+1)\exp\left(-2(n-1)\left(\varepsilon_2-\frac{1}{n-1}\right)\right)\right).\]

Hence, if an arm is pulled a lot, its empirical distribution will be with high probability into a Levy ball of size $\epsilon_2$ around its true distribution. Hence, it remains to upper bound the term 
\[\bar A := \bE\left[\sum_{t=1}^T \ind(A_t=k, c_{k, t}^\alpha< c_1^\alpha-\epsilon_1)\right]\,.\]

We follow the exact same steps as in section~\ref{sec::precv_multi} and obtain again an expression of the form
\begin{align*}
\bar A \leq & \sum_{n=1}^T \bE_{X_1, \dots, X_n}\left[\frac{\bP_{w \sim \cD_{n}}\left(C_\alpha(\cX_n, w) \leq c_1^\alpha - \epsilon_1\right)}{1- \bP_{w \sim \cD_n}\left(C_\alpha(\cX_n, w)\leq c_1^\alpha - \epsilon_1\right)}\right] \;, 
\end{align*}

where we write here $\cX_n$ as the support of the empirical distribution of arm $1$ after receiving $n$ observations. Let us recall that $\cD_n$ is the uniform probability on the simplex of size $n$, namely the Dirichlet distribution with parameter $1_n = (1, \dots, 1)$. Thanks to the fact that $c_{A_t, t} < c_1^\alpha - \epsilon_1 \Rightarrow c_{1, t} < c_1^{\alpha} - \epsilon$ we can upper bound this term by the probability that the best arm under-performs.

As in section~\ref{sec::precv_multi}, we split this expectation into different regions depending of the value of the CVaR of the empirical distribution (that includes the term $x_0=B$ added at the beginning of the history of observations).

We split the upper bound on $\bar A $ into three terms $$\bar A \leq \bar A_1 + \bar A_2 + \bar A_3 \;, $$ where
\begin{itemize}
	\item $\bar A_1=\sum_{n=1}^T \bE_{\cX_n}\left[\frac{\bP_{w \sim \cD_{n}}\left(C_\alpha(\cX_n, w) \leq c_1^\alpha - \epsilon_1\right)}{1- \bP_{w \sim \cD_n}\left(C_\alpha(\cX_n, w)\leq c_1^\alpha - \epsilon_1\right)} \ind\left(C_\alpha(\cX_n) \geq c_1^\alpha -\epsilon_1/2\right)\right] \;,$
	\item $\bar A_2=\sum_{n=1}^T \bE_{\cX_n}\left[\frac{\bP_{w \sim \cD_{n}}\left(C_\alpha(\cX_n, w) \leq c_1^\alpha - \epsilon_1\right)}{1- \bP_{w \sim \cD_n}\left(C_\alpha(\cX_n, w)\leq c_1^\alpha - \epsilon_1\right)} \ind\left(c_1^\alpha - \epsilon_1 \leq C_\alpha(\cX_n) \leq c_1^\alpha -\epsilon_1/2\right)\right]$,
	\item $\bar A_3=\sum_{n=1}^T \bE_{\cX_n}\left[\frac{\bP_{w \sim \cD_{n}}\left(C_\alpha(\cX_n, w) \leq c_1^\alpha - \epsilon_1\right)}{1- \bP_{w \sim \cD_n}\left(C_\alpha(\cX_n, w)\leq c_1^\alpha - \epsilon_1\right)} \ind\left(C_\alpha(\cX_n) \leq c_1^\alpha -\epsilon_1\right)\right]$.
\end{itemize}
We now upper bound each of these three terms, for any value of $\varepsilon_1$. 

\subsubsection{Upper bound on $\bar A_1$}\label{app::A1_cvts}

The first case is again easier than the two others, because in this case the CVaR of arm $1$ is greater than $c_1^\alpha-\epsilon_1/2$ and so we can upper bound the term $\bar A_1$ by upper bounding 
\[\bP_{w \sim \cD_n}\left(C_\alpha(\cX_n, w)\leq c_1^\alpha - \epsilon_1\right)\;.\]

This term should be small as the empirical CVaR does not belong to the sub-space defined by this inequality. Furthermore, we can use a quantization of the random observations $X_1, \dots, X_n$, defining a number of bins $M$ that we will specify later, and for any $i \in \{0, \dots, n\}$  the random variables $\widetilde X_i = \frac{\lfloor M X_i \rfloor}{M}$. Denoting the corresponding set of truncated observations $\widetilde \cX_n$, since for all $i\in \{0, \dots, n\}, X_i-1/M \leq \tilde X_i \leq X_i$ we then obtain that \[C_\alpha(\cX_n)-\frac{1}{M} \leq C_\alpha(\widetilde \cX_n)  \leq C_\alpha(\cX_n)\;.\]

A similar control holds for $C_\alpha(\cX_n, w)$ and $C_\alpha(\widetilde \cX_n, w)$. Interestingly, these properties directly follow from the monotonicity of the CVaR, which is itself a property of any coherent risk measure (see \citet{acerbi2002coherence}).

Using these properties, we first have that \[\bP_{w \sim \cD_n}\left(C_\alpha(\cX_n, w)\leq c_1^\alpha - \epsilon_1\right) \leq \bP_{w \sim \cD_n}\left(C_\alpha(\widetilde \cX_n, w)\leq c_1^\alpha - \epsilon_1\right)  \;, \]

as well as

\[\ind\left(C_\alpha(\cX_n) \geq c_1^\alpha -\epsilon_1/2\right) \leq  \ind\left(C_\alpha(\widetilde \cX_n) \geq c_1^\alpha -\epsilon_1/2-\frac{1}{M}\right) \;.\]

Therefore, we have shown that we can upper bound the first term $\bar A_1$ by 
\begin{align*} \bar A_1 & \leq \sum_{n=1}^T\bE_{\cX_n}\left[\ind\left(C_\alpha\left(\widetilde \cX_n, w\right) \geq c_1^\alpha-\frac{\epsilon_1}{2}-\frac{1}{M}\right) \frac{\bP_{w \sim \cD_n}(C_\alpha\left(\widetilde \cX_n, w\right)\leq c_1^\alpha - \epsilon_1)}{1-\bP_{w \sim \cD_n}(C_\alpha\left(\widetilde \cX_n, w\right)\leq c_1^\alpha - \epsilon_1)}\right] \;. \\
\end{align*}

We then choose the discretization step $\frac{1}{M}$. First, we want this step to be small enough in order to preserve the order of the CVaRs, this in turns can be done by choosing $\epsilon_1$ small enough. Secondly, we want that $c_1^\alpha-\frac{\epsilon_1}{2}-\frac{1}{M} > c_1^\alpha - \epsilon_1$. This condition requires $M> 2/\epsilon_1$, so we choose (for instance) $M=\lceil 3/\epsilon_1 \rceil $.

We can now resort to Lemma~13 of \citet{Honda} in order to upper bound the probability involving the Dirichlet distribution, namely

\[\bP_{w \sim \cD_n}(C_\alpha\left(\widetilde \cX_n, w\right)\leq c_1^\alpha - \epsilon_1) \leq C_1 n^{M/2} \exp\left(-n \cK_{\text{inf}}^{\widetilde \cX_n}\left(\widetilde F_{k, n}, c_1^\alpha - \frac{\epsilon_1}{2}- \frac{1}{M}\right)\right) \;,\]

where $\widetilde F_{k, n}$ is the cdf of the empirical distribution corresponding to $\widetilde \cX_n$. Since we could then transform our problem in order to consider multinomial random variables, we can use the same steps as in the corresponding part of the regret analysis of \MCVTS. Hence, following similar steps as in Appendix~\ref{app::bound_A1_mcvts}, leads to the bound
\[\bar A_1 = O(1) \;.\]

\subsubsection{Upper bound on $\bar A_2$}\label{app::A2}

In order to control the term $A_2$ in Appendix~\ref{app::A2_MCVTS} we ignore the numerator and write \[\bar A_2 \leq \sum_{n=1}^T\bE_{\cX_n}\left[\ind\left(c_1^\alpha- \epsilon_1 \leq C_\alpha(\cX_n) \leq c_1^\alpha-\epsilon_1/2\right) \frac{1}{\bP_{w \sim \cD_n}(C_\alpha(\cX_n, w)\geq c_1^\alpha - \epsilon_1)}\right]\;. \]

We then use Lemma~\ref{lem::lower_lem17_main} in order to upper bound the right hand term, which yields 
\[\bar A_2 \leq \bE_{x_1, \dots, x_n}\left[\ind\left(c_1^\alpha- \epsilon_1 \leq C_\alpha(\cX_n) \leq c_1^\alpha-\epsilon_1/2\right) \frac{25n^3 \ind(Y_1 < c_1^\alpha-\epsilon_1)}{Y_{\lceil n \alpha \rceil}-Y_1}\right]\;.\]

Here, we have introduced $Y_1, \dots, Y_n$ to denote the ordered list of $(X_1, \dots, X_n)$ (i.e $Y_1 \leq Y_2 \leq \dots \leq Y_n$), where for any $j\in \N^*$ the variable $X_j$ represents the $j$-th observation collected from arm $1$. We also added the indicator $\ind(Y_1 \leq c_1^\alpha-\epsilon_1)$ because it is a necessary element of the next steps of the proof that aims at controlling $Y_1$. The inequality holds because if $Y_1 \geq c_1^\alpha-\epsilon_1$ then $C_\alpha(\cX_n, w) \geq c_1^\alpha - \epsilon_1$ for any $w \in \cP^n$. Then, under the events we consider it also holds that \[Y_{\lceil n \alpha \rceil} \geq C_\alpha(\cX_n) \geq c_1^\alpha-\epsilon_1 \;.\]

Note that it is impossible to conclude at this step in general because the variable $Y_{\lceil n \alpha \rceil}-Y_1$ may be arbitrarily small in case all the $n$ observations are very concentrated. However, if $n$ is large and the distribution is \textit{continuous} this event can only happen with a very low probability. This is a place in the proof where continuity is crucial.
To do so, we upper bound the rest of the terms with a peeling argument on the values of $Y_1$. This is done using the closed-form formulas for the distribution of the minimum of $n$ random variable that are independent and identically distributed. Indeed, if $f_1$ denotes the density of arms $1$, and we write the cdf and pdf of the minimum of $n$ independent observations of $\nu_1$ respectively $L_n$ and $l_n$, then it holds that $\forall x \in [0, B]$

\[L_n(x) = 1-(1-F_1(x))^n  \;.\]
 
Now, since $\nu_1$ is continuous it follows that in each point the density is $l_n(x)=n f_1(x) (1-F_1(x))^{n-1}$. The next step consists in defining a strictly decreasing sequence $(a_k)_{k \geq 0}$, and to look at the intervals $[c_1^\alpha-a_k-\epsilon_1, c_1^\alpha-a_{k+1}-\epsilon_1]$. On each of these intervals we obtain
by construction that $Y_{\lceil n \alpha \rceil}\geq c_1^\alpha-\epsilon_1\geq Y_1+a_{k+1}$, and thus
\[\bE_{\cX_n}\left[\frac{25n^3}{Y_{\lceil n \alpha \rceil}-Y_1} \ind\left(Y_1 \in [c_1^\alpha-a_k-\epsilon_1, c_1^\alpha-a_{k+1}-\epsilon_1]\right)\right] \leq \frac{25 n^3}{a_{k+1}}\times \bP\left(Y_1 \in [c_1^\alpha-a_k-\epsilon_1, c_1^\alpha-a_{k+1}-\epsilon_1]\right)\;.\]

Using the properties of the density  $l_n$ it holds that \begin{align*} \bP\left(y_1 \in [c_1^\alpha-a_k-\epsilon_1, c_1^\alpha-a_{k+1}-\epsilon_1]\right) &= \int_{c_1^\alpha -\epsilon_1- a_k}^{c_1^\alpha-\epsilon_1 - a_{k+1}} n f_1(x) (1-F_1(x))^{n-1} dx \\
& \leq \sup_{x \in [0, B] }f_1(x) \int_{c_1^\alpha-\epsilon_1 - a_k}^{c_1^\alpha - \epsilon_1 - a_{k+1}} n (1-F_1(x))^{n-1} dx \\ 
& \leq \sup_{x \in [0, B] }f_1(x) (a_k - a_{k+1}) n (1-F_1(c_1^\alpha - \epsilon_1-a_k))^{n-1} 
\;.\end{align*}

With these results at hand, we can now aim at upper bounding $\bar A_2$.
To this end, we first introduce

\begin{align*}
\bar A_2 & \leq \bE_{x_1, \dots, x_n}\left[\ind\left(c_1^\alpha- \epsilon_1 \leq C_\alpha(\cX_n) \leq c_1^\alpha-\epsilon_1/2\right) \frac{25n^3}{Y_{\lceil n \alpha \rceil}-Y_1}\right] \\
& \leq \underbrace{\bE_{x_1, \dots, x_n}\left[\ind\left(c_1^\alpha- \epsilon_1 \leq C_\alpha(\cX_n) \leq c_1^\alpha-\epsilon_1/2\right) \frac{25n^3}{Y_{\lceil n \alpha \rceil}-Y_1} \ind(y_1 \leq c_1^\alpha-\epsilon_1-a_0)\right]}_{A_{21}}  \\ & + \underbrace{\bE_{x_1, \dots, x_n}\left[\ind\left(c_1^\alpha- \epsilon_1 \leq C_\alpha(\cX_n) \leq c_1^\alpha-\epsilon_1/2\right) \frac{25n^3}{Y_{\lceil n \alpha \rceil}-Y_1} \ind(y_1 \geq c_1^\alpha-\epsilon_1-a_0)\right]}_{A_{22}}\,.
\end{align*}

The left-hand side term can be handled thanks to Brown's inequality \cite{brown_ineq}, that we restate in Lemma~\ref{lem:brown}  for completeness, and discuss in Appendix ~\ref{app::A2_MCVTS}. Using that $Y_{\lceil n \alpha \rceil} - Y_1 \geq a_0$ on the  considered interval, we obtain

\[A_{21} \leq \frac{25n^3}{a_0} e^{-2n \left(\frac{\alpha(a_0+\epsilon_1)}{B_k}\right)^2} \;.\]

Regarding the second term $A_{22}$ we have 

\[A_{22} \leq \sup_{x \in [0, B]} n f_1(x) \times \sum_{k=0}^{+\infty} \frac{a_k-a_{k+1}}{a_{k+1}} (1-F_1(c_1^\alpha-\epsilon_1-a_k))^{n-1} \;. \]

We first use that the cdf is increasing, which enables to upper bound $(1-F_1(c_1^\alpha-\epsilon_1-a_k))^{n-1}$ by the quantity $(1-F_1(c_1^\alpha-\epsilon_1- a_0))^{n-1}$. It remains to choose the sequence $(a_k)$ in order to make the sum $\sum_{k=0}^{+\infty} \frac{a_k- a_{k+1}}{a_{k+1}}$ converge. We define recursively the sequence as $a_{k+1}=\frac{2^k}{2^k+1} a_k$, starting from $a_0=\frac{c_1^\alpha-\epsilon_1}{2}$. This way, $\sum_{k=0}^{+\infty} \frac{a_k- a_{k+1}}{a_{k+1}} = \sum_{k=0}^{+\infty} \frac{1}{2^k} = 2$. This shows that

\[A_{22}\leq 50 n^4 \sup_{x \in [0, B]} f_1(x) \exp\left(-n \log(1-F_1(c_1^\alpha-\epsilon_1))\right) \;. \]

Hence, both terms $A_{21}$ and $A_{22}$ are asymptotically negligible, hence we can write that $\bar A_2 = O(1)$.

\subsubsection{Upper bound on $\bar A_3$}

We now turn to the last term $\bar A_2$, and first upper bound it as
\[\bar A_3 \leq \sum_{n=1}^T\bE_{\cX_n}\left[ \ind\left(C_\alpha(\cX_n)  < c_1^\alpha-\epsilon_1\right) \frac{1}{\bP_{w \sim \cD_n}(C_\alpha(\cX_n, w)\geq c_1^\alpha - \epsilon_1)}\right]\,.\]

We can use the same discretization arguments as in Appendix~\ref{app::A1_cvts} to handle this term. More precisely, we introduce a number of bins $M'$ that is specified later in the proof, and for any $i \in \{0, \dots, n\}$  we again define $\widetilde X_i = \frac{\lfloor M X_i \rfloor}{M}$ and $\widetilde \cX_n$ the corresponding set of truncated observations. Thanks to these definitions we can upper bound $\bar A_3$ as

\begin{align*} \bar A_3 & \leq \sum_{n=1}^T\bE_{\cX_n}\left[\ind\left(C_\alpha\left(\widetilde \cX_n, w\right) < c_1^\alpha-\epsilon_1\right) \frac{1}{\bP_{w \sim \cD_n}(C_\alpha(\tilde \cX_n, w)\geq c_1^\alpha - \epsilon_1 - 1/M)}\right] \\
& \leq \sum_{n=1}^T\bE_{\cX_n}\left[\ind\left(C_\alpha\left(\widetilde \cX_n, w\right) < c_1^\alpha-\epsilon_1 - \frac{1}{M}\right) \frac{1}{\bP_{w \sim \cD_n}(C_\alpha(\tilde \cX_n, w)\geq c_1^\alpha -\epsilon_1-1/M)}\right]\,.
\end{align*}

Now, we use the first result of Lemma~\ref{lem::lemma14_main} that provides with the additional observation on the last item
\[\bP_{w \sim \cD_n}(C_\alpha(\tilde \cX_n, w)\geq c_1^\alpha -\epsilon_1-1/M)\geq \frac{1}{n} \frac{n!}{\prod_{i=1}^M \beta_i}\prod_{j=1}^M (p_j^\star)^{\beta_j}\;,\]
where $p^\star \in \cP^M$ is again a well chosen weight vector. We then use the same proof techniques as in Appendix~\ref{app::A1_cvts}, only requiring $\epsilon_1$ to be small enough in order to keep the same order for the CVaR of the discretized distributions. To proceed, we then choose $M$ of the same order as $1/\epsilon_1$, writing $\epsilon_1+1/M = \epsilon_1'$. Using with a slight abuse of notations $\mathcal{M}$ the set of allocations considered in the expectation, we have now
\begin{align*}
\bar A_3 &\leq \sum_{n=1}^T \sum_{\beta \in \cM}n \frac{\frac{n!}{\prod_{i=1}^M \beta_i!} \prod_{j=1}^M (\widetilde p_i)^{\beta_i}}{\frac{n!}{\prod_{i=1}^M \beta_i!} \prod_{j=1}^M (p_i^\star)^{\beta_i}} \\
& = \sum_{n=1}^T n \sum_{\beta \in \cM} \prod_{j=1}^M \left(\frac{\widetilde p_i}{p_i^\star} \right)^{\beta_i} \\
& = \sum_{n=1}^T n \sum_{\beta \in \cM} \exp\left(-n \left(\KL\left(\frac{\beta}{n}, \widetilde p \right)- \KL\left(\frac{\beta}{n}, p^\star \right)\right) \right)
\end{align*}

Then we remark that this new formulation is exactly equivalent to the one we got in \ref{sec::bound_A3_M}. Hence, introducing
 \[\delta_2= \inf_{p \in \cP^{M}: C_\alpha(\cX, p) \leq c_1^{\alpha} - \varepsilon_1'} \left[\cK_{\text{inf}}^{\alpha, \tilde \cX}(p, c_1^\alpha)-\cK_{\text{inf}}^{\alpha, \tilde \cX}(p, c_1^\alpha-\epsilon_1')\right]\;, \]

where $\widetilde \cX$ is $\widetilde \cX_n$ where each item is only repeated once (the set built from $\widetilde \cX_n$), the same steps allow us to finally obtain 

\begin{align*}
\bar A_3 \leq \sum_{n=1}^T n^{M+1} \exp(-n \delta_2) = \mathcal{O}(1) \\ 
\end{align*}

Hence, this is again upper bounded by a constant. This final result concludes the proof of Equation~\eqref{eq::pre_cv} for continuous bounded distribution, which states that for \CVTS{} \[\text{(Pre-CV)} = O(1) \;.\]
\section{Lower Bound and properties of $\kinfcvgen$}\label{app::asymp_opt}

In the classical bandit setting, asymptotic optimality is an important notion that has guided the design of algorithms, and we investigate in this section the optimal (problem-dependent) scaling of the \CVAR-regret. We start by proving Theorem~\ref{thm:LB}, and then investigate some properties of the obtained lower bound that permit to derive concentration inequalities for the Dirichlet distributions.

\subsection{Proof of Theorem~\ref{thm:LB}} \label{app:LB}

In this section, we prove Theorem~\ref{thm:LB}. We rely on the fundamental inequality (6) of \cite{menard_asymp} which shows that, if $\nu$ and $\nu'$ are two bandit models in $\cD$, for any $\cF_T$-measurable random variable $Z \in [0,1]$, \[\sum_{k=1}^K \bE_{\pi,\nu}[N_k(T)]\KL(\nu_k, \nu_k') \geq \kl(\bE_{\pi,\nu}[Z],\bE_{\pi,\nu'}[Z])\,,\] 
where $\kl(x, y)=x \log\left(\frac{x}{y}\right) + (1-x)\log\left(\frac{1-x}{1-y} \right)$ denotes the binary relative entropy. 

Fix $\nu = (\nu_1,\dots,\nu_K) \in \cD$ and let $k$ be a sub-optimal arm in $\nu$, that is $c_k^{\alpha} < c^\star$. Assume that there exists $\nu'_k \in \cD_k$ such that $\CVAR_{\alpha}(\nu'_k) > c^\star$ (if this does not hold, $\cK_{\inf}^{\alpha,\cD_k}(\nu_k,c^\star) = +\infty$ and the lower bound holds trivially). Then considering the alternative bandit model $\nu'$ in which $\nu'_i = \nu_i$ for all $i\neq k$ and $\nu'_k$ is defined above, we obtain 
\[\bE_{\pi,\nu}[N_k(T)]\KL(\nu_k, \nu_k') \geq \kl\left(\bE_{\pi,\nu}\left[\frac{N_{k}(T)}{T}\right],\bE_{\pi,\nu'}\left[\frac{N_{k}(T)}{T}\right]\right).\]
Exploiting the fact that the strategy $\pi$ has its CVaR-regret in $o(T^{\beta})$ for any $\beta > 0$, one can prove that, for any $\beta$,  
\[\bE_{\pi,\nu}\left[N_{k}(T)\right] = o(T^{\beta}) \ \ \text{ and } \ \ T - \bE_{\pi,\nu'}\left[N_{k}(T)\right] = o(T^{\beta})\]
since arm $k$ is the (unique) optimal arm under $\nu'$. Using the exact same arguments as \cite{menard_asymp} enables to prove that 
\[\liminf_{T\rightarrow \infty} \frac{\kl\left(\bE_{\pi,\nu}\left[\frac{N_{k}(T)}{T}\right],\bE_{\pi,\nu'}\left[\frac{N_{k}(T)}{T}\right]\right)}{\log(T)} \geq 1,\]
which yields 
\[\liminf_{T\rightarrow \infty} \frac{\bE_{\pi,\nu}[N_k(T)]}{\log(T)} \geq \frac{1}{\KL(\nu_k,\nu_k')}.\]
Taking the infimum over $\nu'_k \in \cD_k$ such that $\CVAR_{\alpha}(\nu'_k) > c^\star$ yields the result, by definition of $\cK_{\inf}^{\alpha,\cD_k}$.

\subsection{Discussion on the scaling of the minimal regret}\label{app::discussion_scaling_lb} Using Lemma A.2 of \citet{tamkindistributionally}, we can show that for any distributions $\nu_F$ and $\nu_G$ with  respective CDFs $F$ and $G$ that are supported in $[0, 1]$, \[|\CVAR_\alpha(F)-\CVAR_\alpha(G)|\leq \frac{1}{\alpha}||F-G||_\infty\;.\]
It follows from Pinsker's inequality that  
$\KL(\nu_F, \nu_G) \geq \alpha^2\left(\CVAR_\alpha(F)-\CVAR_\alpha(G)\right)^2/2$.
Therefore, in a bandit model in which all $\nu_k$ are supported in $[0,1]$ (that is, all $\cD_k$ are equal to $\cP([0,1])$, the set of probability measures on $[0,1]$), it follows that  
\[\kinfcv(\nu_k, c^*)\geq(\alpha \Delta_k^\alpha)^2/2.\]

Combining this inequality together with the lower bound of Theorem~\ref{thm:LB}, we obtain that the regret of an algorithm matching the lower bound is upper bounded by $\cO\left(\sum_{k : c_k^{\alpha} < c^\star}\tfrac{\log(T)}{\alpha^2\Delta_k^\alpha}\right)$, which is precisely the scaling of the CVaR regret bounds obtained for the U-UCB \cite{cassel} and CVaR-UCB \cite{tamkindistributionally}. Assuming the above inequalities are tight for some distributions (which may not be the case), one may qualify these algorithms as "order-optimal", as their CVaR regret makes appear the good scaling in the gaps (and in $\alpha$), just like the UCB1 algorithm \citep{auer2002finite} for $\alpha=1$. 
In this paper we go beyond order-optimality, and we strive to design algorithms that are asymptotically optimal.

\subsection{Lemma~\ref{lem:berge}: continuity of $\kinfcvx$ for multinomial distributions}\label{app::berge}

We state the following result on the continuity of the $\kinfcvx$ functional for multinomial distributions.

\begin{lemma}\label{lem:berge}
	The mapping $\kinfcvx : \cP^{M} \times [x_1,x_M) \mapsto \R$ is continuous in its two arguments. Furthermore, for all $p \in \cP^{M}$ the mapping $c \mapsto \kinfcvx(p, c)$ is increasing on $(C_\alpha(\cX, p), x_M]$.
\end{lemma} 

\begin{proof}
We recall that a multinomial distribution $\nu$ is characterized by its finite support $\cX=(x_1, \dots, x_M)$ and a probability vector $p \in \cP^{M}=\left\{q \in \R^{M}: \forall i, q_i\geq 0, \sum_{j=1}^M q_j =1\right\}$ such that $\bP_{X \sim \nu}(X = x_k) = p_k$ for all $k \in \{1,\dots,M\}$.  We assume that $x_1 \leq x_2 \leq \dots \leq x_M$. 

Moreover, by a slight abuse of notation, we use $\cK_{\inf}^{\alpha,\cX} (p,c)$ as a shorthand for $\cK_{\inf}^{\alpha,\cD_{\cX}} (\nu,c)$ where $\cD_{\cX}$ is the set of multinomial distribution supported on $\cX$. That is, for all $p \in \cP^{M}$ and $c \in [x_1,x_M]$,  
\[\kinfcvx(p, c) = \inf_{q \in \cP^{M}} \left\{\text{KL}(p, q):C_\alpha(\cX, q) \geq c\right\}\,,\]
where $\KL(p, q) = \sum_{i=1}^M p_i \log\left(\frac{p_i}{q_i}\right)$. 

More precisely, we use the Berge's theorem (see, e.g. \cite{berge1997topological}) to prove the continuity of the mapping $\cK_c: p \in \cP^{M} \rightarrow \kinfcvx(p, c)$ for any $c\in (0, x_M)$ and that of $\cK_p: c \in [0, x_M) \rightarrow \kinfcvx(p, c)$ for any $p \in \cP^{M}$. Those functions are of the form 
\[\cK_c(p) = \!\! \inf_{q \in \Gamma(p)} \KL(p,q) \ \ \text{and} \ \ \cK_p(c) = \!\! \inf_{q \in \Gamma(c)} \KL(p,q)\,,\]
where $\Gamma(c) = \Gamma(p) = \{q \in \cP^{M} : C_\alpha(\cX,q) \geq c\}$. As $\KL(p,q)$ is continuous in both arguments and the feasible set is compact, it is sufficient to prove that the correspondences $c \mapsto \Gamma(c)$ and $p \mapsto  \Gamma(p)$ are hemicontinuous, i.e that they are non-empty, lower hemicontinuous and upper hemicontinuous. For $\Gamma(p)$ the  lower and upper hemicontinuity is trivial as the feasible set does not depend on $p$. Moreover, this set is non empty as the Dirac in $x_M$ (represented by $q = (0,\dots,0,1)$) belongs to $\Gamma(p)$, for any $c \in [x_0,x_M]$. 
To conclude, it thus remains to prove that $c \mapsto \Gamma(c)$ is both a lower and upper hemicontinuous correspondence.

We first prove the lower hemicontinuity in every $c_0 \in [x_0,x_M)$. Consider an open set $V \in \cP^{M+1}$ satisfying $V \cap \Gamma(c_0) \neq \emptyset$. We must prove that there exists $\epsilon>0$ such that for all $c\in (c_0-\varepsilon,c_0+\varepsilon)$, $V \cap \Gamma(c) \neq \emptyset$. Since $c \mapsto \Gamma(c)$ is nonincreasing in the sense of the set inclusion, it is sufficient to justify that $V \cap \Gamma(c_0+\epsilon) \neq \emptyset$ for some $\varepsilon>0$. Let $q_0 \in V \cap \Gamma(c_0)$. It holds that $C_\alpha(\cX,q_0)\geq c_0$. If the inequality is strict, there exists $\varepsilon$ such that $q_0 \in V \cap \Gamma(c_0+\epsilon)$. Assume that $C_\alpha(\cX, q_0)=c_0$. As $V$ is open, there exists $\varepsilon'$ such that any $B(q_0,\varepsilon') \subseteq V$. Now there must exist $q\in B(q_0,\varepsilon')$ such that $C_\alpha(\cX, q) > C_\alpha(\cX,q_0)=c_0$. Indeed, in order to construct such a probability vector, we can take out some mass from the components of $q$ corresponding to $x_0$ and assign it to the component corresponding to $x_M$. This increases the \CVAR{} while staying in the ball provided that the change is small enough. Hence, there exists $\varepsilon >0$ such that $q \in V \cap \Gamma(c_0+\varepsilon)$. 

To prove the upper hemicontinuity, we use the following sequential characterization: if $c_n$ is a sequence taking values in $[x_0,x_M)$ that converges to $c$ and $q_n$ is a sequence taking values in $\cP^{M+1}$ that converges to $q$, with $q_n \in \Gamma(c_n)$ for all $n$, one has to prove that $q \in \Gamma(c)$. This fact is a simple consequence of the continuity of $q \mapsto C_\alpha(\cX,q)$ on $\cP^{M+1}$, which is obvious as this function is the supremum of affine functions, as can be seen in Equation~\eqref{eq::cvar_discrete_def}. 

We know prove that the mapping $\cK_p(c)$ is strictly increasing on $(C_\alpha(\cX,p),x_M)$. 
The fact that this mapping is non-decreasing is a simple consequence of the fact that for $c<c'$, $\Gamma(c') \subseteq \Gamma(c)$. 
In order to prove the strict monotonicity it is sufficient to prove that the constraints are binding t the optimum. Assume that this is not the case, i.e. for some $c\in(C_\alpha(\cX,p),x_M)$, $\exists p_c^*: \kinfcvx(p, c) = \text{KL}(p, p_c^*)$ such that $C_\alpha(\cX,p_c^*) = c + \delta$ for some $\delta >0$. By continuity of the \CVAR{}, there exists some $\epsilon>0$ such that $\cB(p_c^*,\epsilon) \subset \Gamma(c+\delta/2) \subset \Gamma(c)$, where $\cB(p_c^*,\epsilon) = \{q \in \cP^{M}: d(q, p_c^*)=||q-p_c^*||_\infty \leq \epsilon\}$. By definition of $p_c^*$ we should have that for any distribution $q\in \cB(p_c^*, \epsilon)$, $\KL(p, q)\geq \KL(p, p_c^*)$. Consider a distribution $\widetilde{p}$ satisfying for some $(i, j)$, $\widetilde{p}_i = p_{c,i}^* + \epsilon$, $\widetilde{p}_j = p_{c,j}^* - \epsilon$
and $\widetilde{p}_\ell = p_{c, \ell}^*$ for $\ell \neq i, j$. Then, it holds that:
\begin{align*}
	\KL(p, p_c^*) - \KL(p, \widetilde{p}) &= p_i \log \left(\frac{p_{c, i}^*+\epsilon}{p_{c, i}^*}\right) + p_j \log \left(\frac{p_{c, j}^*-\epsilon}{p_{c, j}^*}\right)\;. \\ 
\end{align*}	
By a simple Taylor expansion. we have $	\KL(p, p_c^*) - \KL(p, \widetilde{p}) = \epsilon \left(\frac{p_i}{p_{c, i}^*}-\frac{p_j}{p_{c, j}^*}\right) + o(\epsilon^2)$. So, if $\epsilon$ is chosen small enough, the difference has the same sign as $\left(\frac{p_i}{p_{c, i}^*}-\frac{p_j}{p_{c, j}^*}\right)$. Since $p_c^*\neq p$ we are sure to find some coordinates satisfying $p_i>p_{c,i}^*$ and $p_j<p_{c,j}^*$, hence this term can be made positive for an appropriate choice of $(i,j)$. This means that if the constraint is not binding at the optimum we can necessarily find a distribution in the feasible set with a lower $\KL$-divergence with $p$, which is a contradiction. Hence, $\cK_p$ is strictly increasing on $(C_\alpha^\cX, x_M)$.
\end{proof}

\subsection{Proof of Lemma~\ref{lem::kinf_dual_main}: dual form of the $\kinfcv$ of multinomial distributions}

In this section we derive the dual form of the function $\kinfcvx$, where $\cX$ is some finite support $\cX=(x_1, \dots, x_m) \in [0, 1]^M$. We let $\cP^M$ denote he simplex of dimension $M$. We rewrite the optimization problem, defined for any $p \in \cP^M$, $\alpha \in (0,1]$ and $c \in [0,1]$ as
\[\kinfcv(p, c) = \inf_{q \in \cP^M} \left\{\KL(p, q): C_\alpha(\cX, q) \geq c \right\}\;. \]

First of all, we recall that $C_\alpha(\cX, q) = \sup_{x \in \cD} \left\{x-\frac{1}{\alpha}\bE_{X \sim  q}\left((x-X)^+\right)\right\}$. We then introduce the set
\begin{align*}
	\cP_{y, \alpha, c}^M &= \left\{q \in \cP^M: y-\frac{1}{\alpha}\bE_{X \sim  q}((y-X)^+)\geq c \right\} \\
	&= \left\{q \in \cP^M: \bE_{X \sim  q}((y-X)^+)\leq (y-c)\alpha \right\}
	\;.
\end{align*}

Thanks to this definition we can rewrite the problem as

\[\kinfcv(p, c) = \min_{y \in \cD}\left\{ \inf_{q \in \cP^M} \left\{\KL(p, q): y-\frac{1}{\alpha}\bE_{X \sim  q}((y-X)^+)\geq c \right\} \right\} \;, \]

where we used that $\{q: C_\alpha(\cX, q)\geq c\}=\cup_{y \in \cD} \left\{q \in \cP^M: y- \frac{1}{\alpha}\bE_{X\sim q }\left((X-y)^+\right) \geq c\right\}$.

Now, we can first solve the problem $\inf_{q \in \cP_{y, \alpha, c}^M} \KL(p, q) $ for a fixed value of $y$, satisfying $y > c $ (else the feasible set is empty). We write the Lagrangian of this problem:
\[H(q, \lambda_1, \lambda_2) = \sum_{i=1}^M p_i \log \left(\frac{p_i}{q_i}\right) + \lambda_1 \left(\sum_{i=1}^M q_i-1\right) + \lambda_2 \left(\sum_{i=1}^M q_i (y-x_i)^+ - \alpha (y-c)\right) \;,\]

and want to solve $\max_{\lambda_1>0, \lambda_2>0} \min_q H(q, \lambda_1, \lambda_2)$. To this end, we write
\[\frac{\partial H}{\partial q_i}  = -\frac{p_i}{q_i} + \lambda_1 + (y-x_i)^+ \;.\]

Setting the derivative to $0$ yields \[q_i = \frac{p_i}{\lambda_1 + \lambda_2 (y-x_i)^+} \;.\]

We can check that the inequality constraint is achieved. Moreover, exploiting the two constraints leads to $\lambda_1 + \lambda_2 \alpha (y-c) =1 $. This finally gives 
\[q_i = \frac{p_i}{1 - \lambda_2 ((y-c)\alpha - (y-x_i)^+)} \;.\]

Note that this solution is only valid if $\lambda_2 \leq \frac{1}{\alpha(y-c)}$. We have two possibilities: 1) the maximum is achieved in $[0, \frac{1}{\alpha (y-c)})$, in this case we have
\begin{align*}\kinfcv(p, c) = &\inf_{y \in \cD} \max_{\lambda \in \left[0, \frac{1}{\alpha(y-c)}\right)} \sum_{i=1}^M p_i \log\left(1 - \lambda_2 ((y-c)\alpha - (y-x_i)^+) \right) \\ 
	=& \inf_{y \in \cD} \max_{\lambda \in \left[0, \frac{1}{\alpha(y-c)}\right)} \bE_{X \sim p} [\log\left(1 - \lambda_2 ((y-c)\alpha - (y-X)^+)\right)]\;.\end{align*}

The other possibility is that the function is still increasing in $\lambda_2=\frac{1}{\alpha  (y-c)}$. For this case, we check the sign of $\frac{\partial \bE_{X \sim F} [\log\left(1 + \lambda_2 ((y-c)\alpha - (y-X)^+)\right)]}{\partial \lambda}$ at point $\lambda = \frac{1}{\alpha(y-c)}$, that is  of $(y-c)\alpha \left(1-\bE_F\left(\frac{(y-c)\alpha}{(y-X)^+}\right)\right)$. We see that the function can only be increasing if $\bE_F\left(\frac{(y-c)\alpha}{(y-X)^+}\right)<1$, and the solution is then $q_i=\frac{p_i (y-c)\alpha}{y-x_i}$, which provides $\kinfcv(p, c) = \inf_y \bE_F\left(\frac{(y-X)^+}{(y-c)\alpha}\right)$. This concludes the proof.

We remark that these results coincide with those of \citet{HondaTakemura10} with $\alpha=1$ and $y=1$. 
\section{Auxiliary results}\label{app::auxiliary_results}

In this Section we provide some technical tools about the CVaR. In particular, we prove several results that were presented in Section~\ref{sec::analysis}, namely Lemma~\ref{lem::lemma13_main}, Lemma~\ref{lem::lemma14_main}, Lemma~\ref{lem::lem_15_main}, and Lemma~\ref{lem::lower_lem17_main}.

\subsection{Some basic CVaR properties}\label{app::basic_cvar_prop}

In this section we develop some well-known properties of the CVaR. First, the definition of the CVaR as the solution of an optimization problem was first introduced by \citet{rockafellar2000optimization}, to formalize previous heuristic definitions of the CVaR as an average over a certain part of the distribution. The  definition \eqref{eq::cvar_def} is indeed appealing as it applies to any distribution for which $\bE[(x-X)^+]$ is defined, including both discrete and continuous distributions. To understand the CVaR it is particularly useful to look at its expression in these two particular cases. First, for any continuous distribution $\nu$ of CDF $F$ it can be shown (see, e.g. \citet{acerbi2002coherence}) that
\[\CVAR_\alpha(\nu) = \bE_{X \sim \nu}\left[X|X\leq F^{-1}(\alpha) \right].\]

This expression provides a good intuition on what the CVaR represents, as the expectation of the distribution after excluding the best scenarios covering a fraction $(1-\alpha)$ of the total mass. A similar definition exists for real-valued distributions $\nu$ with discrete support $\cX=(x_1,x_2,\dots)$ (either finite or infinite). Assuming that the sequence $(x_i)$ is increasing and letting $p_i = \bP_{X \sim \nu}(X =  x_i)$, one has 
\begin{equation}\label{eq::cvar_discrete_def}\CVAR_\alpha(\nu) = \sup_{x_n \in \cX}\left\{x_n - \frac{1}{\alpha} \sum_{i=1}^{n-1}p_i (x_n-x_i) \right\}\;.\end{equation} 
Indeed, the function to maximize in \eqref{eq::cvar_def} is piece-wise linear, so the maximum is necessarily achieved in a point of discontinuity. In particular, we can easily prove that if $n_\alpha$ is the first index satisfying $\sum_{i=1}^{n_\alpha} p_i \geq \alpha$, then the supremum is achieved in $n_\alpha$ and
\begin{align*} \CVAR_\alpha(\nu) &= x_{n_\alpha} - \frac{1}{\alpha} \sum_{i=1}^{n_\alpha-1}p_i(x_{n_\alpha}-x_i)\\
	&= \frac{1}{\alpha} \left(\sum_{i=1}^{n_\alpha-1}p_i x_i + \left(\alpha - \sum_{i=1}^{n_\alpha-1}p_i\right) x_{n_\alpha}\right)\;. \end{align*}
Hence in that case the CVaR can also be seen as an average when we consider the lower part of the distribution before reaching a total mass $\alpha$.

From the general definition \eqref{eq::cvar_def}, one can also observe that for $\alpha=1$, $\CVAR_\alpha(\nu)=\bE_{X \sim \nu}(X)$. Moreover, the mapping $\alpha \mapsto \CVAR_\alpha(\nu)$ is continuous on $(0,1]$. Thus, considering CVaR bandits allows to smoothly interpolate between classical bandits (that correspond to $\alpha = 1$) and risk-averse problems. 

We also prove in this section a technical result needed in the proof of Theorem~\ref{th::regret_MCVTS} that relates the CVaR of two distributions that are close in terms of the $L^\infty$ distance defined in Appendix~\ref{app::notations}. 

\begin{lemma}[CVaR of two discrete distributions in a $L^\infty$ ball]\label{lem::comp_cvar}
	Let $p$ and $q$ be the probability vectors of two discrete distribution of with shared support $\cX= \{x_1,\dots, x_M\}$, then for any $\alpha \in (0, 1]$ and any $\epsilon>0$:
	\[ C_\alpha(\cX, p) - \frac{M ||p-q||_\infty}{\alpha}x_M \leq C_{\alpha}(\cX, q) \leq C_\alpha(\cX, p) + \frac{M ||p-q||_\infty}{\alpha}x_M\,. \]
\end{lemma}

\begin{proof}
	For any $p \in \cP^M$ and $q \in \cP^M$ we write for simplicity $ \epsilon = \sup_{i \in \{0, 1, \dots, M \}} |p_i - q_i| = ||p-q||_\infty$, so $\forall i$: $q_i - \epsilon \leq p_i \leq q_i + \epsilon$. Let's consider the optimisation problem used to compute the CVaR, $\forall x$:
	\begin{align*}
	x-\frac{1}{\alpha} \sum_{i=0}^M p_i \left(x- x_m \right)^+ & \leq x-\frac{1}{\alpha} \sum_{i=0}^M (q_i - \epsilon) \left(x- x_m \right)^+ \\
	& = x-\frac{1}{\alpha} \sum_{i=0}^M q_i \left(x- x_m \right)^+ + \frac{\epsilon}{\alpha} \sum_{i=0}^M \left(x- x_m \right)^+ \\
	& \leq x-\frac{1}{\alpha} \sum_{i=0}^M q_i \left(x- x_m \right)^+ + \frac{\epsilon}{\alpha} (M+1)x_M\,.
	\end{align*}
	Taking the supremum on all possible values $x$ on the left side of the inequality and then on the right side ensures the result. Then, replacing $p$ by $q$ proves the other inequality.
\end{proof}

\subsection{Proof of Lemma~\ref{lem::lemma13_main}}

In this section we prove Lemma~\ref{lem::lemma13_main}, introduced in Section~\ref{sec::analysis}.

\lemupM*

We recall that the set $\mathcal{Q}_n^M$ is defined as

\[\mathcal{Q}_n^M= \left\{(\beta, p) \in \N^{*n} \times \cP^M: p=\frac{\beta}{n} \right\}\;. \]

The proof relies on Lemma 13 of \cite{Honda} that we re-state below for completeness.

\begin{lemma}[Lemma 13 in \cite{Honda}]\label{lem::right_tail_bound}
	Assume $w \sim \mathrm{Dir}(\beta)$ a Dirichlet distribution over the probability simplex $\cP^{M}$. We assume that $\beta^T 1 = n$ and $\forall j \in \{1, \dots, M\}, \beta_j\geq 0$. We denote by $p=\frac{1}{n}\beta$ the mean of the Dirichlet distribution. Let $S \subset \cP^{M+1}$ a closed convex set included in the probability simplex. The following bound holds:
	\[\bP_{w \sim \mathrm{Dir}(\beta)}\left(w \in S \right) \leq C_1 n^\frac{M}{2} \exp\left(-n \KL\left(p, p^*\right) \right)\,,\]
	
	where $p^*= \text{argmin}_{x \in S} \KL(p,x)$.
\end{lemma}

Using the notation of Section~\ref{sec::analysis}, for any $w \in \cP^{M}$, $\bP(w \in \Sa) \leq \sum_{m=1}^M \bP(w \in \Sm)$.
Then, using Lemma~\ref{lem::right_tail_bound} for each subset $\Sm$, which is closed and convex, we have
\[\bP(w \in \Sa) \leq C_1 n^{M/2} \sum_{m=1}^M \exp\left(-n\KL\left(\frac{\beta}{n}, p_m^*\right) \right)
\;, \]
where $p_m^*= \text{argmin}_{x \in \Sm} \KL\left(\frac{\beta}{n},x\right)$. 

We conclude by using that there exists some $i\in \{1, \dots, M\}$ such that $\kinfcvx\left(\frac{\beta}{n}, c\right) = \KL\left(\frac{\beta}{n}, p_i^*\right)$, and for all $m \neq i$ $\KL(q,p_m^*) \geq \KL(q,p_i^*)$.

\subsection{Proof of Lemma~\ref{lem::lemma14_main}}

We prove the Lemma~\ref{lem::lemma14_main} presented in Section~\ref{sec::analysis}.
\lemlowM*

We follow the sketch of the proof of Lemma 14 of \cite{Honda} using Equation~\eqref{eq::cvar_discrete_def}. We start by stating that there exists some $p^*$ such that $\kinfcvx\left(\frac{\beta}{n}, c\right)=\KL\left(\frac{\beta}{n}, p^*\right)$. The existence of $p^*$ is ensured by the fact that the function $\kinfcvx$ is the solution of the minimization of a continuous function on a compact set.
We consider the set \[\cS_2 = \{w \in \cP^{M+1}: w_i \in [0, p_i^*], \forall j \leq M-1, w_M \geq p_M^*\}.\]
Let us remark that $\forall p \in \cS_2, C_\alpha(\cX, p) \geq C_\alpha(\cX, p^*)\geq c$. Indeed, if we transfer some of the mass from some items of the support to largest items we can only increase the CVaR. It then holds that
\begin{align*}
\bP_{w \sim Dir(\beta)}\left(C_\alpha(\cX, w) \geq c \right) & \geq \bP_{{w \sim Dir(\beta)}}\left(w \in \cS_2\right)\\
&=\frac{\Gamma(n)}{\prod_{i=1}^M \Gamma(\beta_i)} \int_{x \in S_2} \prod_{i=1}^M x_i^{\beta_i-1} dx \\
& \geq \frac{\Gamma(n)}{\prod_{i=1}^M \Gamma(\beta_i)} (p_M^*)^{\beta_M-1} \prod_{j=1}^{M-1} \int_{x_j=0}^{p_j^*}  x_j^{\beta_j-1} dx_j\\
& = \frac{\Gamma(n)}{\prod_{i=1}^M \Gamma(\beta_i)} (p_M^*)^{\beta_M-1} \prod_{j=1}^{M-1} \frac{(p_j^*)^{\beta_j}}{\beta_j} \\
& = \frac{\Gamma(n)}{\prod_{i=1}^M \Gamma(\beta_i)} \frac{\beta_M}{p_M^\star} \prod_{j=1}^{M} \frac{(p_j^*)^{\beta_j}}{\beta_j} \;, \\
\end{align*}
which proves the first inequality. We then exhibit the KL-divergence between two multinomial distributions using
\begin{align*}
\bP_{{w \sim Dir(\beta)}}\left(w \in \cS_2\right) &\geq \frac{\Gamma(n)}{\prod_{i=1}^M \Gamma(\beta_i)} \prod_{j=1}^{M} \frac{(p_j^*)^{\beta_j}}{\beta_j} \\
&= \frac{\Gamma(n)}{\prod_{i=1}^M \Gamma(\beta_i)} \prod_{j=1}^{M}  \left(\frac{p_j^*}{\beta_j}\right)^{\beta_j}\times \prod_{j=1}^{M} \beta_j^{\beta_j-1}\\
&= \frac{\Gamma(n)}{\prod_{i=1}^M \Gamma(\beta_i+1)} \prod_{j=1}^{M}  \left(\frac{p_j^*}{\beta_j/n}\right)^{\beta_j}\times \prod_{j=1}^{M} \left(\frac{\beta_j}{n}\right)^{\beta_j} \\
& = \frac{\Gamma(n)}{\prod_{i=1}^M \Gamma(\beta_i+1)}  \prod_{j=1}^{M} \left(\frac{\beta_j}{n}\right)^{\beta_j} \exp\left(-n\KL\left(\frac{\beta}{n}, p^\star\right)\right) \;.
\end{align*}
This corresponds to the second inequality in the lemma, as we can remark that the terms before the exponential correspond to the desired multinomial distributions, up to a factor $1/n$. We finally provide a lower bound of this quantity using Stirling formula (similarly to the proof scheme of \cite{Honda}),
\[\sqrt{2\pi n}\left(\frac{n}{e}\right)^n \leq n! \leq \sqrt{2\pi n}\left(\frac{n}{e}\right)^n \left(1+C(n)\right) \;,\]
with $C(n)=\frac{1}{12n}+\frac{1}{288n^2}$. We then obtain that
\begin{align*}
\frac{\Gamma(n)}{\prod_{i=1}^M \Gamma(\beta_i+1)}  \prod_{j=1}^{M} \left(\frac{\beta_j}{n}\right)^{\beta_j} &= \frac{1}{n} \frac{n!}{n^n} \prod_{j=1}^M \frac{\beta_j^{\beta_j}}{\beta_j!} \\
&\geq \sqrt{\frac{2\pi}{n}} e^{-n} \times \prod_{j=1}^M  \frac{1}{1+C(\beta_j)} \frac{e^{\beta_j}}{\sqrt{2\pi \beta_j}} \\
&= \sqrt{\frac{2\pi}{n}} \prod_{j=1}^M \frac{1}{(1+C(\beta_j))\sqrt{2\pi \beta_j}}\\
&\geq \sqrt{\frac{2\pi}{n}} e^{\frac{-M}{12}} \prod_{j=1}^M \frac{1}{(1+C(\beta_j))\sqrt{2\pi \beta_j}} \\
& \geq \sqrt{\frac{2\pi}{n}} e^{\frac{-M}{12}} \frac{1}{\sqrt{2\pi}^M} \times \left(\frac{M}{n}\right)^{\frac{M}{2}} \\
& \geq \frac{e^{-\frac{M}{12}}M^\frac{M}{2}}{\sqrt{2\pi}^{M-1}} n^{-\frac{M+1}{2}} \\
& \geq \sqrt{2\pi} \left(\frac{M}{2.13}\right)^\frac{M}{2}\times n^{-\frac{M+1}{2}} \\
& = C_M n^{-\frac{M+1}{2}} \;,
\end{align*}
where we used that $C(n)$ is maximum when $n=1$ and that $1/12 \geq \log(25/288)$, and on the other hand that $\prod_{j=1}^M \beta_j$ is minimized when all $\beta_j$ are equal to $n/M$ (if we allow continuous values).

\subsection{Proof of Lemma~\ref{lem::lem_15_main}}
We prove Lemma~\ref{lem::lem_15_main} that is introduced in Section~\ref{sec::analysis}.

\lemupC*

\begin{proof}
	We first use that $\{q: C_\alpha(\cX, q)\geq c\}=\cup_{y \in [c, B]} \left\{q \in \cP^{n+1}: y- \frac{1}{\alpha}\bE_{X\sim q }\left((X-y)^+\right) \geq c\right\}$ to write	
	\begin{align*}
		\bP_w\left(C_\alpha(\cX, w) \geq c\right) \leq & \bP_w\left(\sup_{y \in [c, B]} \left\{y- \frac{1}{\alpha}\sum_{i=0}^n w_i (y-x_i)^+\right\} \geq c \right) \\
		&\leq \int_{c}^B \bP_w\left(y- \frac{1}{\alpha}\sum_{i=0}^n w_i (y-x_i)^+ \geq c \right) \bP_w\left(y = \text{argsup}_{y \in [c, B]} \left\{y- \frac{1}{\alpha}\sum_{i=0}^n w_i (y-x_i)^+\right\} \right) dy\\
	\end{align*}
	The second term can have an arbitrarily complicated form, but we use the fact that the support is bounded, which enables to uniformly bound it by $1$. We bound the first term by its supremum on $[0, B]$, hence
	\begin{align*}
		\bP_w\left(C_\alpha(\cX, w) \geq c\right) \leq & B \sup_{y \in [c, B]} \bP_w\left(y- \frac{1}{\alpha}\sum_{i=0}^n w_i (y-x_i)^+ \geq c \right) \\
		& \leq B \sup_{y \in [c, B]} \bP_w\left(\alpha(y-c)-\sum_{i=0}^n w_i (y-x_i)^+ \geq 0 \right) \;.\\
	\end{align*}
	
	We then handle $\bP\left(\alpha(y-c)-\sum_{i=0}^n w_i (y-x_i)^+ \geq 0 \right)$ for a fixed value of $y$. We here follow the path of \citet{Honda}, using that a Dirichlet random variable $w=(w_0, \dots, w_n)$ can be written in terms of $n+1$ independent random variables $R_0, \dots, R_n$ following an exponential distribution, as $w_i=\frac{R_i}{\sum_{j=0}^n R_j}$. Using this property and multiplying by $\sum_{j=0}^n R_j$ we obtain
	\begin{align*}
		\bP\left(\alpha(y-c)-\sum_{i=0}^n w_i (y-x_i)^+ \geq 0 \right) \leq & \bP\left( \sum_{i=0}^n R_i\left(\alpha(y-c)- (y-x_i)^+\right) \geq 0 \right) \\
		\leq & \bE\left[\exp\left(t \sum_{i=0}^n R_i\left(\alpha(y-c)- (y-x_i)^+\right) \right)  \right] \;, \\
	\end{align*}
	where we used Markov's inequality for some $t\in \left[0, \frac{1}{(y-c)\alpha}\right)$. We then isolate the first term, writing 
	
	\begin{align*}
		\leq & \prod_{i=0}^n \bE\left[\exp\left(R_i t \left(\alpha(y-c)- (y-x_i)^+\right)\right)\right] \\
		\leq &  \exp\left(-\sum_{i=0}^n \log\left(1-t\left(\alpha(y-c)- (y-x_i)^+\right)\right)\right) \\
		\leq & \frac{1}{1-t\alpha(y-c)} \exp\left(- \sum_{i=1}^n \log\left(1-t\left(\alpha(y-c)- (y-x_i)^+\right)\right)\right) \\ 
		\leq &  \frac{1}{1-t\alpha(y-c)}  \left\{\exp\left(- N \bE_{\widehat{F}}\left[ \log\left(1-t\left(\alpha(y-c)- (y-X)^+\right)\right)\right]\right) \right\} \;.
	\end{align*}
	
	Since the term $1-t\alpha(y-c)$ can be arbitrarily small, we have to control the values of $t$ in order to ensure that the constant before the exponential is not too large. Hence, we choose some constant $\eta>0$, and write that for any $t \in [0, \frac{1-\eta}{\alpha(y-c)}]$ we have
	
	\[\bP_w\left(C_\alpha(\cX, w) \geq c\right) \leq \frac{B}{\eta} \exp\left(- N \bE_{\widehat{F}}\left[ \log\left(1-t\left(\alpha(y-c)- (y-X)^+\right)\right)\right]\right) \;, \]
	
	which leads to
	\begin{align*}
	\bP_w\left(C_\alpha(\cX, w) \geq c\right) \leq& \frac{B}{\eta} \inf_{t \in \left[0, \frac{1-\eta}{(y-c)\alpha}\right]} \exp\left(- N \bE_{\widehat{F}}\left[ \log\left(1-t\left(\alpha(y-c)- (y-X)^+\right)\right)\right]\right) \\
	\leq & \frac{B}{\eta} \exp\left(- N \sup_{t \in \left[0, \frac{1-\eta}{(y-c)\alpha}\right]} \bE_{\widehat{F}}\left[ \log\left(1-t\left(\alpha(y-c)- (y-X)^+\right)\right)\right]\right)\,.
	\end{align*}
	
	At this step the dual form of the function $\kinfcvx(\widehat F)$ start to appear, however, we have to handle the interval on which the supremum is taken is $\left[0, \frac{1-\eta}{\alpha(y-c)}\right]$ instead of $\left[0, \frac{1}{\alpha(y-c)}\right]$. As in \cite{Honda} we will use the concavity and the regularity of the function in the expectation in order to conclude. We write 
	\[ \phi(t) = \frac{1}{n} \sum_{i=1}^n  \log\left(1-t(\alpha(y-c)-(y-x_i)^+)  \right) \;.\]
	As $\phi$ is concave it holds that for any $t \in \left[\frac{1-\eta}{(y-c)\alpha}, \frac{1}{(y-c)\alpha}\right)$ we have
	\[ \phi(t) \leq  \phi\left(\frac{1-\eta}{\alpha (y-c)}\right) + \frac{\eta}{\alpha (y-c)} \phi'\left(\frac{1-\eta}{\alpha(y-c)}\right) \;.\]
	
	At this step we only need to upper bound $\phi'\left(\frac{1-\eta}{\alpha(y-c)}\right)$ by a constant that would not depend on the values of $x_1, \dots, x_n$ and $y$. To do so, we use that all variables are bounded, and for any $t \in \left[\frac{1-\eta}{(y-c)\alpha}, \frac{1}{(y-c)\alpha}\right)$,
	
	\begin{align*}
		\phi'\left(t\right) &= - \bE_{\widehat F}\left[\frac{(y-c)\alpha - (y-X)^+}{1-t\left[(y-c)\alpha - (y-X)^+\right]}\right] \\
		& \leq - \bE_{\widehat F}\left[\frac{(y-c)\alpha - y}{1-t\left[(y-c)\alpha - y\right]}\right] \\
		& =  \frac{(1-\alpha)y + \alpha c}{1-t(1-\alpha)y - t\alpha c} \;.
	\end{align*} 

We then replace $t$ by $\frac{1-\eta}{(y-c)\alpha}$, which gives
\[\frac{\eta}{\alpha(y-c)}\phi'\left(\frac{1-\eta}{(y-c)\alpha}\right) \leq \eta \frac{(1-\alpha) y + \alpha c}{\eta \alpha (y-c)+(1-\eta)y}\;.\]

Then, we use that $(1-\alpha)y +\alpha c \leq B$, and that $\eta \alpha (y-c)+(1-\eta)y \geq (1-\eta)c \geq c$, so finally
\[\frac{\eta}{\alpha(y-c)}\phi'\left(\frac{1-\eta}{(y-c)\alpha}\right) \leq \eta \frac{(1-\alpha)B + \alpha c}{c} \;.\]

Summarizing these steps, we obtain
\[\sup_{t \in \left[0, \frac{1-\eta}{(y-c)\alpha}\right]} \phi(t) \leq  \sup_{t \in \left[0, \frac{1}{(y-c)\alpha}\right]} \phi(t) + \eta \frac{(1-\alpha) B + \alpha c}{c}  \;.\]

Hence, we finally conclude the proof using Lemma~\ref{lem::kinf_dual_main}, to obtain

\[\bP_w\left(C_\alpha(\cX, w) \geq c\right) \leq \frac{B}{\eta}\exp\left(-n \left(\kinfcvx(\widehat F, c) - \eta \frac{(1-\alpha)B+\alpha c}{c} \right)\right)\;. \]

\end{proof}

\subsection{Proof of Lemma~\ref{lem::lower_lem17_main}}\label{sec::lemma17}

In this section we prove Lemma~\ref{lem::lower_lem17_main} introduced in Section~\ref{sec::analysis}.
\lemlowC*

\begin{proof}
	We assume that $\cX$ is known and ordered, i.e $x_1 \leq x_2 \leq \dots \leq x_n$. We then write \[A = \bP_{w \sim \cD_n}\left(C_\alpha(\cX, w) \geq C_\alpha(\cX)\right) \;.\]
	
	Thanks to the definition of the CVaR provided by Equation~\eqref{eq::cvar_def} it holds that
	
	\[A = \bP_w\left(\sup_{y \in \cX}\{y - \frac{1}{\alpha}\sum_{i=1}^n w_i(y-x_i)^+\} \geq \sup_{z \in \cX}\{z - \frac{1}{\alpha n}\sum_{i=1}^n(z-x_i)^+\}\right) \;.\]
	
First, if we know $x_1, \dots, x_n$ then the second term is deterministic and the sup is actually achieved in $x_{\lceil n\alpha \rceil}$. Secondly, the inequality is true if at least one term in the left element satisfies it, so we can write
	
	\begin{align*} A = &\bP\left(\sup_{z \in \cX} \left\{ z - \frac{1}{\alpha}\sum_{i=1}^n w_i(z-x_i)^+\right\} \geq x_{\lceil n\alpha \rceil} - \frac{1}{\alpha n} \sum_{i=1}^n (x_{\lceil n\alpha \rceil}-x_i)^+\right) \\
		\geq & \bP\left(x_{\lceil n\alpha \rceil} - \frac{1}{\alpha}\sum_{i=1}^n w_i(x_{\lceil n\alpha \rceil}-x_i)^+ \geq x_{\lceil n\alpha \rceil} - \frac{1}{\alpha n}\sum_{i=1}^n(x_{\lceil n\alpha \rceil}-x_i)^+\right) \\
		=& \bP\left(\sum_{i=1}^n w_i(x_{\lceil n\alpha \rceil}-x_i)^+ \leq \frac{1}{n}\sum_{i=1}^n(x_{\lceil n\alpha \rceil}-x_i)^+\right) \\
		=& \bP\left(\sum_{i=1}^n w_i\frac{B-(x_{\lceil n\alpha \rceil}-x_i)^+}{B} \geq \frac{1}{n}\sum_{i=1}^n\frac{B-(x_{\lceil n\alpha \rceil}-x_i)^+}{B}\right) \;.
	\end{align*}
	
	As the variable $\frac{B-(x_{\lceil n\alpha \rceil}-x_i)^+}{B}$ belongs to $[0, 1]$ we can apply the lemma 17 of Riou \& Honda and get \[A \geq \frac{1}{25n^2B}\left(B- \frac{1}{n}\sum_{i=1}^n (B-(x_{\lceil n\alpha \rceil}-x_i)^+)\right) = \frac{1}{25n^3 B}\sum_{i=1}^n (x_{\lceil n\alpha \rceil}-x_i)^+\;. \]
	We conclude by simply omitting all the terms except $(x_{\lceil n\alpha \rceil}-x_1)$ in the sum.
\end{proof}

\section{Brown-UCB a.k.a U-UCB}\label{app::brown_ucb}

In this section, we present the instanciation of the U-UCB algorithm of \cite{cassel} for CVaR bandits, and discuss its links with the Brown-UCB idea proposed by \cite{tamkindistributionally}, which propose to build a UCB strategy based on concentration inequalities proposed by \cite{brown_ineq}. 

\subsection{Explicit form of U-UCB}

The U-UCB bonus is written as $f\left(\frac{C \log t}{N_k(t-1)}\right)$ for some constant $C$ and some function $f$ defined as \[f(x)=\max\left\{2b \left(\frac{x}{a}\right)^{1/2}, 2b \left(\frac{x}{a}\right)^{q/2}\right\}.\] 

Following the Definition 3 in \cite{cassel} we can find the values of the constants $a$, $b$ and $q$. The DKW inequality gives $a=1$, while $b$ and $q$ are found as the smallest parameters satisfying the following inequality: \[ |\CVAR_\alpha(F)-\CVAR_{\alpha}(G)| \leq b(||F-G||_\infty + ||F-G||_\infty^q)\]

If the distributions are upper bounded by some constant $U$ then we have from \cite{tamkindistributionally} that it is sufficient to choose $b=\frac{U}{2 \alpha}$ and $q=1$. This yields the following explicit form for the U-UCB stratey:  
\[A_{t+1}^{\text{U-UCB}} = \argmax{k \in [K]} \left[\CVAR_{\alpha}(\widehat{\nu}_k)+ \frac{U}{\alpha}\sqrt{\frac{C \log t}{2 N_k(t)}} \;\right].\]
In our experimental study, with use the constant $C=2$ in the index of U-UCB (and $U=1$ as we consider distributions that are bounded in $[0,1]$). This choice is motivated by the fact that 
\cite{cassel} show that for $C>2$, U-UCB has a logarithmic \emph{proxy regret}. As explained by \cite{tamkindistributionally}, the proxy regret is an upper bound on the CVaR regret, hence U-UCB is guaranteed to have logarithmic CVaR regret in our setting. 

Interestingly, by following an approach suggest by \cite{tamkindistributionally}, we can recover the exact same algorithm as U-UCB, an propose a simple analysis of this algorithm directly in terms of CVaR regret. 

\subsection{Brown-UCB and its analysis}

The authors of \cite{tamkindistributionally} propose to build on concentration inequalities for the empirical CVaR given by \cite{brown_ineq} to derive a UCB strategy in which the index of each arm adds a confidence bonus to the CVaR of its empirical distribution. However, their derivation of this Brown-UCB algorithm is not correct as they use the concentration inequalities originally given by  \cite{brown_ineq} for the \emph{loss} version of the CVaR. We propose a fix in this section, which consists in adapting the Brown inequalities to the \emph{reward} version of the CVaR which we consider in this paper. 

For bounded distributions there is a clear symmetry between the the two definitions of CVaR. In particular, for any distribution $\nu$ supported in $[0,B]$ \[\CVAR_\alpha(\nu) = B-\CVAR_\alpha^{\text{loss}}(B-\nu)\;, \] where $1-\nu$ denotes the distribution of $1-X$ with $X \sim \nu$ and we write respectively \CVAR{} and $\text{\CVAR}^\text{loss}$ the reward and loss version of CVaR. 
\begin{proof}
	\begin{align*}\CVAR_{\alpha}(\nu) &= \sup_{x \in [0,B]} \left\{x - \frac{1}{\alpha}\bE\left((x-X)^+\right)\right\} \\
	&= \sup_{x \in [0,B]} \left\{x - \frac{1}{\alpha}\bE\left((B-X - (B-x))^+\right)\right\}\\
	& = \sup_{y \in [0,1]} \left\{B-y - \frac{1}{\alpha}\bE\left((B-X - y)^+\right)\right\}\\
	& = B - \inf_{y \in [0, 1]} \left\{y + \frac{1}{\alpha}\bE\left((B-X - y)^+\right)\right\}\\
	& =  1 - \CVAR_\alpha^\text{loss}(B-\nu)
	\end{align*}
\end{proof}

\begin{remark} Applying the same trick to $Y=-X$ provide that for a real random variable $X$ then $\CVAR_\alpha(X)=-\CVAR_\alpha^\text{loss}(-X)$.
\end{remark}

This observation easily yield the following concentration inequalities, which are the counterpart of the Brown inequalities for the reward version of the CVaR. 

\begin{lemma}[Brown's inequalities for $\CVAR_{\alpha}$] \label{lem:brown}
If we write $\widehat{c}_n^\alpha $ the CVAR of an empirical distribution from $n$ variables drawn from a distribution $\nu$ supported in $[0,B]$ and $\emph{CVAR}_\alpha(\nu)= c^\alpha$, we have: \begin{align*} \bP(\widehat{c}_n^\alpha \geq c^\alpha + \epsilon) \leq 3 \exp\left(-\frac{\alpha }{5}  \left(\frac{\epsilon}{B} \right)^2 n \right)\\
\bP(\widehat{c}_n^\alpha \leq c^\alpha - \epsilon) \leq \exp\left(-2 \left(\frac{\alpha \epsilon}{B} \right)^2 n \right)
\end{align*}
\end{lemma}

We note that the upper and lower deviation have their probability bounded by a term whose scaling in $\alpha$ is different. By interverting the two inequalities, \cite{tamkindistributionally} proposed a ``Brown-UCB'' algorithm with an confidence bonus scaling in $1/\sqrt{\alpha}$ instead of $1/\alpha$ and obtained a regret bound that was actually contradicting the lower bound of Theorem~\ref{thm:LB}. 

\paragraph{Expression of Brown-UCB} The inequalities in Lemma~\ref{lem:brown} permit to propose a UCB algorithm of the form 
\[A_{t+1}^{\text{Brown-UCB}} = \argmax{k \in [K]} \ \mathrm{UCB}_k(N_k(t), t)\]
where $\mathrm{UCB}_k(n,t) = \widehat{c}_{k,n}^\alpha + \frac{U}{\alpha} \sqrt{\frac{f(t)}{2n}}$, where $\widehat{c}_{k,n}^\alpha$ is the CVaR of level $\alpha$ of the empirical distribution of the $n$ first observations from arm $k$ and $f(t)$ is an increasing function of $t$ that will be specified later in the analysis. Indeed, one can easily check that \[\bP(\UCB_k(t, n) \leq c^\alpha) = \bP\left(\widehat{c}_{k,n}^\alpha \leq c_{k}^\alpha - \frac{U}{\alpha} \sqrt{\frac{f(t)}{2n}}\right) \leq e^{-f(t)},\]
which justifies that fact that $\UCB_k(t, n)$ is an upper confidence bound on the CVaR of arm $k$. 

Interestingly, we observe that for the choice $f(t) = C\log(t)$, Brown-UCB coincides with the U-UCB algorithm. We upper bound below the CVaR regret of Brown-UCB (or U-UCB) for $C>2$, and recover that this regret is indeed logarithmic. 

We analyze Brown-UCB for distributions supported in $[0, U]$ and a threshold function $f(t) = (2+ \varepsilon)\log(t)$ for some $\varepsilon >0$. For every sub-optimal arm $k$, we start with the classical decomposition 
\begin{align*}
\bE\left[N_k(T)\right] &= \sum_{t=0}^{T-1} \bE[\ind(A_{t+1} = k)] \\
& = 1 + \sum_{t=K}^{T-1} \bE[\ind(A_{t+1} = k, \UCB_1(N_1(t),t)\leq c_1^\alpha)] + \sum_{t=K}^{T-1} \bE[\ind(A_{t+1} = k, \UCB_k(N_k(t),t)\geq c_1^\alpha)]\;.
\end{align*}
We analyze separately these two terms. We use a union bound on the values of $N_1(t)$ and the second inequality in Lemma~\ref{lem:brown} to handle the first term:
\begin{align*}
\sum_{t=K}^{T-1} \bE[\ind(A_{t+1} = k, \UCB_1(t)\leq c_1^\alpha)] &\leq \sum_{t=K}^T \sum_{n=1}^t \bP\left(N_1(t)=n, \UCB_1(n,t)\leq c_1^\alpha\right)\\
& \leq \sum_{t=1}^T \sum_{n=1}^t \bP\left(\widehat{c}_{1,n}^\alpha \leq c_1^\alpha - \frac{1}{\alpha} \sqrt{\frac{f(t)}{2n}}\right)\\
& \leq \sum_{t=1}^T t \exp(-f(t))\;.
\end{align*}
With the choice $f(t) = (2+\varepsilon) \log(t)$, we get $\sum_{t=K}^{T-1} \bE[\ind(A_{t+1} = k, \UCB_1(t)\leq c_1^\alpha)]=O(1)$.

To handle the second term, we write the following:
\begin{align*}
\sum_{t=K}^{T-1} \bE[\ind(A_{t+1} = k, \UCB_k(t)\geq c_1^\alpha)] &\leq \sum_{t=K}^{T-1} \sum_{n=1}^t \bE[\ind(A_{t+1} = k, N_k(t)=n, \UCB_k(N_k(t),t)\geq c_1^\alpha)]\\
& \leq \sum_{t=K}^{T-1} \sum_{n=1}^t \bE\left[\ind\left(A_{t+1}=k, N_k(t)=n, \widehat{c}_{k,n}^\alpha \geq c_1^\alpha - \frac{1}{\alpha} \sqrt{\frac{f(t)}{2n}}\right)\right]\\
& \leq \sum_{t=K}^{T-1} \sum_{n=1}^t \bE\left[\ind\left(A_{t+1}=k, N_k(t)=n, \widehat{c}_{k,n}^\alpha \geq c_1^\alpha - \frac{1}{\alpha} \sqrt{\frac{f(T)}{2n}}\right)\right]\\
& \leq \sum_{n=1}^T \bE\left[\ind\left( \widehat{c}_{k,n}^\alpha \geq c_1^\alpha - \frac{1}{\alpha} \sqrt{\frac{f(T)}{2n}}\right) \sum_{t=K}^{T-1} \ind\left(A_{t+1}=k, N_k(t)=n\right) \right]\\
& \leq \sum_{n=1}^T \bE\left[\ind\left( \widehat{c}_{k,n}^\alpha \geq c_1^\alpha - \frac{1}{\alpha} \sqrt{\frac{f(T)}{2n}}\right) \right] \\
& = \sum_{n=1}^T \bP\left( \widehat{c}_{k,n}^\alpha \geq c_k^\alpha + \left(\Delta_k^\alpha-\frac{1}{\alpha} \sqrt{\frac{f(T)}{2n}}\right)\right) 
\end{align*}
Let $\beta>0$ to be chosen later. Letting $n_0(T) = \left\lceil \frac{f(T)}{2 \alpha^2 (1-\beta)^2 (\Delta_k^\alpha)^2} \right\rceil$, we have that for all $n \geq n_0(T)$, \[\left(\Delta_k^\alpha-\frac{1}{\alpha} \sqrt{\frac{f(T)}{2n}}\right) \geq \beta \Delta_k^\alpha.\]
Therefore, using the first inequality in Lemma~\ref{lem:brown},
\begin{align*}
\sum_{t=1}^T \bE[\ind(A_{t+1} = k, \UCB_k(t)\geq c_1^\alpha)] &\leq n_0(T) + \sum_{n=n_0(T)+1}^T \bP\left(\widehat{c}_{k,n}^\alpha \geq c_k^\alpha + \beta \Delta_k^\alpha \right) \\
& \leq n_0(T) + \sum_{n=1}^T 3 \exp\left(-\frac{\alpha}{5}\beta^2 (\Delta_k^\alpha)^2 n\right)\\
& = n_0(T) + O(1).
\end{align*}

Choosing for example $\beta=1-\sqrt{1/2}$ yields that Brown-UCB with $f(t) = (2+\varepsilon)\log(T)$ satisfies 
\[\bE[N_k(T)] \leq \frac{(2+\varepsilon)\log(T)}{\alpha^2 (\Delta_k^\alpha)^2} + O_{\varepsilon}(1).\]
This permits to prove that Brown-UCB has a regret of the same order of magnitude as the CVaR-UCB algorithm proposed by \cite{tamkindistributionally}. 

\section{Complementary Experiments}\label{app::experiments}

In this section we provide a more complete overview of the results of our experiments introduced but not detailed in Section~\ref{sec::experiments}. The first part of this section contains a comprehensive set of experiments on synthetic data in order to illustrate particular properties of the \MCVTS{} and \CVTS{} algorithms. The second part provides additional experiments performed with the DSSAT crop-simulator, that complete the first set of experiments provided in Section~\ref{sec::experiments}.

\subsection{Experiments on synthetic examples}

Before testing the algorithms on a real-world use-case we performed experiments on simulated data in order to illustrate their empirical properties compared to existing UCB-like algorithms. For all the experiments we generally consider $\alpha$ ranging in $\{10\%, 50\%, 90\%\}$.

\subsubsection{Experiments on Multinomial arms} 

\begin{figure}[H]
		\captionsetup{justification=centering}
	\begin{minipage}[c]{0.5\linewidth}
		\begin{figure}[H]
			\centering
			\includegraphics[scale=0.5]{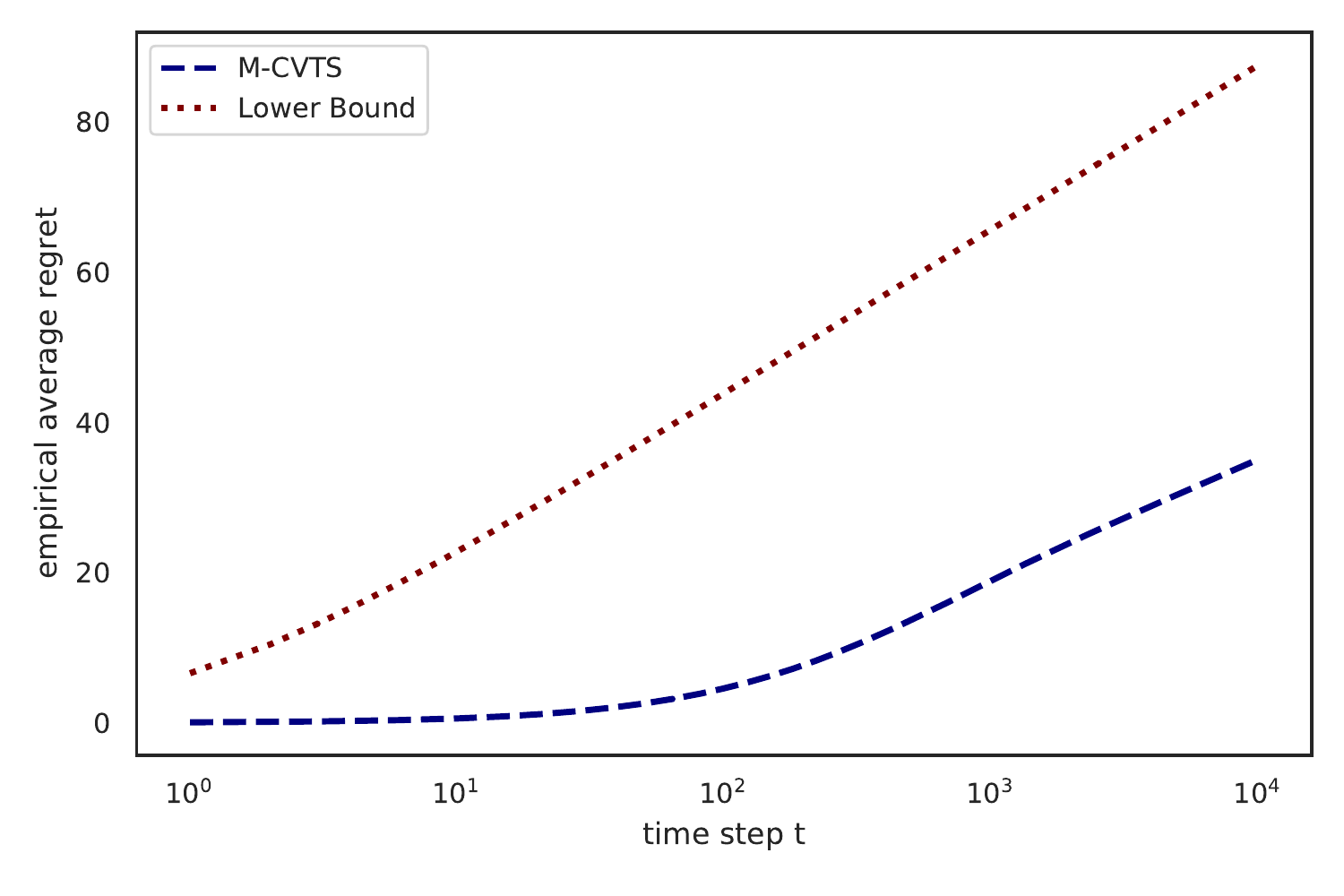}
			\caption{Experiment 1 with Multinomial arms, \\ regret of M-CVTS and lower bound (abscissa log scale),\\ $\alpha=10\%$ \label{fig::xp1_mult10_lb}}	
		\end{figure}
	\end{minipage}
	\hfill
	\begin{minipage}[c]{0.5\linewidth}	
		\begin{figure}[H]
			\centering
			\includegraphics[scale=0.5]{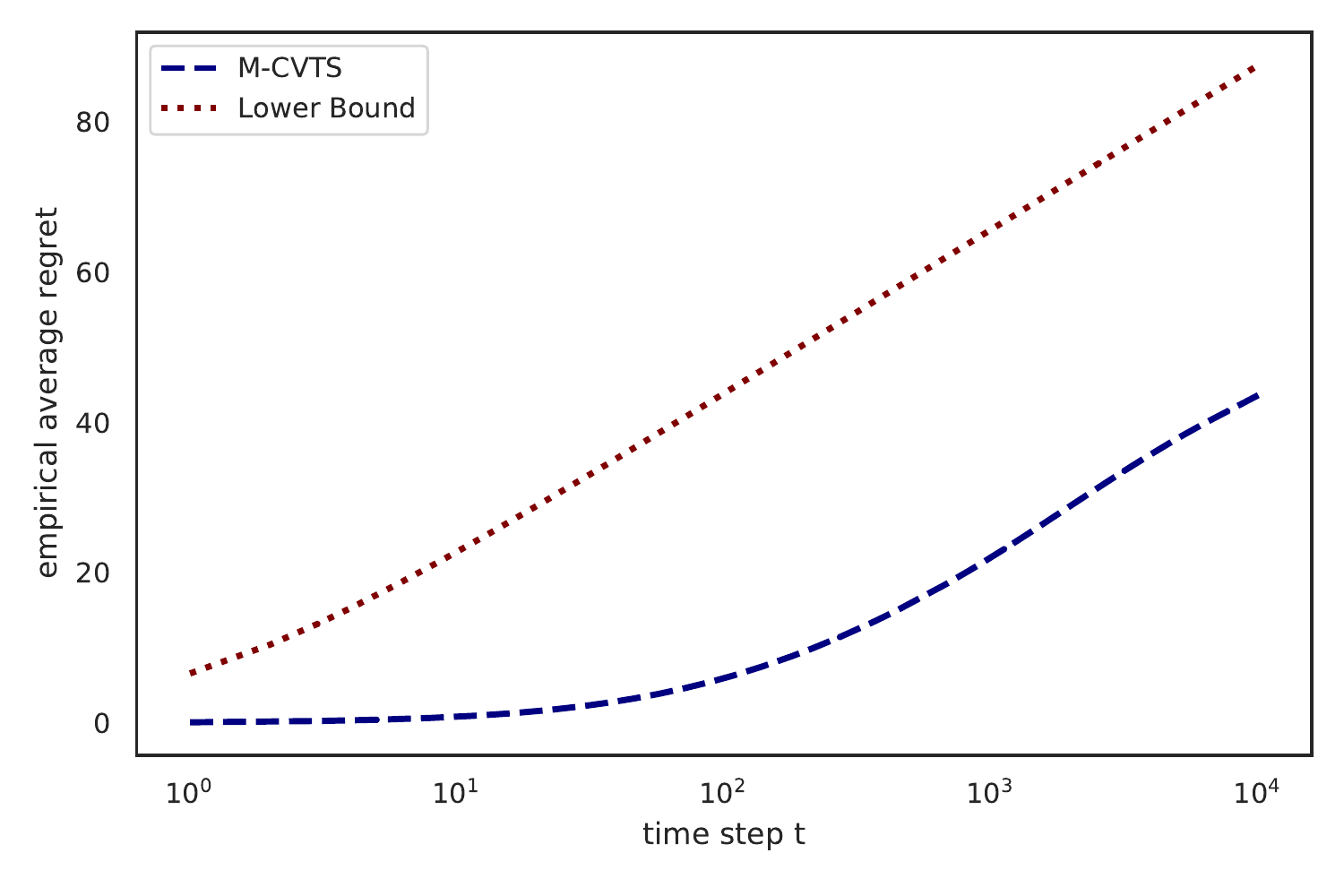}			
			\caption{Experiment 1 with Multinomial arms, \\ regret of M-CVTS and lower bound (abscissa log scale) \\ $\alpha=90\%$ \label{fig::xp1_mult90_lb}}
		\end{figure}	
	\end{minipage}
\end{figure}

We first introduce some experiments with multinomial arms in order to check the empirical performance of the \MCVTS{} algorithm. We ackowledge that \MCVTS~ has some advantage over its competitors as it is aware of the full support of the multinomial distribution while the UCBs only know an upper bound. For this reason we do not comment extensively on the performance gaps between the algorithms, but we are more interested in checking the \textit{asymptotic optimality} of \MCVTS{}. Indeed, for multinomial distribution we implemented the lower bound described in Section~\ref{sec::analysis} and illustrated it in Figures \ref{fig::xp1_mult10_lb} and \ref{fig::xp1_mult90_lb} for one of our experiments.

\begin{figure}[h]
	\centering
	\includegraphics[scale=0.5]{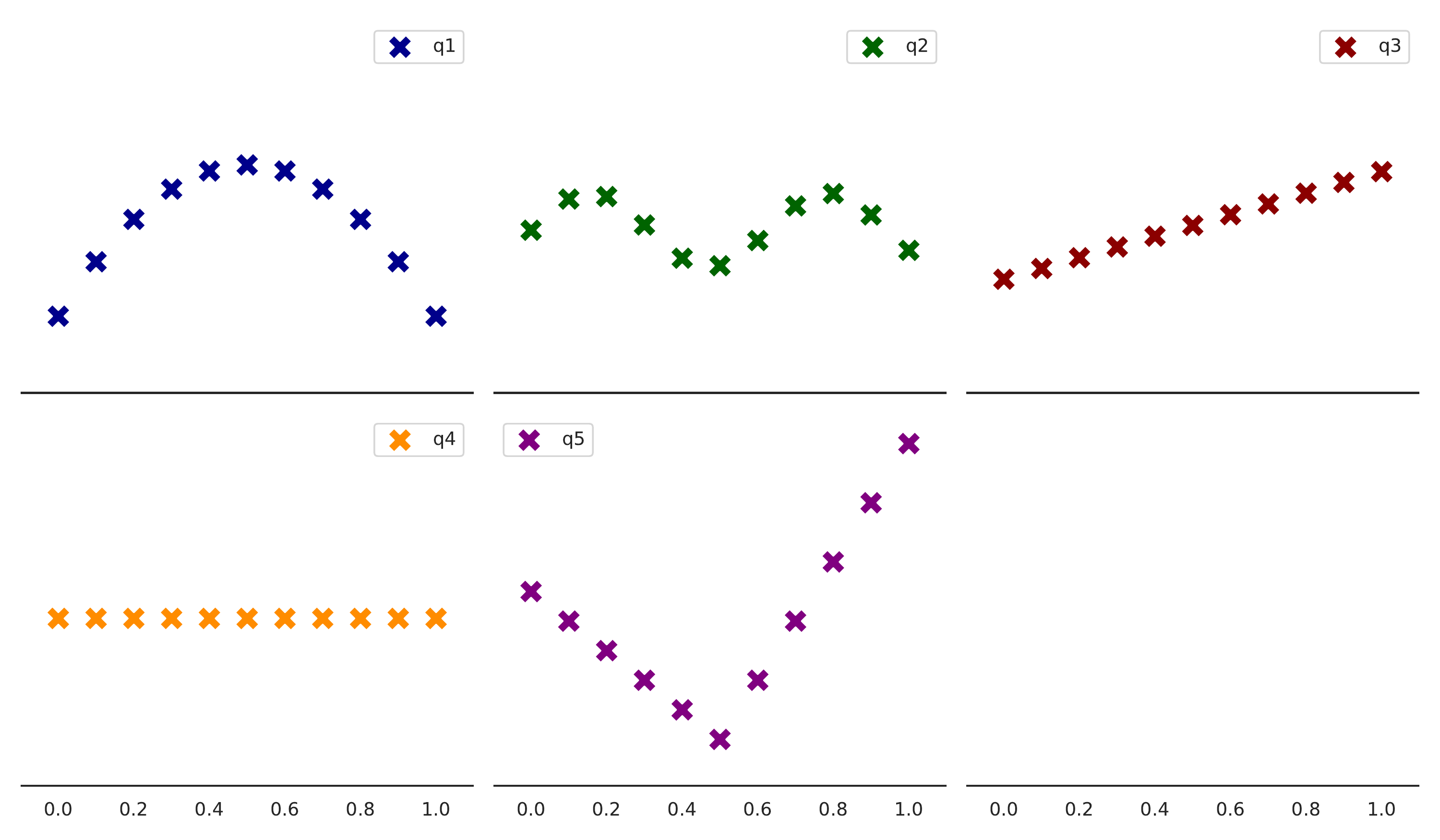}
	\caption{Visual representation of multinomial distributions $q_1, q_2, q_3, q_4, q_5$} \label{fig::mult_visual}
\end{figure}

\paragraph{Multinomial Experiments 1 to 4: Choice of the distributions} We run \MCVTS{} on different multinomial bandit problems in which all arms have the common support $(x_0, x_1,\dots, x_{10})=\{0,0.1, \dots, 0.9, 1\}$. In this setting we consider 5 multinomial distributions, that we write $(q_i)_{i \in \{1, \dots, 5\}}$, and visually represent in Figure~\ref{fig::mult_visual}. Those arms provide different distributions with interesting shapes and properties, using simple formulas to generate the probabilities. 

\begin{figure}[H]
	\centering
	\includegraphics[scale=0.8]{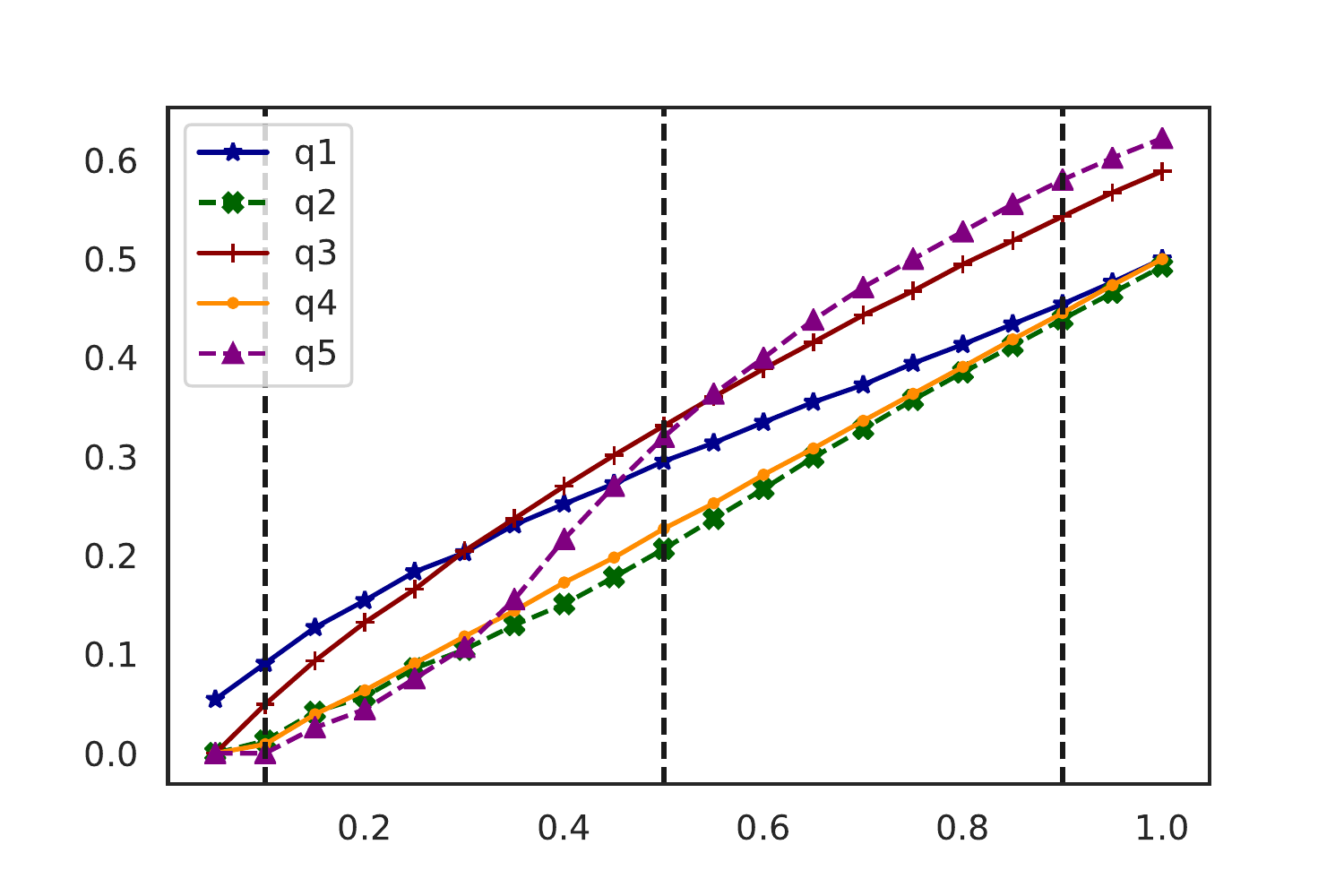}
	\caption{CVaR of the distributions $(q_i)_{i \in \{1, \dots, 5 \}}$  as a function of $\alpha$} vertical lines are the thresholds $\alpha=\{5\%, 50\%, 90\%\}$. \label{fig::cvar_mult_app}
\end{figure}

As the order of arms' CVaR varies substantially depending on the value of $\alpha$, a bandit algorithm aiming at minimizing the CVaR regret is necessary. For instance $q_1$ is the best arm for $\alpha \leq 20\%$, $q_3$ for $\alpha \in [30\%, 55\%]$, and $q_5$ for $\alpha\geq 55\%$. Furthermore, $q_5$ is typically a distribution that a \textit{risk-averse} practitioner would like to avoid as its expectation is large at the cost of potential high losses, while $q_1$ is not satisfying for someone maximizing the expected reward due to the high concentration around $0.5$. Interestingly, despite different shapes $q_2$ and $q_4$ are actually close in terms of CVaR, hence a bandit problem defined over  these two distributions only is hard to solve. We illustrate the CVaRs of these different arms as a function of $\alpha$ in Figure~\ref{fig::cvar_mult_app}.

We implemented four experiments with different subsets of these arms, for $\alpha$ in $\{10\%, 50\%, 90\%\}$: \begin{itemize}
	\item Experiment $1$, $Q_1 = [q_1, q_2, q_3, q_4, q_5]$
	\item Experiment $2$ (Hard for risk-averse learner), $Q_1 = [q_4, q_5]$
	\item Experiment $3$ (Large gaps), $Q_1 = [q_1, q_2, q_3]$
	\item Experiment $4$ (Small gaps), $Q_1 = [q_2, q_4]$
\end{itemize} 

\paragraph{Experimental setup} Each experiment consists in $N=5000$ runs of each algorithm (namely U-UCB, CVaR-UCB and \MCVTS) up to a horizon $T=10000$. 
We report the results in Tables~\ref{tab::xp1_multapp},~\ref{tab::xp2_mult}, ~\ref{tab::xp3_mult} and ~\ref{tab::xp4_mult}.

\paragraph{Analysis of the results} The results reveal that \MCVTS{} clearly outperforms the baselines for any level of $\alpha$ in all experiments. Experiment $4$ is the only one for which the gap is not very large, because this CVaR bandit problem is a hard instance, and no algorithm reaches its asymptotic regime after $10^4$ time steps. However, it is interesting to notice that \MCVTS{} can be better than the baselines even when it is not in this asymptotic regime. 

\begin{figure}[H]
		\captionsetup{justification=centering}
	\begin{minipage}[c]{0.5\linewidth}
		\begin{figure}[H]
			\centering
			\includegraphics[scale=0.5]{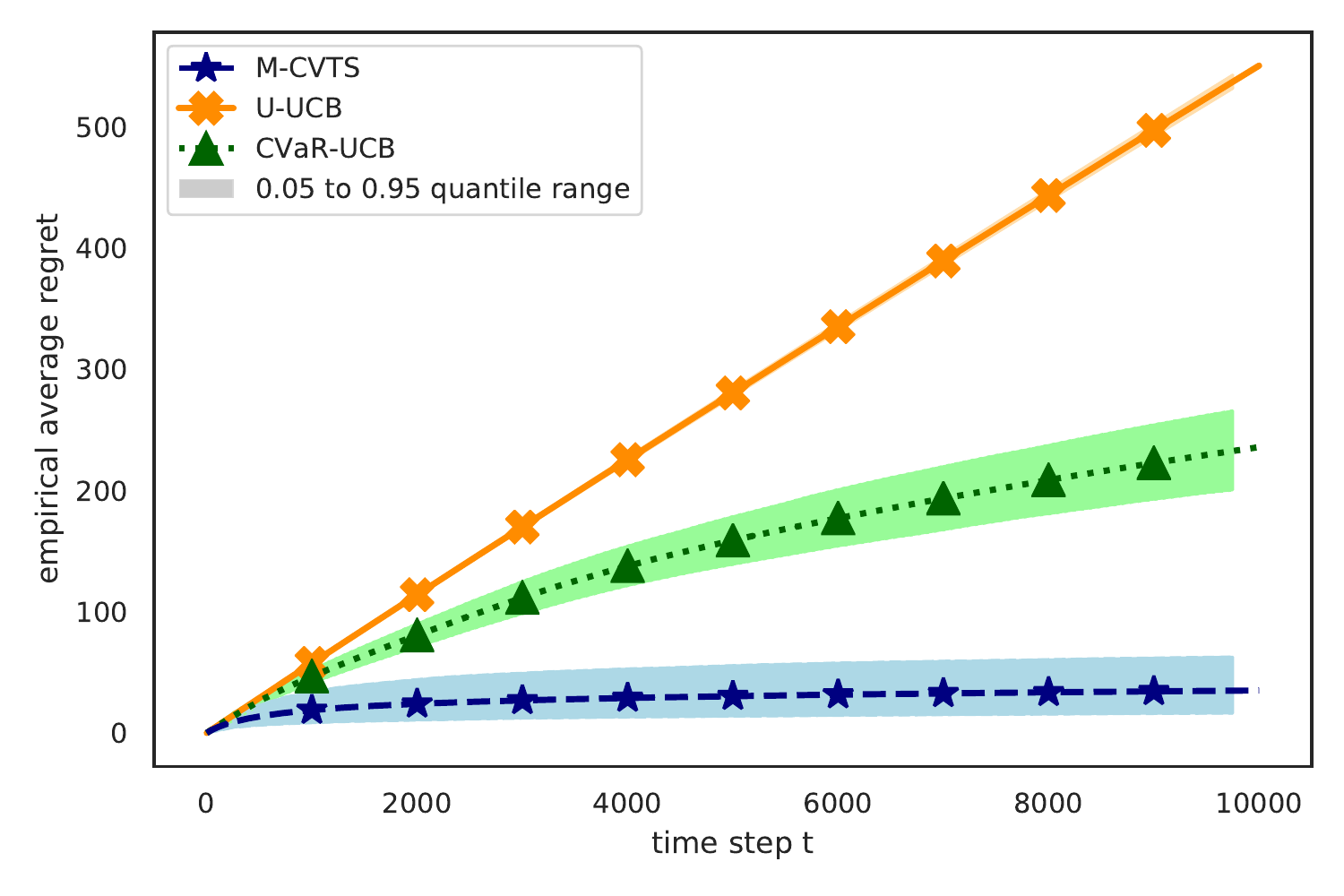}
			\caption{Experiment 1 with Multinomial arms, \\ all algorithms, $\alpha=10\%$ \label{fig::xp1_mult10}}	
		\end{figure}
	\end{minipage}
	\hfill
	\begin{minipage}[c]{0.5\linewidth}	
		\begin{figure}[H]
			\centering
			\includegraphics[scale=0.5]{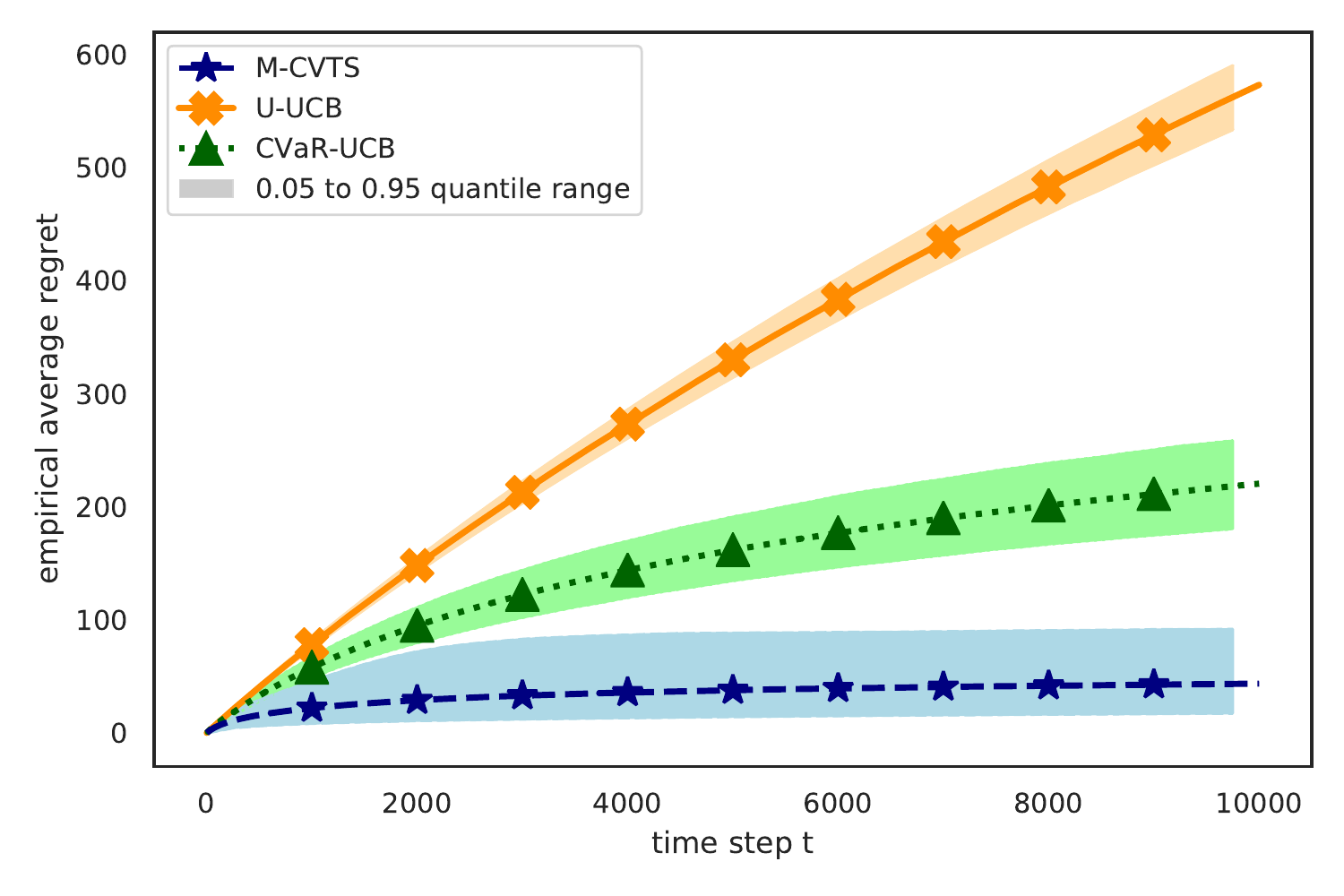}
			\caption{Experiment 1 with Multinomial arms,\\ all algorithms, $\alpha=90\%$ \label{fig::xp1_mult90}}
		\end{figure}	
	\end{minipage}
\end{figure}

For Experiment $1$ and $\alpha \in \{10\%, 90\%\}$ we display the regret curves of both M-CVTS, CVaR-UCB and U-UCB in Figure~\ref{fig::xp1_mult10} and Figure~\ref{fig::xp1_mult90}. We also add a $5\%-95\%$ confidence bound around each curve. We see that the regret of U-UCB is linear for this time horizon with $\alpha=10\%$, while the regrets of CVaR-UCB and M-CVTS have similar shapes and confidence intervals for the two values of $\alpha$, hence they appear to be more robust to the parameter $\alpha$ than U-UCB in this setting. Nonetheless, in both cases M-CVTS largely outperforms CVaR-UCB. It is also interesting to remark that M-CVTS becomes clearly better than its competitors very early in the competition, which shows that M-CVTS can be a good choice for practitioners who would consider shorter horizons.

\paragraph{Optimality of \MCVTS} Still on Experiment $1$, we illustrate in Figures~\ref{fig::xp1_mult10_lb} and ~\ref{fig::xp1_mult90_lb} the \textit{asymptotic optimality} of M-CVTS by representing its regret (in logarithmic scale on the $x$ axis) along with the asymptotic lower bound described in Section~\ref{sec::analysis}, again for $\alpha \in \{10\%, 90\%\}$. The fact that \MCVTS{} matches the asymptotic lower bound is verified in this experiment, as the regret of M-CVTS converges to a straight line which is parallel to the lower bound (still in logarithmic scale on the $x$ axis). The small difference of slopes in Figure~\ref{fig::xp1_mult10_lb} might be due to the fact that we used a solver to solve the optimization problem involved in the lower bound, and we noticed that this solver was less precise for small values of $\alpha$. 

\begin{table}[h]
	\captionsetup{justification=centering}
\begin{minipage}[c]{0.45\linewidth}
		\caption{Results for Multinomial Experiment 1 at $T=10000$ for 5000 replications. Standard deviations in parenthesis. $Q = [q_1, q_2, q_3, q_4, q_5]$} \label{tab::xp1_multapp}
	
		\begin{center}
			\begin{tabular}{llll}
				\textbf{$\alpha$}  & U-UCB & CVaR-UCB & M-CVTS \\
				\hline \\
				$10\%$  & 549.4 (3.3)& 235.9 (19.6)&\textbf{35.1 (14.7)} \\
				$50\%$  & 283.6 (16.3)& 181.5 (17.9) & \textbf{65.4 (30.6)} \\
				$90\%$  & 221.1 (23.7)& 220.5 (23.7) & \textbf{43.7 (36.3)}\\
			\end{tabular}
		\end{center}
	\end{minipage}
	\hfill
	\begin{minipage}[c]{0.45\linewidth}
		\caption{Results for Multinomial Experiment 2 at $T=10000$ for 5000 replications. for 5000 replications. $Q=[q_4, q_5]$} \label{tab::xp2_mult}
		\begin{center}
			\begin{tabular}{llll}
				\textbf{$\alpha$}  & U-UCB & CVaR-UCB & M-CVTS \\
				\hline \\
				$10\%$  & 44.5 (0.4) & 26.7 (5.0) & \textbf{11.9 (7.8)} \\
				$50\%$  & 137.5 (18.9) & 55.4 (13.6) & \textbf{17.7 (23.8)} \\
				$90\%$  & 53.3 (11.0) &  54.3 (11.4) & \textbf{8.0 (5.6)}\\
			\end{tabular}
		\end{center}	
	\end{minipage}
	\vfill
	\vspace{4mm}
	\begin{minipage}[c]{0.5\linewidth}
		\caption{Results for Multinomial Experiment 3 at \\ $T=10000$ for 5000 replications. $Q=[q_1, q_2, q_3]$} \label{tab::xp3_mult}
		\begin{center}
			\begin{tabular}{llll}
				\textbf{$\alpha$}  & U-UCB & CVaR-UCB & M-CVTS \\
				\hline \\
				$10\%$  & 360.1 (3.9)& 149 (16.8)& \textbf{23.0 (13.8)} \\
				$50\%$  & 217.2 (17)& 117.6 (18.7) & \textbf{29.0 (25.6)} \\
				$90\%$  & 124.2 (16.5)& 116.6 (16.0) & \textbf{17.3 (10.8)}\\
			\end{tabular}
		\end{center}
	\end{minipage}
	\hfill
	\begin{minipage}[c]{0.5\linewidth}
		\caption{Results for Multinomial Experiment 4 at \\$T=10000$ for 5000 replications. $Q=[q_2, q_4]$} \label{tab::xp4_mult}
		\begin{center}
			\begin{tabular}{llll}
				\textbf{$\alpha$}  & U-UCB & CVaR-UCB & M-CVTS \\
				\hline \\
				$10\%$  & 17.7 (0.2)& 16.4 (2.2)& \textbf{13.6 (8.6)} \\
				$50\%$  & 79 (7.8)& 68.3 (14.4) & \textbf{27.8 (26.4)} \\
				$90\%$  & 27.1 (4.4)& 26.0 (4.3) & \textbf{21.3 (15.6)}\\
			\end{tabular}
		\end{center}
	\end{minipage}
	\label{fig:ma_fig}
\end{table}

\subsubsection{Experiments on general Bounded distributions: Truncated Gaussian Mixtures} 

In this section we consider bounded multi-modal distributions, built by truncated Gaussian Mixture models in $[0, 1]$. We simply call these distributions \text{Truncated Gaussian Mixtures} (TGM for short). We first remark that these distributions are not continuous because they can have a positive mass in $0$ and $1$, but it is still a good illustrative example to check the performance of \CVTS{}. Indeed, we also performed the same exact  experiments making the distributions continuous (instead of truncating, we re-sampled observations until they lied in $[0, 1]$) and the results were deemed to be exactly the same.

\paragraph{Continuous Experiments 1 to 4: Bi-modal Gaussian mixtures } We first consider experiments with two modes, each mode being equiprobable and having the same variance for simplicity ($\sigma=0.1$) in all experiments. It is interesting to compare settings where some arms have modes that are both close to $0.5$, and where other arms have a large mass of probability close to the two support bounds (one mode close to $1$ and one close to $0$).
 
We experiment 4 possible configurations of the modes, given by:
\begin{itemize}
	\item $\mu_1=(0.2, 0.5)$
	\item $\mu_2=(0, 1)$
	\item $\mu_3=(0.3, 0.6)$
	\item $\mu_4=(0.1, 0.65)$
\end{itemize} 

We indifferently call the arms by their means (saying arm $\mu_1$ for the TGM arm with the parameter $\mu_1$ along with the parameters we fixed).

\begin{figure}[h]
	\centering
	\includegraphics[scale=0.7]{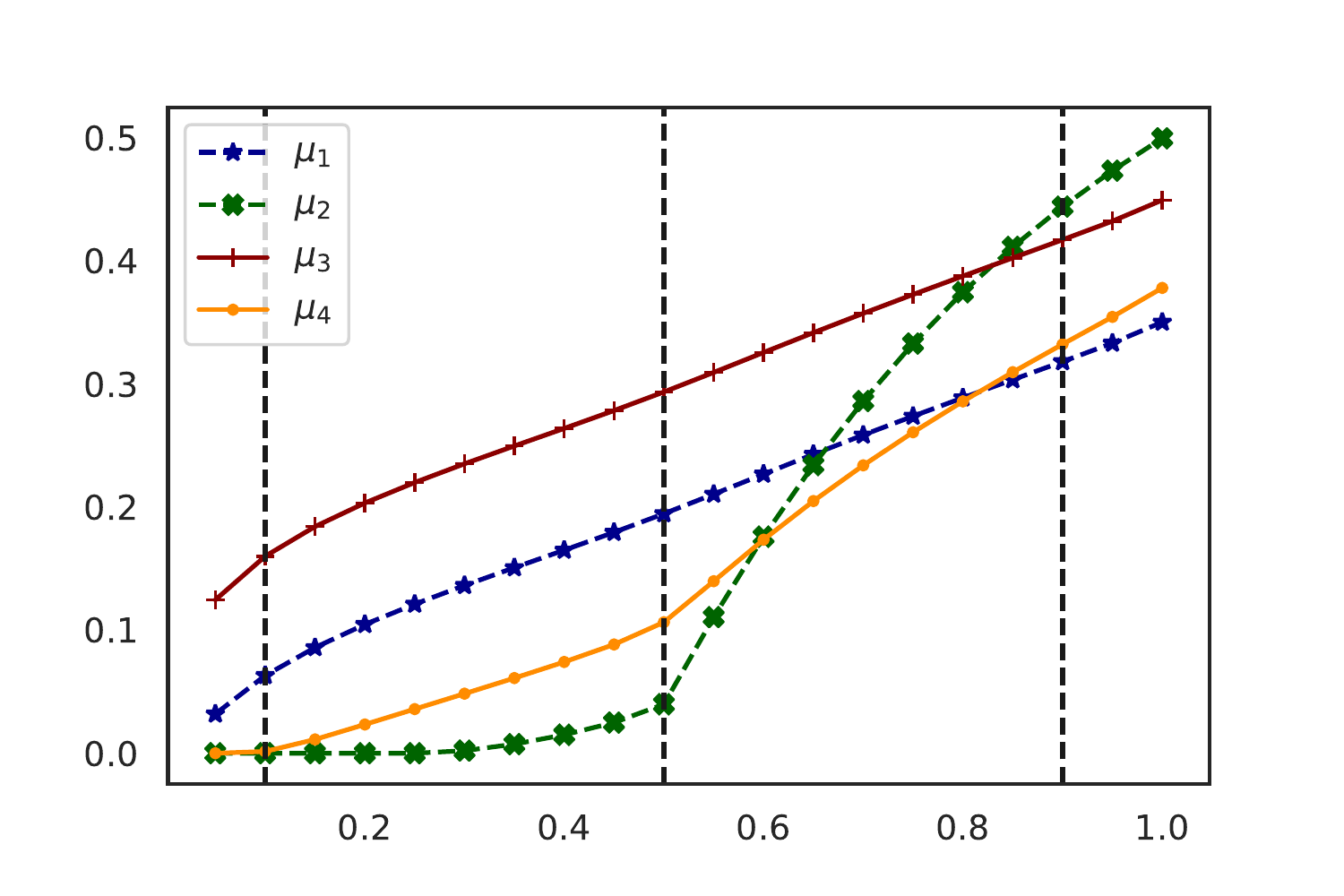}
	\caption{CVaR of each TGM distribution $\nu_i$ (with centers $\mu_i$), $i=1,\dots,4$ for different values of $\alpha$} \label{fig::CVARTBG}	
\end{figure}

As for the discrete setup in the previous section, we highlight some basic properties of their CVaRs: the distribution with parameter $\mu_2$ has a larger mean than the one with $\mu_1$, but the $50\%$ CVaR of $\mu_1$ is larger. We represented the CVaR for each parameter for different values of $\alpha \in (0, 1]$ in Figure~\ref{fig::CVARTBG}, with the thresholds $\alpha \in \{0 \%, 10\%, 90\%\}$ represented by the vertical lines. Interestingly, with these arms the most difficult problems are not necessarily those with smallest values of $\alpha$. Indeed, for $\alpha=80\%$ it may be particularly difficult to choose between $\mu_2$ and $\mu_3$, or between $\mu_1$ and $\mu_4$, while $\mu_3$ is the clear winner for $\alpha=10\%$ due to the distribution being very concentrated around $0.5$. Furthermore, the distribution $\mu_2$ is very concentrated around the bounds of the support but has a larger mean than the others, hence it becomes the best arm for values of $\alpha$ that are close to $1$.

We run the algorithms for $\alpha=10\%, 50\%$ and $90\%$ on four bandit problems with the following respective distributions: 
\begin{itemize}
	\item  $\nu_1=(\mu_1, \mu_2)$
	\item $\nu_2=(\mu_1, \mu_3)$
	\item $\nu_3=(\mu_1, \mu_4)$
	\item  $\nu_4=(\mu_i)_{i \in \{1,2,3,4\}}$
\end{itemize} 

\paragraph{Results} In Tables~\ref{fig::xp1_BBG}, ~\ref{fig::xp2_BBG},~\ref{fig::xp3_BBG} and ~\ref{fig::xp4_TBG} we report the results for the four considered problems (mean regret and standard deviation at $T=10000$). We also provide the regret curves, as for multinomial distributions, in order to check the logarithmic order of the regret of \CVTS{} and its rate when $T$ is large.

\begin{figure}[h]
	\begin{minipage}[c]{0.45\linewidth}
		\begin{table}[H]
			\caption{Results for TGM Experiment 1 at $T=10000$ for 5000 replications. Standard deviations in parenthesis. \label{fig::xp1_BBG}}
			\begin{center}
				\begin{tabular}{llll}
					\textbf{$\alpha$}  & U-UCB & CVaR-UCB & \CVTS \\
					\hline \\
					$10\%$  & 274.9 (1.8)& 5.3 (1.5)& \textbf{1.1 (0.5)} \\
					$50\%$  & 127.0 (19.3)& 135.3 (41.1) & \textbf{29.8 (17.2)} \\
					$90\%$  & 80.5 (10.4)& 53.5 (6.7) & \textbf{10.2 (17.9)}\\
				\end{tabular}
			\end{center}
		\end{table}
	\end{minipage}
	\hfill
	\begin{minipage}[c]{0.45\linewidth}
		\begin{table}[H]
			\begin{center}
				\caption{Results for TGM Experiment 2 at $T=10000$ for 5000 replications. \label{fig::xp2_BBG}}
				\begin{tabular}{llll}
					\textbf{$\alpha$}  & U-UCB & CVaR-UCB & \CVTS \\
					\hline \\
					$10\%$  & 373.7 (4.1)& 72.8 (9.6)& \textbf{4.1 (2.3)} \\
					$50\%$  & 135.8 (8.9)& 37.9 (7.5) & \textbf{5.5 (2.7)} \\
					$90\%$  & 62.6 (7.1)& 43.9 (5.1) & \textbf{5.0 (1.8)}\\
				\end{tabular}
			\end{center}
		\end{table}
	\end{minipage}
	\vfill
	\begin{minipage}[c]{0.45\linewidth}
		\begin{table}[H]
			\caption{Results for TGM Experiment 3 at $T=10000$ for 5000 replications. \label{fig::xp3_BBG}}
			\begin{center}
				\begin{tabular}{llll}
					\textbf{$\alpha$}  & U-UCB & CVaR-UCB & \CVTS \\
					\hline \\
					$10\%$  & 269.4 (1.8)& 23.2 (4.8)& \textbf{2.8 (1.5)} \\
					$50\%$  & 138.5 (12.4)& 71.8 (19.0) & \textbf{14.7 (8.3)} \\
					$90\%$  & 53.1 (6.6)& 34.5 (6.6) & \textbf{20.2 (22.4)}\\
				\end{tabular}
			\end{center}
		\end{table}
	\end{minipage}
	\hfill
	\begin{minipage}[c]{0.45\linewidth}
		\begin{table}[H]
			\caption{Results for TGM Experiment 4 at $T=10000$ for 5000 replications. \label{fig::xp4_TBG}}
			\begin{center}
				\begin{tabular}{llll}
					\textbf{$\alpha$}  & U-UCB & CVaR-UCB & \CVTS \\
					\hline \\
					$10\%$  & 958.9 (4.8)& 230.5 (25.3)& \textbf{10.4 (3.2)} \\
					$50\%$  & 318.4 (12.2)& 147.7 (17.9) & \textbf{21.2 (6.4)} \\
					$90\%$  & 154.3 (11.9)& 119.5 (11.7) & \textbf{25.1 (14.1)}\\
				\end{tabular}
			\end{center}
		\end{table}
	\end{minipage}
\end{figure}

Again, the TS approach significantly outperforms the two UCB algorithms, which is a very interesting result: contrarily to the multinomial case, this time the three algorithms had the same level of information on arms' distributions. \CVTS{} is consistently the best for all four problems we implemented and for all $\alpha$ levels. 

\paragraph{Continuous Experiment 5: Robustness to small $\alpha$} We then check the robustness of \CVTS{} to a smaller value of the parameter $\alpha$ by setting $\alpha=1\%$, referred as Experiment 5. The bandit of Experiment 5 has six TGM arms with respective mean and variance parameters $\mu_{135}=(0.3, 0.6)$, $\mu_{246}=(0.25, 0.65)$, $\sigma_{12}=0.05$, $\sigma_{34}=0.06$, $\sigma_{56}=0.07$. This experiment allows to additionally check if adding different variances to the arms affects the performance of the algorithms. However, we keep the probability of each mode to $0.5$. This problem provides the following CVaR values for each arm at level $1 \%$, respectively:  $c^{0.01}_{1:6} = [0.18, 0.13, 0.15, 0.10, 0.13, 0.08]$. The results are reported in Table~\ref{tab::xp_small_alpha}, in which we  observe a very large performance gap between \CVTS{} and UCB algorithms. This is particularly interesting because it shows that the UCB algorithms are not really able to learn for very small values of $\alpha$ (indeed $\alpha=1\%$ is very small when drawing only a total number of $10^4$ observations) before the horizon becomes extremely large. We already observed this behavior for CVaR-UCB in previous experiments, but this time we can see as well that its average regret is even higher than the one of U-UCB, and its variance spiked. On the other hand, \CVTS{} seems to learn smoothly even for $\alpha=1\%$, as its average regret only doubles between $T=1000$ and $T=5000$, and increases even less  between $T=5000$ and $T=10000$.

\begin{table}[h]
	\caption{Results for TGM Experiment 5  ($\alpha=1\%$) at $T=10000$ for 5000 replications.} \label{tab::xp_small_alpha}
	\begin{center}
		\begin{tabular}{llll}
			T  &U-UCB & CVaR-UCB & \CVTS \\
			\hline \\
			1000  & 49.1 (0.3)& 53.2 (5.6)& \textbf{18 (37)}  \\
			5000  & 245 (1.1)& 263.2 (24.7) & \textbf{35.5 (51)} \\
			10000  & 489.1 (2.2)& 518.4 (45.0) & \textbf{41 (66)}\\
		\end{tabular}
		\vspace{0.2cm}
	\end{center}
\end{table}

\begin{table}[h]
	\centering
	\caption{Results for TGM Experiment 6, at $T=10000$ averaged over 400 random instances with $K=30$ truncated Gaussian mixtures with 10 modes.} \label{tab::xp_MTG}
	\begin{center}
		\begin{tabular}{llll}
			T  &U-UCB & CVaR-UCB & \CVTS \\
			\hline \\
			10000  & 2149.9 (263)& 2016.0 (265)& \textbf{210.9 (6.4)}  \\
			20000  & 4276.4 (538)& 3781.3 (521) & \textbf{237.1 (15.4)} \\
			40000  & 8493.4 (1085)& 6894.1 (985) & \textbf{263.5 (17.9)}\\
		\end{tabular}
	\end{center}
\end{table}

\paragraph{Continuous Experiment 6: Random Problems with more modes and more arms} Finally, we further check the robustness of \CVTS{} to more arms and more diverse distribution profiles by increasing the number of possible modes. To do so, we implement an experiment with $K=30$ arms, with TGM distributions with 10 modes exhibiting different means and variances, which covers a large variety of shapes of distributions. All of those parameters are drawn uniformly at random, and we summarize their distributions as $(\mu, \sigma)\sim \mathcal{U}([0.25, 1]^{10} \times [0, 0.1]^{10})$, and $p \sim \cD_{10}$ (uniform distribution on the simplex, presented in Section~\ref{sec::analysis}). We name this setting TGM Experiment 6. The results of this experiment are reported in Table~\ref{tab::xp_MTG} for a parameter $\alpha=0.05$ averaged over 400 random instances. Again, we choose a smaller value for $\alpha$ than in the previous extensive sets of experiments because problems with small $\alpha$ seem to be more challenging. The results highlight that best performances are obtained by \CVTS{}.

\paragraph{Conclusions} We preliminary evaluated the CVaR bandit algorithms on synthetic problems before testing them on realistic-world bandit environment in the next section. These experiments seem to highlight a greater robustness of \CVTS{} to many different settings regarding different parameters: $\alpha$ level, the number of arms $K$ and the different possible shapes of the distributions (symbolized by the number of modes in our synthetic experiments). In particular, \CVTS{} is the only algorithm that has not shown to be affected by the value of $\alpha$, as the two UCB algorithms had their respective performances degraded in some extent depending on $\alpha$ values.

\subsection{Experiments with \texttt{DSSAT} crop-model}
In this section we keep comparing ~\CVTS~ with U-UCB and CVaR-UCB for $\alpha \in \{5\%, 10\%,80\%\}$ as described in section~\ref{sec::experiments}. \texttt{DSSAT} is still parameterized with the same challenging conditions, but we generate two different problems thanks to the crop-simulator. For both presented experiments we consider $N=1040$ runs of each algorithm up to a time horizon $T=10000$. As explained in section~\ref{sec::experiments}, all \texttt{DSSAT} arms' distributions are empirically estimated from $10^6$ samples in both experiments.

\paragraph{\texttt{DSSAT} Experiment 1: 7 armed planting date bandit} We consider a bandit instance that consists of 7 arms, each arm corresponds to a planting date spaced of 15 days from the previous one. An illustration of the underlying distributions is given in Figure~\ref{fig::E1distPlot}. In this case, the best arm is consistent with all values of $\alpha$, as shown in Table~\ref{tab::E1distTab}. Nevertheless, arms exhibit different gaps when considering different values of $\alpha$. This experiment intends to evaluate \CVTS~robustness for a greater number of real-world alike arms with a diversity of reward distribution shapes.

The results of this experiment are reported in Table~\ref{tab::E1resTab}. The regret curves for the three algorithms, with considered values of $\alpha$ parameter are illustrated in Figures~\ref{fig::E1a5p},~\ref{fig::E1a20p}, and ~\ref{fig::E1a80p}. 

In this experiment, by exhibiting superior performances ~\CVTS~ appears to be more robust than the UCB CVaR bandit algorithms relative to an increase in the number of arms. In practice for the planting-date problem, a global, few months planting-window is known but needs further refinements e.g. to identify the best two-week time slot for planting. That is to say, the number of arms is unlikely to be greater that what has been tested in this experiment, making \CVTS{} a particularly fit-for-purpose candidate in this setup.

\paragraph{\texttt{DSSAT} Experiment 2: Impact of support upper bound over-estimation} This configuration is the same than the one presented in Section ~\ref{sec::experiments}, but here we largely over-estimate the yield upper-bound to 30 t/ha, when a close to reality yield upper bound is about 10 t/ha. From an agronomic point of view, this yield value is a very unlikely over-estimation in the given conditions. This experiment intends to empirically evaluate how a rough arms' upper-bound estimation affects algorithms' performances, when little expert knowledge is available.  An illustration of the underlying distributions and how the upper-bound estimation is exaggerated is given in Figure~\ref{fig::E2distPlot} and corresponding metrics are reported in Table~\ref{tab::E2distTab}.

We provide the results of this experiment in Table~\ref{tab::E2resTab}, and display the regret curves in Figures~\ref{fig::E2a5p},~\ref{fig::E2a20p}, and ~\ref{fig::E2a80p}.

Experiment 2 addresses one possible concern for practitioners: the prerequisite of rewards' support upper bound. We empirically demonstrate that with realistic simulations, when a highly over-estimated, unrealistic  support upper-bound is given -- triple of expert's estimation --,~\CVTS~ keeps outperforming UCB-like CVaR bandit algorithms. We shown that this over-estimation did not affect ~\CVTS~ performances compared to the situation of correct support upper-bound identification as presented in Section~\ref{sec::experiments}. In particular, it even slightly improved its performance for $\alpha = 80\%$. This result is counter-intuitive, but it can be explained by the fact that the extra exploration induced by the larger upper bound may have sped up learning in this particular case, improving overall performances. On the other hand, CVaR-UCB seems much more impacted by this over-estimation (regret is respectively increased by about $150\%$, $75\%$ and $78\%$ for $\alpha \in \{5\%, 20\%,80\%\}$). Similarly U-UCB shown altered performances, despite its already unsatisfying results when considering the true upper bound.

\paragraph{Conclusions}\CVTS~ appeared to be a satisfying candidate for real-world alike problems, as shown with the planting date bandits. We empirically showed the ~\CVTS~ was best able to deal with a greater number of planting date arms than its UCB counterparts. We showed as well that \CVTS~ remained the best performer despite considering a very unlikely support upper-bound estimation. We think that in many physical resource-based problems, this should be reassuring for practitioners, in particular when compared with UCB algorithms' sensibility to the input upper bound.

{\setlength{\belowcaptionskip}{-3em}
	\hfill
	\begin{figure}[H]
		\begin{minipage}[c]{0.45\linewidth}
			\begin{figure}[H]
				\centering
				\includegraphics[width=\linewidth]{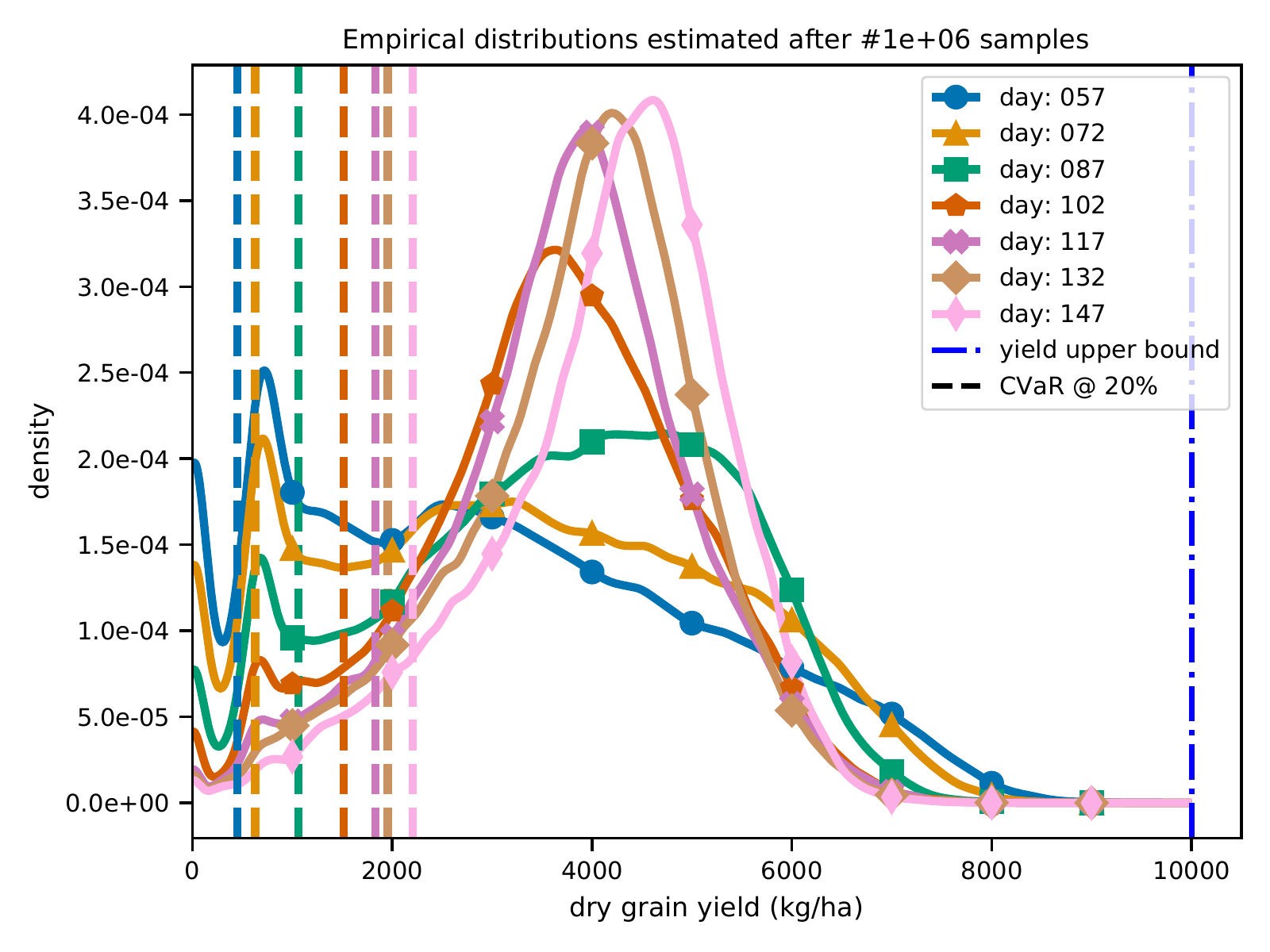}
				\vskip -5mm
				\caption{Experiment 1, 7 armed \texttt{DSSAT} environment empirical distributions ; $10^6$ samples. \label{fig::E1distPlot}}
			\end{figure}
		\end{minipage}
		\hfill
		\begin{minipage}[c]{0.45\linewidth}	
			\begin{figure}[H]
				\centering
				\includegraphics[width=\linewidth]{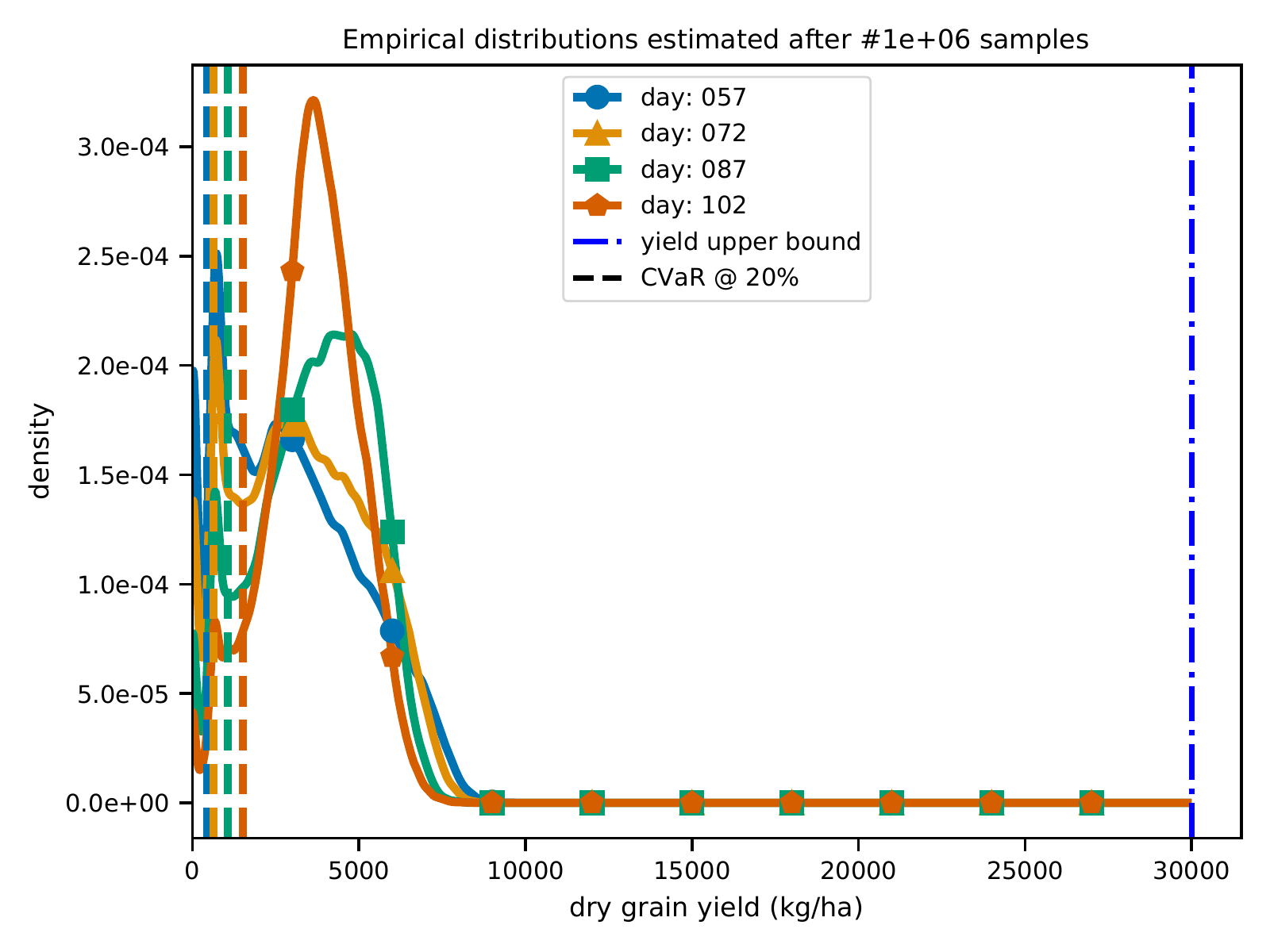}
				\vskip -5mm
				\caption{Experiment 2, 4 armed \texttt{DSSAT} environment empirical distributions with over-estimated support ; $10^6$ samples. \label{fig::E2distPlot}}
			\end{figure}
		\end{minipage}
		\hfill
	\end{figure}
	\hfill
	\begin{table}[h!]
		\begin{minipage}[c]{0.45\linewidth}
			\caption{\texttt{DSSAT} Experiment 1 distribution metrics in kg/ha estimated from $10^6$ samples.. \label{tab::E1distTab}}
			\begin{center}
				\begin{tabular}{lllll}
					day (action) & & & $\text{CVaR}_{\alpha}$ & \\ \hline
					& $5\%$ & $20\%$ & $80\%$ & $100\%$ (mean) \\ \cmidrule{2-5}
					057  & 0 & 448 & 2238 & 3016\\
					072  & 46 & 627 & 2570 & 3273\\
					087  & 287 & 1059 & 3074 & 3629\\
					102  & 538 & 1515 & 3120 & 3586\\
					117  & 808 & 1832 & 3299 & 3716\\
					132  & 929 & 1955 & 3464 & 3850\\
					147  & \textbf{1122} & \textbf{2203} & \textbf{3745} & \textbf{4112}\\
				\end{tabular}
			\end{center}
		\end{minipage}
		\hfill
		\begin{minipage}[c]{0.45\linewidth}
			\vspace*{-12mm}
			\caption{\texttt{DSSAT} Experiment 2 distribution metrics in kg/ha estimated from $10^6$ samples.\label{tab::E2distTab}}		
			\begin{center}
				\begin{tabular}{lllll}
					day (action)  & & & $\text{CVaR}_{\alpha}$ & \\ \hline
					& $5\%$ & $20\%$ & $80\%$ & $100\%$ (mean) \\ \cmidrule{2-5}
					057  & 0 & 448 & 2238 & 3016\\
					072  & 46 & 627 & 2570 & 3273\\
					087  & 287 & 1059 & 3074 & \textbf{3629}\\
					102  & \textbf{538} & \textbf{1515} & \textbf{3120} & 3586\\
				\end{tabular}
			\end{center}
		\end{minipage}
		\hfill
	\end{table}
	\hfill
	\begin{table}[h!]
		\begin{minipage}[c]{0.45\linewidth}
			\caption{Results for \texttt{DSSAT} Experiment 1, empirical regret at $T=10000$ in t/ha for 1040 replications. Standard deviations in parenthesis. \label{tab::E1resTab}}
			\begin{center}
				\begin{tabular}{llll}
					\textbf{$\alpha$}  & U-UCB & CVaR-UCB & \CVTS \\
					\hline \\
					$5\%$  & 5687 (5)& 1891 (18)& \textbf{700 (22)} \\
					$20\%$  & 6445 (10)& 1795 (19) & \textbf{489 (17)} \\
					$80\%$  & 3367 (14)& 1580 (15) & \textbf{293 (8)}\\
				\end{tabular}
			\end{center}	
		\end{minipage}
		\hfill
		\begin{minipage}[c]{0.45\linewidth}
			\caption{Results for \texttt{DSSAT} Experiment 2, empirical regret at $T=10000$ in t/ha for 1040 replications. \label{tab::E2resTab}}
			\vspace*{4mm}
			\begin{center}
				\begin{tabular}{llll}
					\textbf{$\alpha$}  & U-UCB & CVaR-UCB & \CVTS \\
					\hline \\
					$5\%$  & 3179 (2)& 759 (14)& \textbf{195 (11)} \\
					$20\%$  & 5644 (6)& 1020 (17) & \textbf{202 (10)} \\
					$80\%$  & 2642 (10)& 888 (13) & \textbf{284 (12)}\\
				\end{tabular}
			\end{center}
		\end{minipage}
		\hfill
	\end{table}
	\hfill
	\begin{figure}[H]
		\captionsetup{justification=centering}
		\begin{minipage}[c]{0.5\linewidth}
			\begin{figure}[H]
				\centering
				\includegraphics[height=.25\textheight]{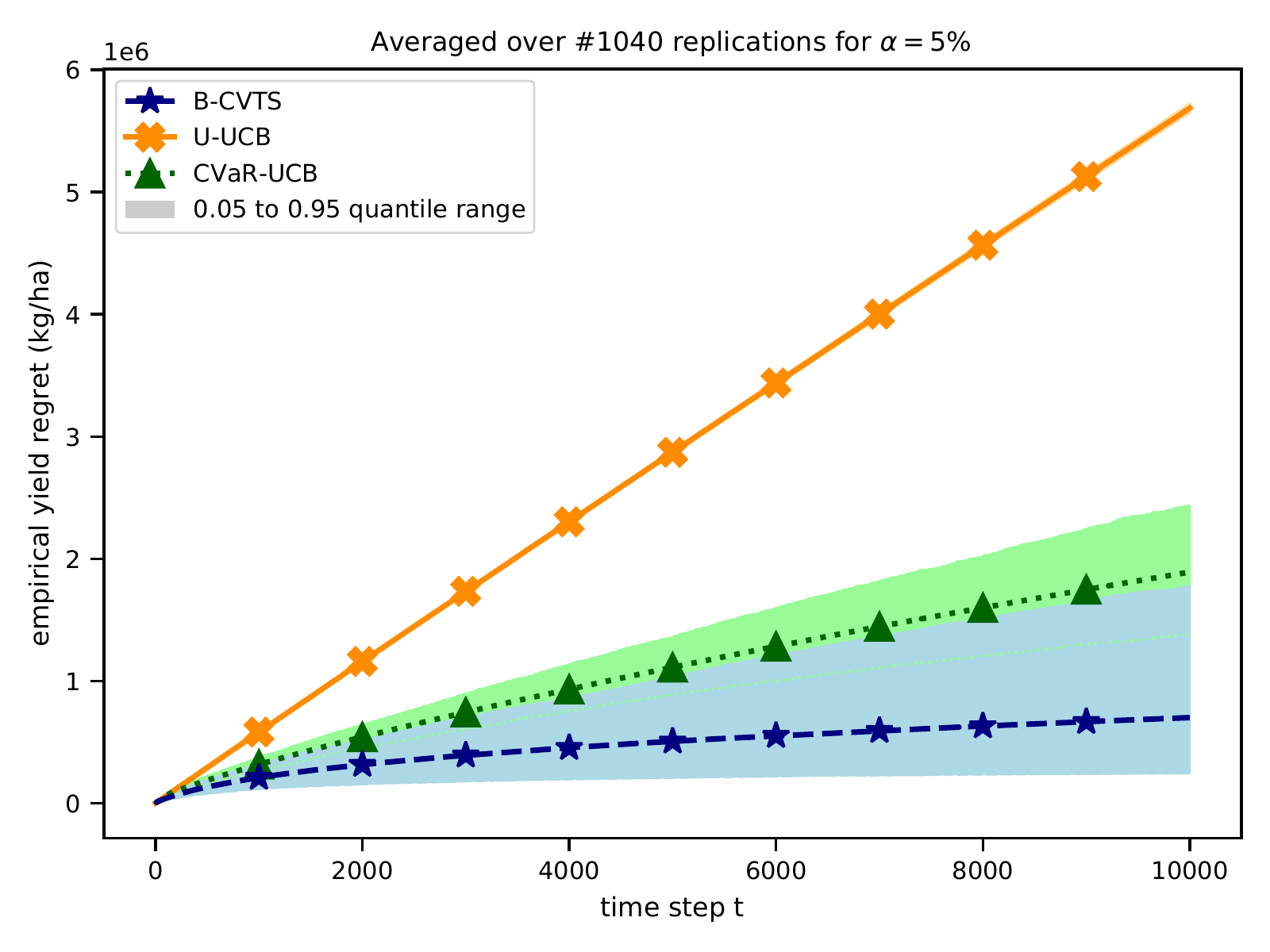}
				\vskip -5mm
				\caption{\texttt{DSSAT} Experiment 1, 7 armed bandit\\ all algorithms, $\alpha=5\%$ ; 1040 replications. \label{fig::E1a5p}}
		\end{figure}
		\end{minipage}
		\hfill
		\begin{minipage}[c]{0.5\linewidth}
			\vspace*{4mm}
			\begin{figure}[H]
				\centering
				\includegraphics[height=.25\textheight]{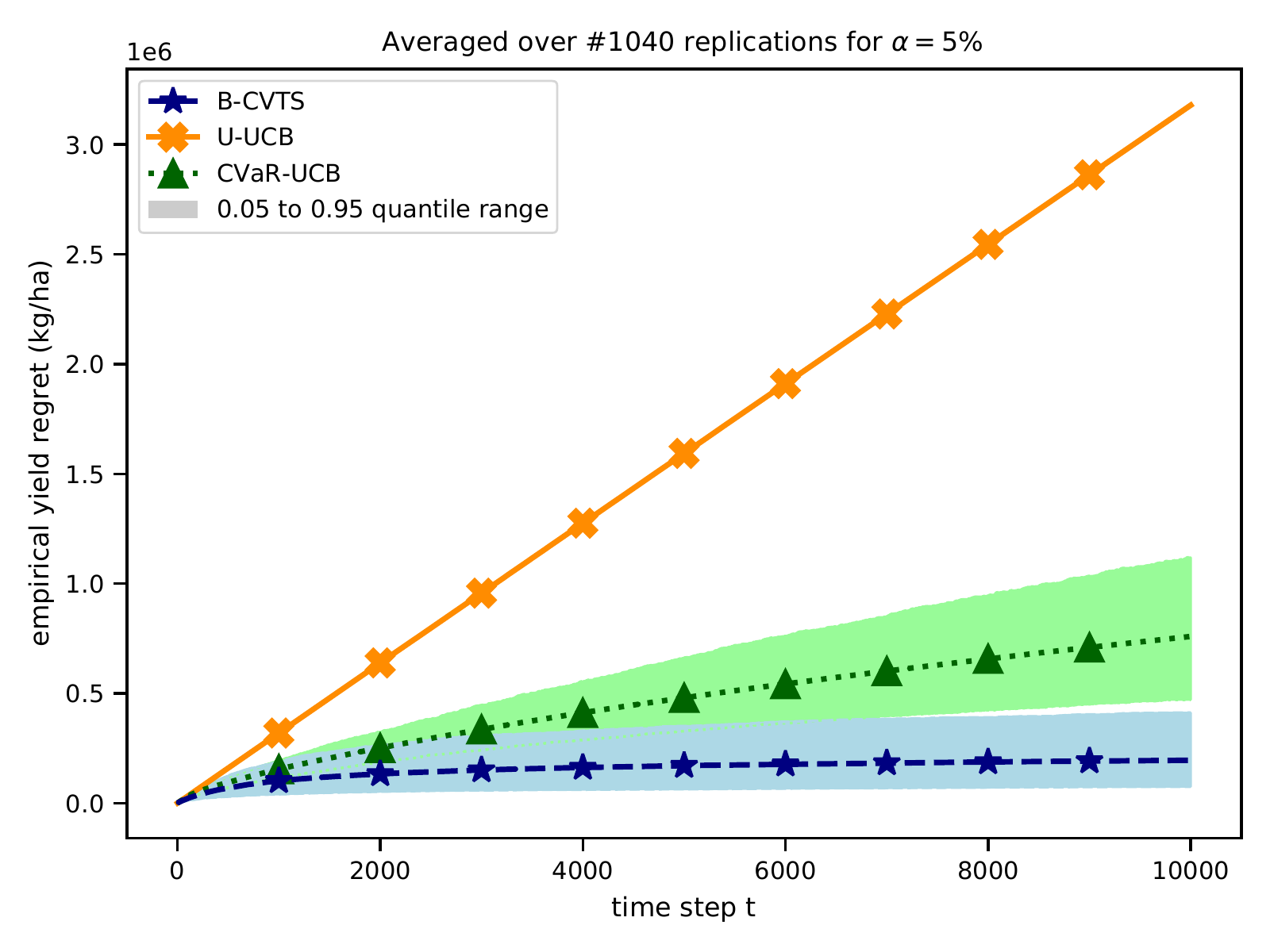}
				\vskip -5mm
				\caption{\texttt{DSSAT} Experiment 2, \\ 4 armed over-estimated support upper bound, \\ all algorithms, $\alpha=5\%$ ; 1040 replications.\\ \label{fig::E2a5p}}
			\end{figure}	
		\end{minipage}
		\hfill
	\end{figure}
	\hfill
	\begin{figure}[H]
		\captionsetup{justification=centering}
		\begin{minipage}[c]{0.5\linewidth}	
			\begin{figure}[H]
				\centering
				\includegraphics[height=.25\textheight]{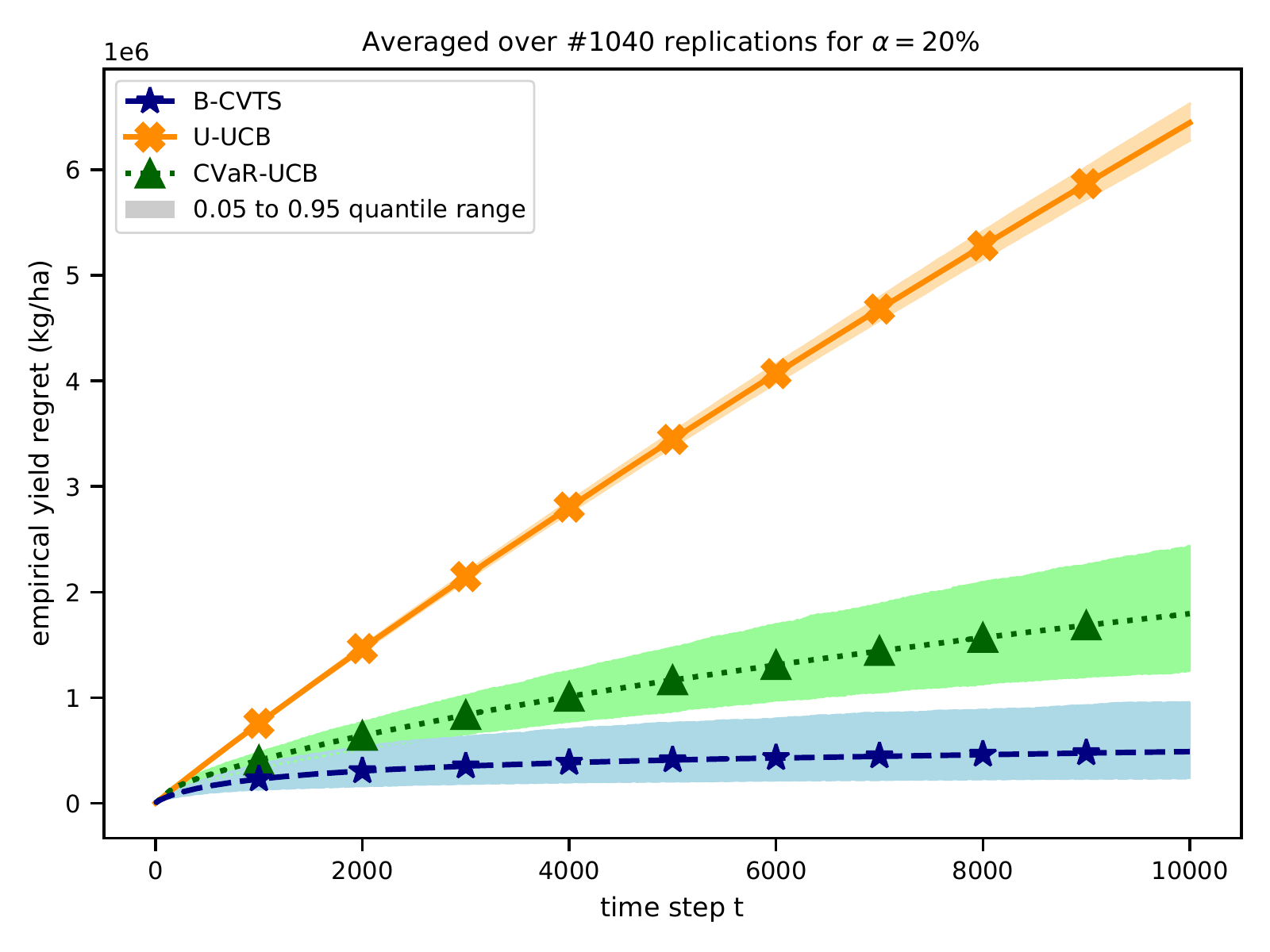}
				\vskip -5mm
				\caption{\texttt{DSSAT} Experiment 1,\\ all algorithms, $\alpha=20\%$ \label{fig::E1a20p}}
			\end{figure}	
		\end{minipage}
		\hfill
		\begin{minipage}[c]{0.5\linewidth}
			\begin{figure}[H]
				\centering
				\includegraphics[height=.25\textheight]{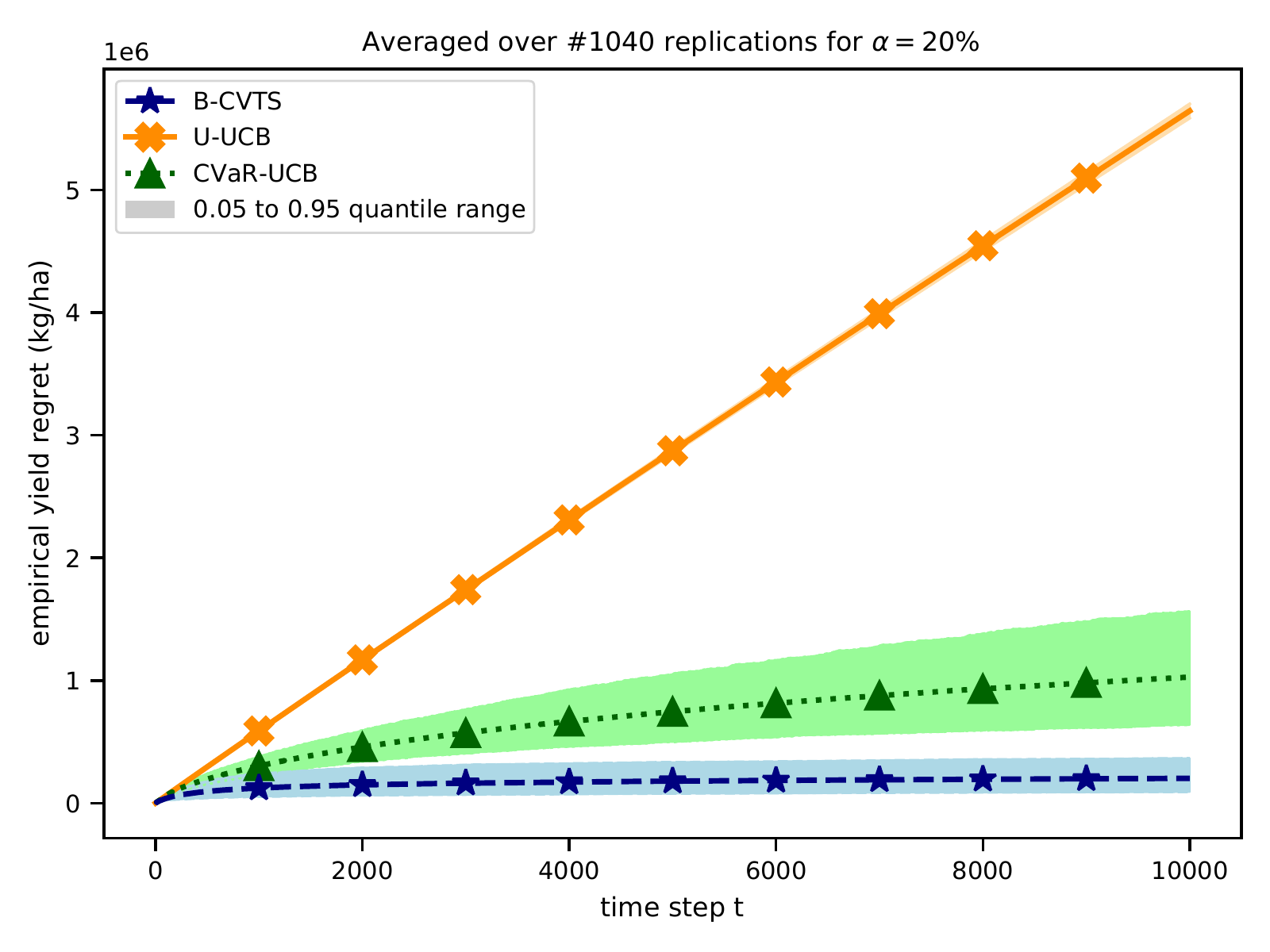}
				\vskip -5mm
				\caption{\texttt{DSSAT} Experiment 2,\\ all algorithms, $\alpha=20\%$ \label{fig::E2a20p}}
			\end{figure}
		\end{minipage}
		\hfill
	\end{figure}
	\hfill
	\begin{figure}[H]
		\captionsetup{justification=centering, belowskip=0pt}
		\begin{minipage}[c]{0.5\linewidth}
			\begin{figure}[H]
				\centering
				\includegraphics[height=.25\textheight]{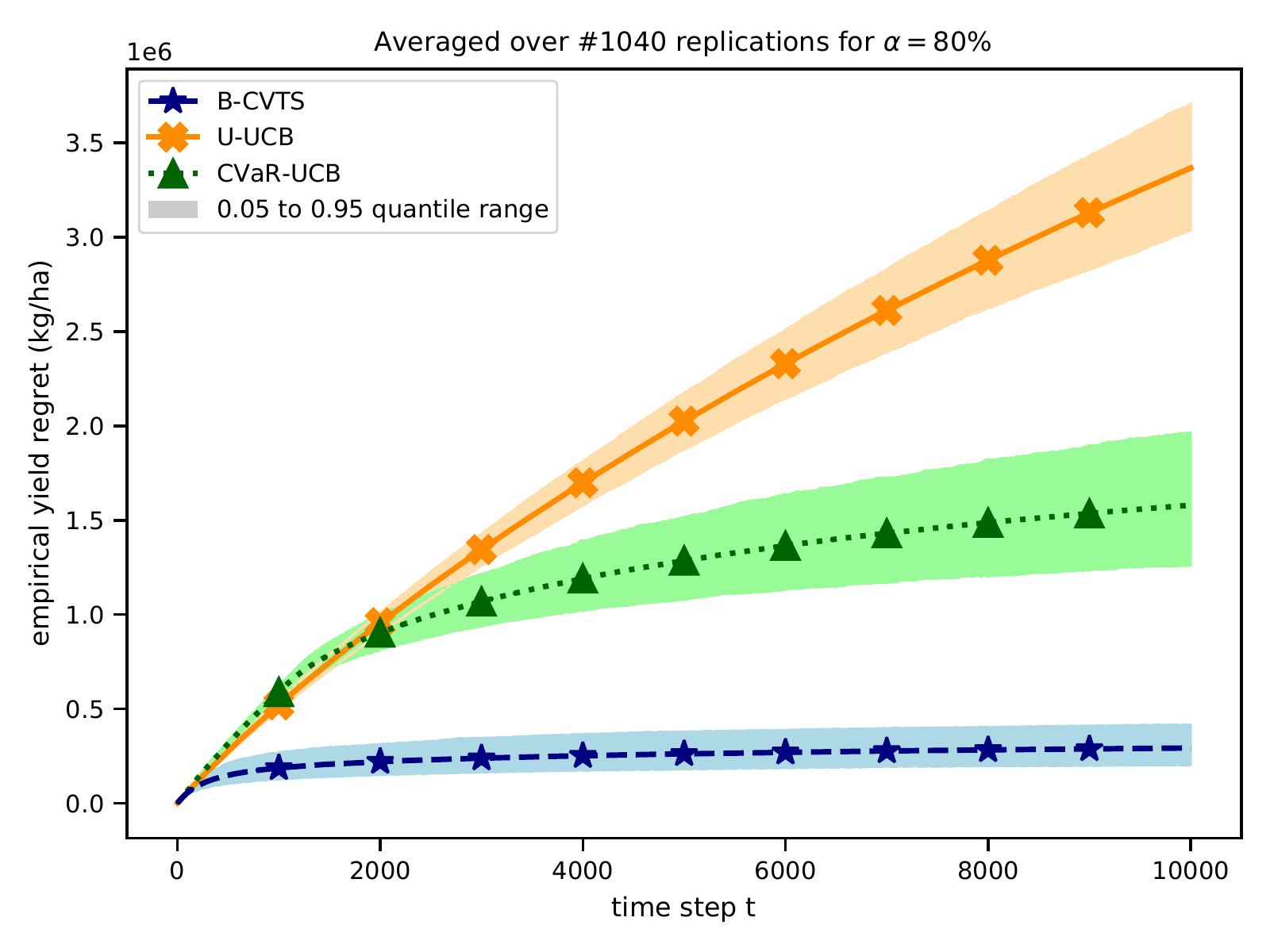}
				\vskip -5mm
				\caption{\texttt{DSSAT} Experiment 1,\\ all algorithms, $\alpha=80\%$ \label{fig::E1a80p}}
			\end{figure}
		\end{minipage}
		\hfill
		\begin{minipage}[c]{0.5\linewidth}	
			\begin{figure}[H]
				\centering
				\includegraphics[height=.25\textheight]{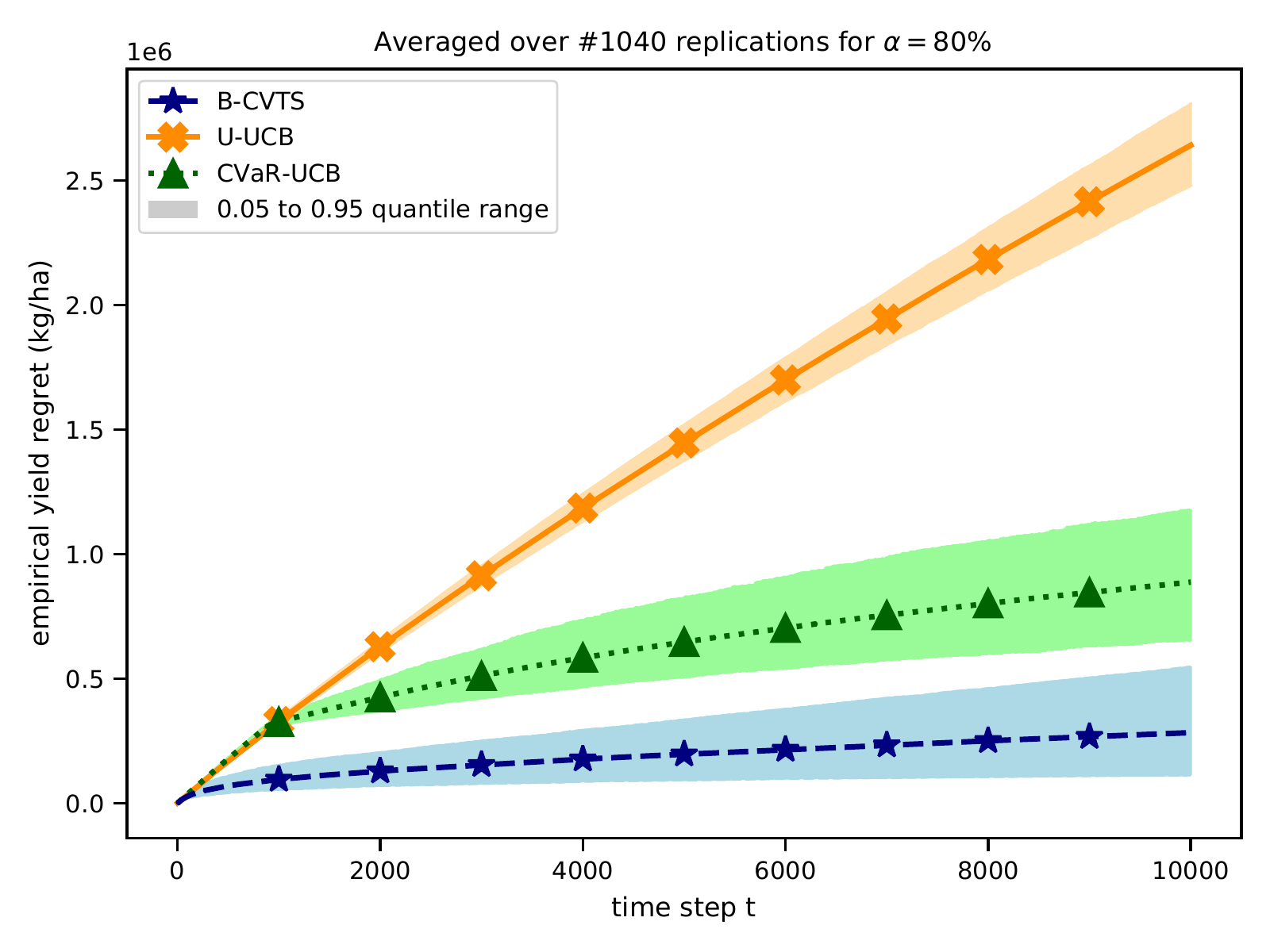}
				\vskip -5mm
				\caption{\texttt{DSSAT} Experiment 2,\\ all algorithms, $\alpha=80\%$ \label{fig::E2a80p}}
			\end{figure}	
		\end{minipage}
		\hfill
	\end{figure}
}

\end{document}